\documentclass[twoside,11pt]{article}

\usepackage{blindtext}
\usepackage{amssymb}
\usepackage{amsmath}
\usepackage{graphicx}
\usepackage{enumerate}
\usepackage{amsthm}

\usepackage{float}
\usepackage{subfigure}
\usepackage{algorithm}
\usepackage{bm}
\usepackage[noend]{algpseudocode}
\usepackage{lscape}
\usepackage[colorinlistoftodos]{todonotes}

\usepackage{jmlr2e}

\newtheorem*{theorem*}{Theorem}

\def\t{\textrm}
\def\d{\partial}

\def\R{\mathbb{R}}
\def\M{\mathbb{M}}
\def\E{\mathbb{E}}

\def\S{\mathbb{S}}
\def\M{\mathbb{M}}
\def\R{\mathbb{R}}

\def\g{\mathbf{g}}
\def\f{\mathbf{f}}

\def\A{\mathcal{A}}

\def\X{\mathcal{X}}
\def\S{\mathcal{S}}
\def\G{\mathcal{G}}

\def\g{\mathbf{g}}
\def\f{\mathbf{f}}

\usepackage{lastpage}
\jmlrheading{25}{2025}{1-\pageref{LastPage}}{1/21; Revised 5/22}{9/22}{21-0000}{Author One and Author Two}

\ShortHeadings{Atlas GP}{Niu et al.}
\firstpageno{1}

\begin{document}

\title{Atlas Gaussian processes on restricted domains and point clouds}

\author{\name Mu Niu \email mu.niu@glasgow.ac.uk \\
\name Yue Zhang \email 2416950z@student.gla.ac.uk \\
       \addr School of Mathematics and Statistics,\\
       University of Glasgow\\
       \AND
       \name Ke Ye \email keyk@amss.ac.cn \\
       \addr 
     KLMM, Academy of Mathematics and Systems Science,\\
     Chinese Academy of Sciences, China
    \AND
       \name Pokman Cheung \email pokman@alumni.stanford.edu\\
       \addr{} \\
       \name Yizhu Wang \email 2603214w@student.gla.ac.uk \\
\name Xiaochen Yang \email xiaochen.yang@glasgow.ac.uk \\
       \addr School of Mathematics and Statistics,\\
       University of Glasgow}

\editor{
}
\maketitle

\begin{abstract}
In real-world applications, data often reside in restricted domains with unknown boundaries, or as high-dimensional point clouds lying on a lower-dimensional, nontrivial, unknown manifold. Traditional Gaussian Processes (GPs) struggle to capture the underlying geometry in such settings. Some existing methods assume a flat space embedded in a point cloud, which can be represented by a single latent chart (latent space), while others exhibit weak performance when the point cloud is sparse or irregularly sampled. The goal of this work is to address these challenges. The main contributions are twofold:
(1) We establish the Atlas Brownian Motion (BM) framework for estimating the heat kernel on point clouds with unknown geometries and nontrivial topological structures; (2) Instead of directly using the heat kernel estimates, we construct a Riemannian
corrected kernel by combining the global heat kernel with local RBF kernel and leading to the formulation of Riemannian-corrected Atlas Gaussian Processes (RC-AGPs). The resulting RC-AGPs are applied to regression tasks across synthetic and real-world datasets. These examples demonstrate that our method outperforms existing approaches in both heat kernel estimation and regression accuracy. It improves statistical inference by effectively bridging the gap between complex, high-dimensional observations and manifold-based inferences.
\end{abstract}



\section{Introduction}\label{intro}
In recent years, data collection has increasingly involved  complex, constrained spaces with unknown boundaries or high-dimensional point clouds residing on a lower-dimensional, nontrivial, unknown manifold. For example, pollution data may be collected within geographic regions like lakes, where the exact boundaries are complex and unknown. Similarly, object rotations and movements in images or videos can form lower-dimensional manifolds embedded in high-dimensional pixel spaces.
In this work, we focus on the following regression model for such point cloud data: 
\begin{align} \label{eqn:reg}
y_j &= f(s_j) +\epsilon_j, \ \ \  \epsilon_j \sim \mathcal{N}(0, \sigma_{\epsilon}^2) 
\end{align}
where  $f:\M \rightarrow \R$ is an unknown regression function. $y_j \in \R$ is a response variable. $s_j= (s_{j}^1,\ldots,s_{j}^p) \in \M \subseteq \R^p$ is an observed predictor on a complete  compact and nontrivial  Riemannian manifold $\M$ embedded in $\mathbb{R}^p$. The manifold $\M$ can be parameterised by an atlas, which is a collection of some $q$-dimensional local coordinate systems(charts), where $q\leq p$. Let $\mathcal{D} =  \{ (s_j, y_j), j = 1, \ldots, n_d \} $ be the set of labeled observations with $n_d \geq 1$. 
Suppose we also have an unlabeled dataset $\mathcal{U} = \{ s_j, j=1, \ldots, n_u \}, $ where $n_u \geq 1$. The total number of points in the point cloud is $n=n_d+n_u$. Considering the regression model in \eqref{eqn:reg}, our goal is to make inferences about $f$ using the labeled dataset $\mathcal{D}$ and to predict $y$ values for the unlabeled dataset $\mathcal{U} $. Gaussian Processes (GPs) have been widely used as prior distributions for unknown functions. The GP with squared exponential (a.k.a. Radial Basis Function) kernel is a commonly used prior \citep{Rasmussen2006}. However, covariance functions relying on Euclidean distance can be inadequate when the input space $\M$ deviates from Euclidean geometry. On manifolds, such measures fail to reflect true relationships between points on $\M$, highlighting the need for approaches that incorporate the manifold's intrinsic geometry to support inference on complex data structures.

A major challenge in developing GPs on manifolds lies in constructing valid covariance kernels for these complex spaces. Recent advances in this area assume known manifold structure or simple topology that can be covered by a single chart. For instance, \citet{niu2019} defined intrinsic GPs on known manifolds by simulating Brownian Motion(BM) paths and estimating the transition densities. \citet{niu2023} extended this method to unknown manifolds,  though limited to flat, trivial topologies characterised by a single chart (latent space). \citet{lin2019extrinsic}
proposed extrinsic framework for GP modeling on manifolds, which relies on embedding the manifold into a Euclidean space. However such embeddings are only available for certain manifolds with known geometry.

Alternatively, several works approximate the manifold by constructing graphs over point clouds \citep{li2023inference, he2024scalable, bolin2022, dunson2020diffusion}, leading to Graph Laplacian GPs (GL-GPs). Here, kernels are estimated from eigenpairs of a graph Laplacian built using scaled Euclidean distances. While GL-GPs offer improvements over standard Euclidean GPs for manifold structured data, they can suffer when the point cloud is sparse or irregularly sampled, and are sensitive to hyperparameters such as bandwidth and neighbor counts \citep{garcia2020}. Moreover, smoothing spline approaches \citep{wood, tramsay} address noisy observations within regions of known boundaries in $\R^2$. While cases with unknown boundaries were discussed, they remain an open problem.

It is well known that a nontrivial manifold cannot be covered by a single chart unless it is homeomorphic to an open subset of $\R^q$ \citep[Section~1]{Lee13}. \citet{schonsheck2019chart} and \citet{stolberg2022} emphasize the necessity of using multiple charts to properly capture the structure of manifolds with nontrivial topology. Mapper \citep{singh2007topological}, a tool from topological data analysis, has been widely used to partition complex point clouds into locally simple subsets. Mapper has successfully uncovered global structures (e.g., Trefoil knots) \citep{floryan2022, paik2022atlas}, while Chart Autoencoders \citep{schonsheck2019chart, schonsheck2022semi} have also been employed for learning multiple charts in molecular dynamics data. In this work, we follow \citet{singh2007topological} and adopt Mapper as a preprocessing step to partition the point cloud into subsets. 

{\raggedleft {\bf Main Contribution}}:
We propose a novel methodology for constructing Gaussian processes 
on constrained domains with unknown boundaries and point clouds characterised by unknown geometry and nontrivial topology. Our key contributions are: 
\begin{itemize} 
\item We develop an \emph{Atlas Brownian Motion framework} to estimate the heat kernel on point clouds via simulation of BM paths across multiple learned charts. 
\item Instead of directly using the estimated heat kernel, we construct a \emph{Riemannian-corrected kernel}, combining the global heat kernel and local RBF kernel, and leading to the formulation of \emph{Riemannian-corrected Atlas Gaussian Processes (RC-AGPs)}. \end{itemize} 
The RC-AGP is constructed through three steps. First, we build a probabilistic atlas by learning multiple local charts from subsets of the point cloud, where the subsets are generated using the Mapper(Section~\ref{sec:atlas}). Second, we simulate BM paths on the Atlas by enabling probabilistic coordinate transitions across these charts, and estimate the heat kernel from the transition densities on the Atlas (Section~\ref{sec:density}). Third, we construct RC-AGPs by integrating the estimated heat kernel with the local RBF kernel, proving its semipositive definiteness to guarantee well-posed Gaussian process inference (Section~\ref{sec:RCAGP}).

The Atlas BM framework improves upon existing methods—such as single-chart BM and graph Laplacian approximations—in terms of heat kernel estimation accuracy. It is robust to variations in point cloud distribution and requires significantly fewer points to achieve accurate kernel estimates. By integrating the global heat kernel with local RBF kernels, the proposed RC-AGPs simultaneously capture the global geometry and ensure local smoothness, leading to superior predictive performance. We apply RC-AGPs on four regression tasks: two domains in $\mathbb{R}^2$ with unknown boundaries (the U-shape and the Aral Sea), a torus point cloud in $\mathbb{R}^3$, and a high-dimensional shark prey image cloud in $\mathbb{R}^{900}$. Across all tasks, RC-AGPs consistently outperform Euclidean GPs and GL-GPs.

\section{Background of Riemannian geometry}
\label{sec:2}
A manifold is a curved space that generalises the flat Euclidean space. A classic example is the unit sphere $\mathbb{S} \subseteq \mathbb{R}^3$. In general, a point cloud in $\R^p$ sampled from an unknown manifold can be described by probabilistic mapping functions and the corresponding latent spaces. In the context of differential geometry, the latent spaces can be interpreted as coordinate charts of the unknown manifold to be learned.   
\subsection{Atlas and metrics of manifolds}
\begin{figure}[h!]
\includegraphics[width=0.8\textwidth,height=0.45\textwidth]{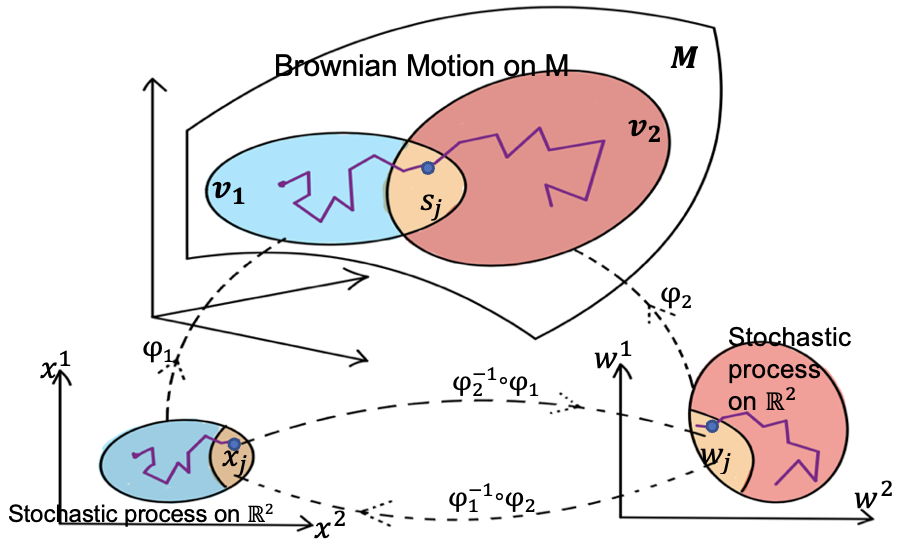} 
\centering\caption{ \label{Fig:ChangeCoord}    Atlas with two charts on $\M$ and change of coordinates operations.  }  
\end{figure}

Let $\M \subseteq \mathbb{R}^p$ be a $q$-dimensional manifold where $q<p$. An \emph{atlas} consists of charts that, roughly speaking, locally parametrizes the manifold. It is a collection $\A=\{ (V_i,\varphi_i)\}_{i \in I}$,  where $I$ is an index set, $V_i\subset \M$ is an open subset for $i\in I$ such that $\bigcup_{i \in I} V_i = \M$. $\varphi_i$ is a mapping (homeomorphism) from an open subset of $\mathbb{R}^q$ to the corresponding region $V_i$ on the manifold. For example, Fig.~\ref{Fig:ChangeCoord} illustrates two overlapping coordinate charts of a $2$-dimensional manifold, namely $\varphi_1: X \rightarrow V_1\subset \M$ (blue) and $\varphi_2: W \rightarrow V_2\subset \M$ (red). Given a point $s_j$ in the intersection $V_1\cap V_2$ (yellow), it is represented via $\varphi_1$ by a point $x_j=\varphi_1^{-1}(s_j)$ in $X$ and also via $\varphi_2$ by a point $w_j=\varphi_2^{-1}(s_j)$ in $W$.
 The two local representations are related by the transition operation $w_j = (\varphi_2^{-1} \circ \varphi_1) (x_j) $, also known as the \emph{change of coordinates}. The BM path on $\M$ can be interpreted as the two stochastic processes in $X$ and $W$.

The \emph{tangent space} of $\mathbb{M}$ at a point $s\in \M$ is the $q$-dimensional subspace $\mathbb{T}_s \mathbb{M} \subseteq \mathbb{R}^p$ such that $s + \mathbb{T}_s \mathbb{M}$ is tangent to $\mathbb{M}$ at $s$. The (induced Riemannian) \emph{metric} \citep{Lee18} on $\mathbb{M} \subseteq \mathbb{R}^N$ is defined for each $s\in \M$ as the inner product on $\mathbb{T}_s \mathbb{M} \subseteq \mathbb{R}^p$, obtained by restricting the standard one on $\mathbb{R}^p$. 
This metric plays a crucial role in simulating BM on 
$\M$, as it encodes the local geometry and governs the stochastic process. It also defines the geodesic distance function, which is in general not equal to the Euclidean distance.
In a chart $(V,\varphi)$ of $\M$, the metric is represented by a symmetric positive definite matrix-valued function $\g$ on $V$. If $\mathcal J$ denotes the Jacobian of $\varphi$, then we have 
\begin{align} \label{eqn:deter_metric}
\g = \mathcal{J}^T \mathcal{J}, \ \ \  \mathcal{J}^{l,k} = \frac{\partial \varphi^l }{ \partial  x^k},
\end{align}
where the superscripts indicate the $l^\text{th}$ dimension of the chart and the $k^\text{th}$ dimension of the observation space. The determinant of $\g$ can express how much a part of chart(latent space) is stretched or bend when mapped to $\M$.
Moreover, the Laplace-Beltrami operator $\Delta_s$ of $\M$ can be expressed in local coordinates as
$\Delta_s f =\sum_{l,k=1}^{q} \frac{1}{\sqrt{G}} \frac{\partial}{\partial x^k} \left (    \sqrt{G} \g^{lk} \frac{\partial f}{\partial x^l}  \right ),$
where $G$ is the determinant of $\g$, $\g^{lk}$ is the $(l,k)$ element of its inverse and $f$ is a smooth function on $\M$. In the special case where $\M = \R^q$ with its standard coordinate chart, $\g$ is the $q\times q$ identity matrix at every point and the Laplace-Beltrami operator $\Delta_s$ becomes the Laplace operator $\Delta = \sum_{l,k=1}^q \frac{\partial^2}{\partial x^l \partial x^k}$.
\subsection{Heat kernel and BM on $\M$ } 
\label{sec:Heat}
Consider the heat equation on $\M$, given by
\begin{align*}
\frac{\d}{\d t}h(s_0,s_j,t)=\frac{1}{2}\Delta_s h(s_0,s_j,t),\qquad  s_0, s_j\in \M,
\end{align*}
where $s_0$ and $s_j$ are points on $\M$, $t \in \R^{+}$ is the diffusion time, and $\Delta_s$ is the Laplacian-Beltrami operator on $\M$. A heat kernel of $\M$ is a smooth function $h(s_0,s_j,t)$ on $ \M \times \M \times \R^{+}$ that satisfies the heat equation and the initial condition $\lim_{t\rightarrow 0}h(s_0,s_j,t)=\delta(s_0,s_j)$, where $\delta$ is the Dirac delta function. It can be interpreted as the amount of heat that is transferred from $s_0$ to $s_j$ in time $t$ via diffusion. The heat kernel becomes unique when we impose a suitable condition along the boundary $\partial \M$, such as the Neumann boundary condition. If $\M$ is a Euclidean space $\R^q$,  the heat kernel has a closed form expression corresponding to a time-varying Gaussian function:
$h(s_0,s_j,t)
=\frac{1}{(2\pi t)^{q/2}}\,
  \exp\left\{-\frac{||s_0-s_j||^2}{ 2t }\right\}, \; s_j\in \mathbb R^q.$
The diffusion time $t$ controls the rate of decay of the covariance.

Solving the heat equation on arbitrary manifolds presents substantial challenges due to the complexity of their geometric structure \citep{chavel1984}. To avoid directly solving the heat equation, \citet{niu2019} proposed estimating the heat kernel by simulating BM paths on a known manifold and using the transition density as an approximation. However, their approach is limited to manifolds that can be described with a single chart.


\section{Learn the probabilistic atlas $\A$ of $\M$} \label{sec:atlas}
\begin{figure}
    \centering
    \includegraphics[width=0.95\linewidth]{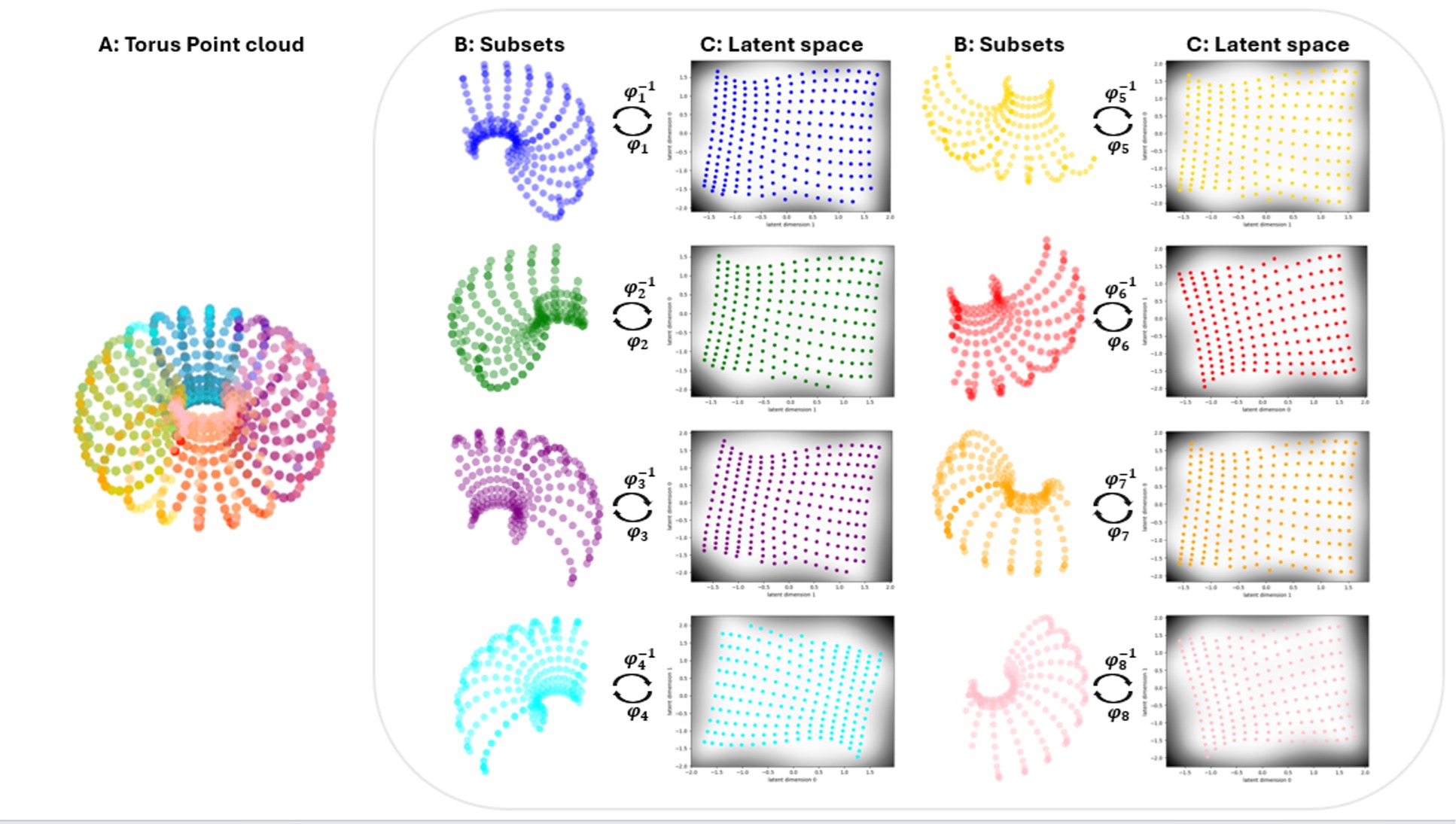}
    \caption{ The torus atlas and point cloud.}
\label{fig:Toruslatent}
\end{figure}

To learn an unknown manifold from a point cloud,  possibly with a nontrivial topology, we first create subsets that form an open cover of the manifold as a preprocessing step \citep{KeplerMapper2019}. Let $\mathcal{S} = \{ s_j | j = 1, \dots, n  \}$, with $s_j \in \mathbb{R}^p$ and $n = n_d + n_u$, represent the points in the cloud, including labeled points from $\mathcal{D}$ and unlabeled points from $\mathcal{U}$. 
The point cloud is covered by $n_v$ subsets: $\S = \S_1 \cup \S_2 \cdots \S_i \cdots \cup \S_{n_v}$, where $\S_i = \{ s_{ij}|j=1,\cdot \cdot \cdot,ns_{i}  \}$ and $ns_i$ is the number of points in the $i$th subset. The intersections of two sets are nonempty, $\S_i \cap \S_{i+1} \neq \O$. An example of torus subsets are shown in column B of Fig.~\ref{fig:Toruslatent}, with each subset marked in a different color. Further details on the preprocessing are provided in Section J. of the Appendix. 

Given a subset $\S_i$, we can learn the coordinate chart and mapping function.  For each subset $\S_i$, we can define a latent variable model, which introduces a set of unobserved variables $\X_i= \{x_{ij} | j= 1,\cdot \cdot \cdot, ns_i \}$, $x_{ij} \in \R^q$. The latent $\X_i$ is associated with $\S_i$ observed in a higher-dimensional space as follows:
\begin{align}
\label{eqn:learnMap}
s_{ij} = \varphi_i(x_{ij}) +\epsilon_{ij}, \ \ \  s_{ij}\in \S_i,
\end{align}
where $\epsilon_{ij}$ is an iid noise term. $x_{ij}$ is a latent variable. $i$ is the index of subset. If $p$, the dimension of the data space, is larger than $q$, the dimension of the latent space, \eqref{eqn:learnMap} can be seen as a model for dimension reduction. Various approaches can be used to learn the mapping $\varphi_i$ and the latent space $\X_i$, such as the Gaussian process latent variable model~(GPLVM,\cite{lawrence2005}) and  autoencoder~(AE,\cite{baldi2012}). Both approaches have been used in this work. We present the construction of  GPLVM atlas in the main paper. The AE atlas is explained in Section B. of the Appendix. If $p$ is the same as $q$, the latent space becomes a subset of $\R^p$. This lead to a natural chart for $\M$.  In the U-shape example in Section \ref{sec:u}, we have $q=p=2$. $\R^2$ becomes a natural chart. The mapping becomes an identity mapping. 
  
GPLVM \citep{lawrence2005} is a GP-based dimensionality reduction method designed to learn a probabilistic mapping between a low-dimensional latent space and observed data. The probabilistic model is computed by placing a GP prior over the mapping, $\varphi_i (x) \sim \mathcal{GP} (0,k(x,x'))$, where $k$ is the RBF kernel. The latent variables $\{ x_{ij} | j=1,\cdot \cdot \cdot, ns_i  \}$ are inferred by maximizing the log-likelihood of the latent variable model. The predictive distribution of GPLVM acts as the forward map, denoted by $\varphi_i$, enabling the transformation of a latent point $x_{i*}$ into its corresponding observation space point $s_{i*}$. Conversely, the backward mapping $\varphi_i^{-1}$ facilitates the estimation of a latent point $x_{i*}$ given an observation $s_{i*}$. In practice, the backward map is approximated via an optimization procedure that maximizes the log-likelihood conditioned on the observed data. Using both the forward and backward maps, we can map points between different charts. Given two charts $i$ and $j$, the {\it transition map} is defined as $\varphi_i \circ \varphi_j^{-1}$. Further details regarding the forward and backward mappings in the GPLVM framework, are provided in Section C. of the Appendix.

An example of Torus atlas is shown in Fig.~\ref{fig:Toruslatent}. Eight subsets are created from the torus point cloud consisting of 625 points. Given a three dimensional torus subset $\S_i$ in column B of Fig.~\ref{fig:Toruslatent}, we learn a two-dimensional latent space $\X_i$ and the corresponding mapping function $\varphi_i$ in column C. The background shading in the latent spaces represents the uncertainty of mapping $\varphi_i$, with darker areas indicating higher variance. Applying GPLVM to each subset separately results in multiple charts. The collection of all learned charts becomes the probabilistic atlas $\A$ of $\M$. The intersections of the subsets $\S_i \cap \S_{j}$ on $\M$ are used to identify the overlapping regions within the latent space (see Section F. of the Appendix for details). 



To learn the associated metric tensor of the $i^\text{th}$ chart (latent space), we also need to estimate the Jacobian of $\varphi_i$. Since the RBF kernel is differentiable, \cite{Rasmussen2006} show that the derivative of a GP is also a GP. Therefore, given a point in the latent space, we can derive the distribution of the Jacobian $p(\mathcal{J}_i | \X_i , S_i ) = \prod_{l=1}^p \mathcal{N} ( \mu_{J_i}^l, \Sigma_{J_i}  )$. The expressions for $\mu_{J_i}^l$ and $\Sigma_{J_i}$ are shown in Section C. of the Appendix. Given \eqref{eqn:deter_metric} , the resulting metric $\g_i$ of the $i^\text{th}$ chart follows a non-central Wishart distribution \citep{anderson1946}. The expected metric tensor $\E(\mathbf{g}_i)$ is defined in \eqref{eqn:GPLVMmetric}. 
\begin{align}
\label{eqn:GPLVMmetric}
\E[\mathbf{g}_i]=\mathcal{G}_i = \E(\mathcal{J}_i^T) \E(\mathcal{J}_i) + p \Sigma_{J_i}.
\end{align}

\section{Estimate the heat kernel via BM on $\M$} \label{sec:density}
Once the atlas is learned, we will estimate the heat kernel as the BM transition density on $\M$. BM paths are simulated using the learned mappings and metrics of the atlas from Section \ref{sec:atlas}.

\subsection{BM on $\M$ as the SDE on the atlas $\A$} \label{sec:bm}

For a single chart manifold, simulating BM sample paths on $\M \subset \R^p$ is equivalent to simulating the stochastic process in the chart in $\R^q$, $q\leq p$. BM on a Riemannian manifold in a local coordinate system (chart) is given as a system of stochastic differential equations (SDEs) in It$\hat{o}$ form \citep{hsu1988,hsu2008}. Given the atlas $\A=\{ (V_i,\varphi_i) | i \in \{ 1,\cdots,n_v  \} \}$ on $\M$, where $n_v$ is the number of the charts and  $\cup_{i=1}^{n_v} V_i = \M$, we can define a group of SDEs on $\A$ that is equivalent to the BM on $\M$. For each $\varphi_i$ in $\A$, we can compute the expected metric $\G_i = \E[\g_i]$ as in \eqref{eqn:GPLVMmetric} and the SDE in the $i^\text{th}$ chart can be derived as 
\begin{align}
\label{eqn:swBM}
dx_i^l(t) = \frac{1}{2}{G_i}^{-1/2} \sum^{q}_{r=1}\frac{\partial}{\partial x_i^r} \left(  {\G_i}^{lr} {G_i}^{1/2} \right) dt + \left( {\G_i}^{-1/2} dB_i(t)\right)_l,
\end{align}
where $x_i^l$ represents the coordinate in the $l^\text{th}$ dimension of chart $i$ (latent space $i$). $\G_i$ is defined in \eqref{eqn:GPLVMmetric}, $G_i$ is the determinant of $\G_i$, and $B_i(t)$ represents an independent BM in the Euclidean space.


\begin{figure}[h!]
\includegraphics[width=1\textwidth,height=0.35\textwidth]{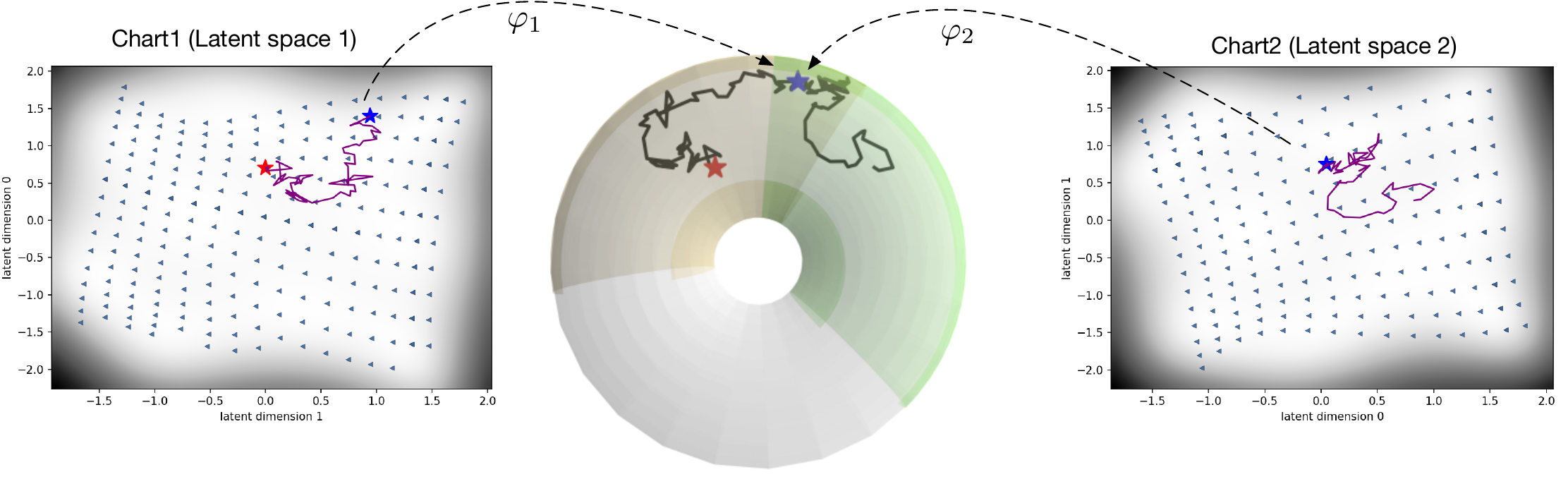} 
\centering\caption{ \label{Fig:switch} A demonstration of simulating a BM path by switching the stochastic process in different charts (or latent spaces). The collection of charts forms the atlas of the torus manifold. The black trajectory on the torus (middle panel) represents a BM path with a red star indicating the starting point. The green region of the torus is described by Chart 2 on the right. The brown region of the torus is describe by Chart 1 on the left. The purple trajectory in Chart 1 represents a stochastic process simulated by \eqref{eqn:swBM} with $i=1$. Once the path reaches the overlapping region, indicated by the blue star $x_{1j}$ on the left panel, the blue star are transferred to Chart 2 using the change of coordinates operation: $x_{2j} = \varphi_2^{-1} \circ \varphi_1 ( x_{1j} )$. The stochastic process continues in Chart 2, as shown by the purple trajectory in the right panel (simulated by \eqref{eqn:swBM} with $i=2$). The projection of both segments of purple lines from the charts onto the torus generates the  BM path shown in the middle panel. }
\end{figure}

Since each chart covers only a part of the original manifold, we introduce the probabilistic change of coordinates operations for transitioning SDEs between charts within an atlas. Specifically, BM trajectories are simulated by transitioning between different local coordinates using \eqref{eqn:swBM}  and the proposed transition map. Fig.~\ref{Fig:switch} serves as a demonstrative example. Consider a point cloud representation of the manifold $\M$ shown in the middle panel. Two overlapping subsets (green and brown) on $\M$ are used to learn two charts. While the atlas $\A$ may consist of multiple charts to cover $\M$, we illustrate the process with two charts for simplicity. Chart 1 (left panel) parameterizes the brown region, and Chart 2 (right panel) parameterizes the green region. The intersection of these regions on $\M$ is non-empty, defining an identifiable overlapping region. This overlapping region serves as the switching domain in the latent space. In the middle panel, the red star denotes the starting point of the BM path. It can be mapped to chart~1 via $\varphi_1^{-1}$. In Chart 1, the stochastic process is depicted as the purple line. Upon entering the switching domain, the step of SDE, marked by a blue star in Chart~1, is mapped back to $\M$ (the overlapping region) using $\varphi_1$ and subsequently to Chart~2 via $\varphi_2^{-1}$. The transition map from Chart~1 to Chart~2 is given by $\varphi_2^{-1} \circ \varphi_1 $. The ensuing trajectory is simulated as the stochastic process( purple line) originating from the blue star in Chart~2 in the right panel. When we map the purple trajectories from Chart~1 and Chart~2 to $\M$, they form the black BM path in the middle panel.

While this example involves two charts covering the overlapping region, in general, multiple charts may cover the same region. In such cases, there are several target charts to which the simulated steps can transit to. According to Theorem~\ref{th:sw}, simulations using any target chart result in equivalent steps on $\M$. We can randomly pick one of the target charts and use the transition map to switch the chart. The proof of Theorem \ref{th:sw} is given in Section A.2 of the Appendix. To ensure that BM steps are directed away from regions of high uncertainty where the mapping is not reliable, boundaries in the latent space are established based on the variance of $\varphi_i$. The process of simulating BM paths is summarised in Algorithm \ref{alg:bm}. 

\begin{theorem}
\label{th:sw}
If the region on $\M$ can be parameterised by multiple charts, the stochastic process defined in \eqref{eqn:swBM} is chart independent. With a given diffusion time, simulations in any choice of charts are equivalent to the same step in $\M$.
\end{theorem}

\begin{algorithm}[!h]
\caption{ { Simulating Brownian motion sample paths on $\M$ } }
\label{alg:bm}
\begin{algorithmic}
\State {\bf Step 1}:Construct a probabilistic atlas by learning multiple charts from point cloud subsets, with chart boundaries defined by mapping uncertainty. Overlapping regions are identified through subset intersections.
\State {\bf Step 2}: Simulate SDE sample paths on the atlas $\A$:
\For{$\kappa = 1, \ldots, N_s$}    \Comment{ {\small $N_s$ is the No. of starting points}  }
\For{$j = 1, \ldots, N_{bm}$}    \Comment{ {\small$N_{bm}$ is No. of sample paths} }
\For{$ \tau = 1, \ldots, T$}    \Comment{ {\small T steps BM, $T*\Delta t$  $\rightarrow$ max diffusion time} }
                 
\State {\bf do} Propose $x_i(\tau)$ using \eqref{eqn:swBM} until $x_i(\tau)$ is inside the boundary of chart $i$ 
\State {\ \  } $q\left(x_i(\tau)|x_i(\tau-1)\right) \leftarrow \mathbb N \left( x_i(\tau) |  \mu \left( x_i\left(\tau-1\right),\Delta t \right)  , \Delta t \G_i^{-1} \right )$  
  
\State {\bf While} $x_i(\tau)$ is outside the boundary of chart $i$
\State{ \Comment{ the boundary is defined by the uncertainty of $\varphi_i$} }
\State {\bf If} $x_i(\tau)$ is in the overlapping region of chart $i$
\State $\ \   \ \  $ Move $x_i(\tau)$ to the target chat $i_s$ using transition map. 
\State $ \ \  \ \  $  If there are multiple target charts,  randomly pick one for $i_s$.  
\State $ \ \ \ \ $ $x_{i_s}(\tau) = \varphi_{i_s}^{-1} \circ \varphi_{i}( x_i(\tau) )$  
 \Comment{ Theorem \ref{th:sw} }

\EndFor
\EndFor
\EndFor
\Return $\bm x$ and chart indices.
\end{algorithmic}
\end{algorithm}



\subsection{Heat kernel estimation as transition density of BM on the atlas $\A$} \label{sec:kernel} 

\begin{figure}[h!]
\centering
\subfigure[Sample paths of atlas BM]{ \label{fig:Multi-BM}  \includegraphics[width=0.45\textwidth,height=0.35\textwidth]{ 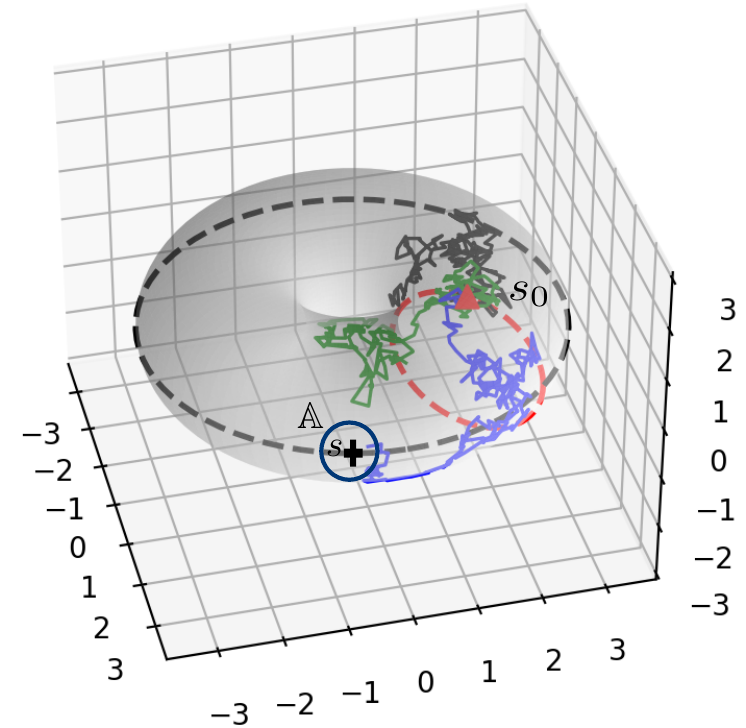} }
\subfigure[ Sample path of single chart SDE ]{ \label{Fig:One-BM}  \includegraphics[width=0.45\linewidth,height=0.35\textwidth]{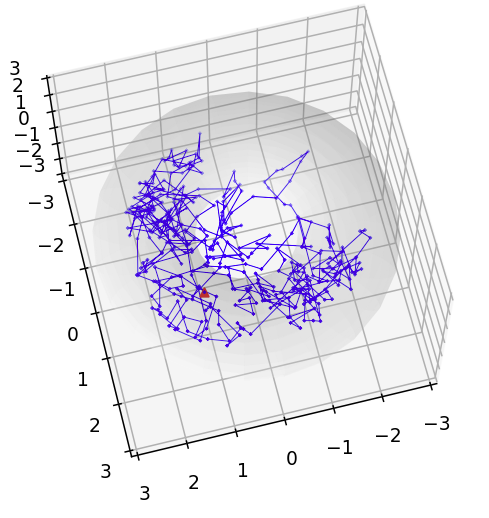}}
\centering\caption{ \label{fig:bmdemo}   { BM paths on $\M$:  (a) $s_0$ (red triangle) is the starting point of three BM paths (black, green and blue solid lines). Only the blue path reaches the neighbourhood $\mathbb{A}$ (blue circle) of $s$ (black cross) at time $t$. Therefore the estimate of the transition probability $p( {S}(t) \in \mathbb{A} \ | \ {S}(0)=s_0 ) = \frac{1}{3}$. (b) 
The blue path, generated using the SDE on a single chart \citep{niu2023}, starts at the red triangle and traverses through the central void of the torus, demonstrating that it does not correspond to a proper BM on the torus.
} }  
\end{figure}

We estimate the heat kernel by simulating BM sample paths on $\M$ and numerically evaluating the transition density. Let $\{ S(t) : t > 0 \}$ denote BM on $\M$, starting at $S(0)= s_0$. We generate $N$ sample paths by \eqref{eqn:swBM} and transition maps. For any $t > 0$ and $s \in \M$, the transition probability $p( {S}(t) \in \mathbb{A} \ | \ {S}(0)=s_0 )$
is approximated by the proportion of BM paths that reach $\mathbb{A}$ at time $t$, where $\mathbb{A}$ is a small neighborhood of $s$. 
An illustrative diagram is presented in Fig.~\ref{fig:Multi-BM}. The red triangle denotes the starting point of the BM paths. Three paths are simulated using the SDES on the atlas of torus as in Fig.~\ref{fig:Toruslatent}. Of these paths, only one reaches the neighborhood of the blue cross (target point), resulting in a transition probability of 
1/3. For comparison, Fig.~\ref{Fig:One-BM} presents a trajectory simulated using SDEs on a single-chart torus, following the approach in \cite{niu2023}. The single-chart approach results in the path leaving the torus surface and crossing through the central void, illustrating the limitations of the single-chart approach.

The transition density of BM at $s$ can be approximated as:
\begin{align}
\label{eqn:kheat}
\hat{h}^t (s_0,s) \approx  \frac{p( {S}(t) \in \mathbb{A} \ | \ {S}(0)=s_0 )}{V(\mathbb{A})} \approx \frac{1}{V(\mathbb{A})}\cdot \frac{N_{A}}{N} 
\end{align}
where $\hat{h}^t (s_0,s)$ is the heat kernel estimator, $N$ is the number of simulated BM paths and $N_{A}$ is the number of BM paths which reach $\mathbb{A}$ at time $t$, $V(\mathbb{A})$ is the Riemannian volume of $\mathbb{A}$. Note that the BM diffusion time $t$ operates akin to the smoothing parameter in the RBF kernel. A larger $t$ implies a higher probability of BM reaching the neighbourhood of the target point, consequently leading to increased covariance, and vice versa. Given Theorem \ref{th:unbiased}, this estimator is asymptotically unbiased and consistent. The proof of Theorem \ref{th:unbiased} is given in Section A.3 of the Appendix.

\begin{theorem}
\label{th:unbiased}
The estimator $\hat{h}^t (s_0,s)$ in \eqref{eqn:kheat} is asymptotically unbiased.
\end{theorem} 

\begin{figure}[h!]
\includegraphics[width=0.7\textwidth,height=0.45\textwidth]{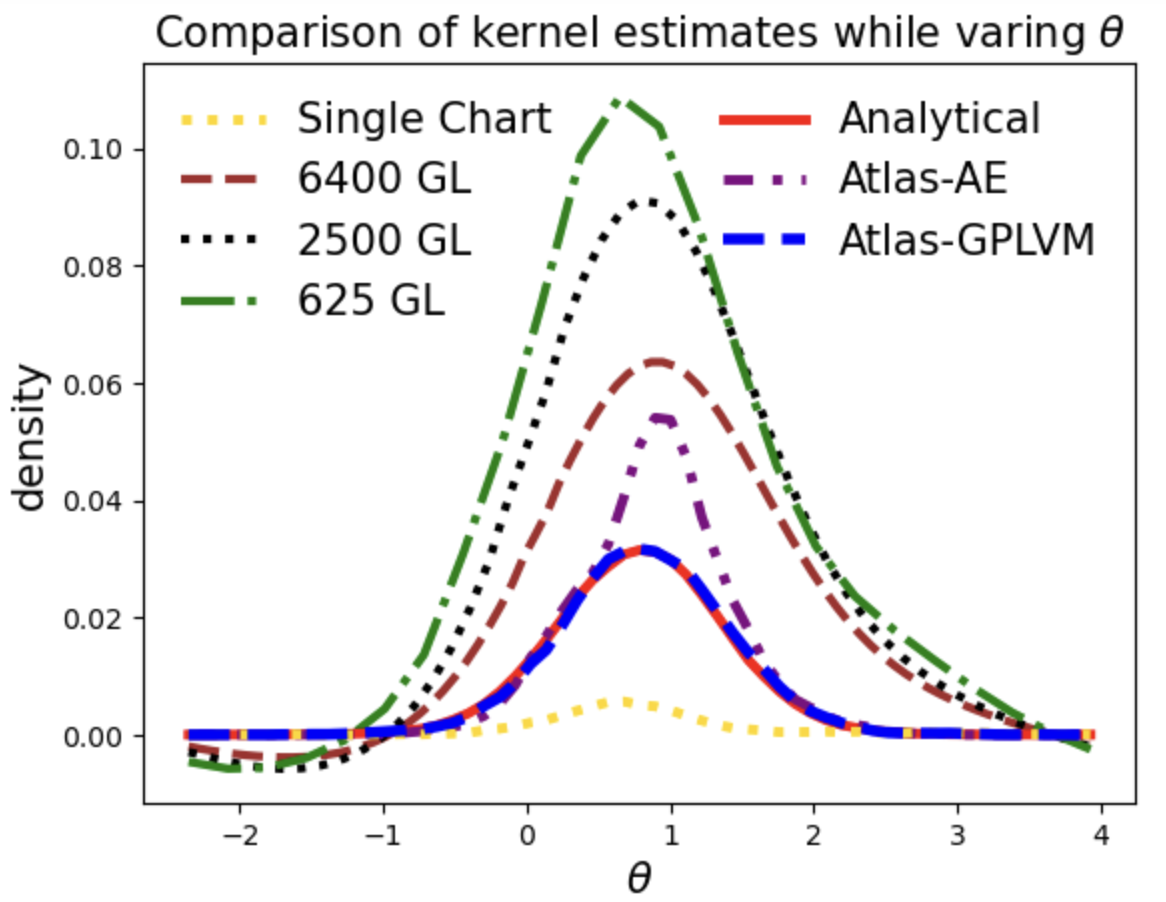} 
\centering\caption{ \label{fig:compKtheta}   {Kernel estimates comparison. The analytical kernel density is plotted as a solid red line. The kernel estimates using BM transition density with the GPLVM atlas are represented by a dashed blue line. The kernel estimates using BM transition density with the AE atlas are shown as a dashed purple line. Both atlases are constructed using 625 points on the Torus. The GL kernel estimates are plotted as a green dash-dotted line when the number of points is 625. We also plot the GL estimates as a black dotted line and a brown dashed line when the number of points increases to 2,500 and 6,400, respectively. The kernel estimates using single chart BM are shown in yellow dotted line.}  }
\end{figure}

When the estimated manifold closely approximates the true manifold, the resulting BM transition density estimates will resemble those obtained through analytical parameterisation. We utilise the Torus as a case study. By assuming the Torus's geometry is known, we can adopt the methodology outlined by \cite{niu2019} to assess the heat kernel  ${h}^t (s_0,s)$ through BM transition density using the analytical metric tensor. The derivation of the analytical metric and the parameterisation of the Torus are detailed in Section I. of the Appendix. Consider the red triangle in Fig.~\ref{fig:bmdemo}  as our starting point for the BM simulation. The BM transition density is evaluated with 60 evenly distributed target points along the black horizontal dashed circle  in Fig.~\ref{fig:bmdemo}. The hyper-parameter $t$ (diffusion time) remains constant at 4. Results are depicted as solid red lines in Fig.~\ref{fig:compKtheta}, where the horizontal axis represents the outer circle angle $\theta$ of the Torus, and the vertical axis represents the transition density.


In the case of the point cloud, the geometry and topology of the Torus are unknown.  Following the approach in Section \ref{sec:atlas}, the atlas can be learned using GPLVM and AE. BM trajectories are simulated on the estimated atlas, and the corresponding heat kernel estimates with GPLVM metrics are represented by blue dashed line in Fig.~\ref{fig:compKtheta}. The estimates using AE metric are plotted as the purple dashed line in Fig.~\ref{fig:compKtheta}. It is evident that these GPLVM results closely match the red solid line. Heat kernel estimates derived from the Graph Laplacian (GL) method \citep{dunson2020diffusion} across various data scenarios are depicted in Fig.~\ref{fig:compKtheta}. Initially, when the Torus dataset comprises 625 points, the GL kernel estimates are represented by the green dash dotted line, noticeably deviating from the solid red line for analytical kernel estimates. However, as the number of grid points escalates to 2500 and 6400, the GL kernel estimates shift to the black dotted line and brown dashed line, respectively. With this increase in the number of points, the estimates progressively converge toward the solid red line. Compared to the GL method, kernel estimates utilising GPLVM metrics exhibit superior performance even with fewer points on the Torus. We also applied the single chart BM method \citep{niu2023}, with the estimates plotted as the yellow dotted line. The results demonstrate that the single chart method does not align with the analytical estimates. Examples of the latent space and BM trajectories for the single chart method are available in Section E. of the Appendix. To further illustrate the robustness of our BM-on-atlas approach, we modified the configuration of the Torus to generate a non-regular, non-uniformly distributed point cloud. Even under these conditions, the kernel estimates produced by our method closely match the analytical results. Additional results are provided in Section D. of the Appendix. The numerical accuracy of the heat kernel estimates for the Torus is examined by varying the number of simulated BM paths, as shown in Fig.~\ref{fig:toruskernel} of Appendix I. With $2e+4$ simulated paths, the estimates closely match the analytical solution.

\section{Construct the Riemannian-Corrected Atlas GP}
\label{sec:RCAGP}
A straightforward approach to constructing the atlas GPs is to use the heat kernel estimates in \eqref{eqn:kheat} for the covariance function in GPs. However, this method requires simulating BM paths for all grid points.
Let $N_{bm}$ denote the number of paths simulated per point  and $n$ represent the total number of points on $\M$. The total number of simulated paths becomes $n \times N_{bm}$, which can lead to substantial computational costs. 

To address this, we propose the Riemannian-Corrected Atlas GP (RC-AGP), which leverages the estimated heat kernel to model the covariance structure between different subsets. The RC-AGP combines the estimated heat kernel with the RBF kernel to balance global and local modeling needs: the heat kernel allows the RC-AGP to respect the intrinsic geometric structure of $\M$, while the RBF kernel enhances local smoothness in predictions. This approach reduces the total number of simulated paths from $n \times N_{bm}$ to $n_v \times N_{bm}$, where $n_v$ is the number of charts and $n_v \ll n$. As a result, RC-AGP improves both computational efficiency and regression accuracy. For comparison, we also developed the Sparse Atlas GP (S-AGP), which integrates the Atlas BM framework with the subset of regressors approximation \citep{smola2000}. While this section focuses on RC-AGP, details on S-AGP can be found in Section G. of the Appendix. As demonstrated in subsequent regression experiments, the RC-AGP consistently outperforms all other comparision methods across most tasks. 

Let the atlas of $\M$ consist of $n_v$ charts,  with the associated data cloud $\S$ represented as the 
 union of $n_v$ subsets: $\S = \S_1 \cup \S_2 \cdots \cup \S_{n_v}$. 
We select the centre points (median points) from each $\S_i$ as the starting points for the BM trajectories  simulated by \eqref{eqn:swBM}. The transition density between these center points is then estimated using \eqref{eqn:kheat}. Given a diffusion time $t$, we have an $n_v
\times n_v$ covariance matrix, named $K_h^t$, the heat kernel matrix. Let $K_{rbf}$ be a $n \times n$ covariance matrix computed from a standard Euclidean RBF kernel. The RC-AGP can be constructed by first expanding the $n_v \times n_v$ matrix $K_h^t$ to $n \times n$ matrix $\tilde{K}_h^t$ with repeating elements, and then taking hadamard product with $K_{rbf}$.

For example, let $\M$ be represented by a point cloud consisting of $n$ points. Assume the point cloud can be partitioned into three overlapped subsets $\S_1$, $\S_2$ and $\S_3$. Three charts are learned from the subsets. By initiating BM paths from the center points of these subsets and simulating the process over a diffusion time $t$, we can derive a $3\times3$ matrix $K_h^t$ which is the heat kernel evaluated at the three center points of $\S_1$, $\S_2$ and $\S_3$:
\begin{align*}
\label{eqn:tilK}
K_{h_u}^t= \left[\begin{array}{ccc}  
h_{11} &h_{12} & h_{13} \\[0.3em]
h_{21} &h_{22} & h_{23} \\[0.3em]
h_{31} & h_{32} & h_{33} \\[0.3em]
\end{array} \right] ,       
\end{align*}
where $h_{ij}$ represents the heat kernel estimated at point $i$ and $j$ by \eqref{eqn:kheat}. Let $n_{S_i}$ be the number of points in $\S_i$. We can expand $K_h$ and create a $n \times n$ matrix $\tilde{K}_h$ by repeating elements.
\begin{align}
\tilde{K}_{h}^t= \left[\begin{array}{ccc}  
h_{11}* \bf{1}_{S1} &h_{12}*\bf{1}_{S12} &h_{13}*\bf{1}_{S13} \\[0.3em]
 h_{21}*\bf{1}_{S21} & h_{22}*\bf{1}_{S2} & h_{23}*\bf{1}_{S23} \\[0.3em]
  h_{31}*\bf{1}_{S31} & h_{32}*\bf{1}_{S32}  & h_{33}*\bf{1}_{S3}  \\[0.3em]
\end{array} \right]       
\end{align}
$\bf{1}_{S_i}$ represent a $n_{S_i} \times n_{S_i}$ matrix of ones. $\bf{1}_{S_{ij}}$ is a $n_{S_i} \times n_{S_j}$ matrix of ones. The Riemannian corrected kernel matrix can be calculated as:
\begin{align} \label{eqn:RCkernel}
K_{RC}= K_{rbf} \odot \tilde{K}_h^t,
\end{align}
where $K_{rbf}$ is an $n \times n$ matrix created from the RBF kernel. $\odot$ stand for the Hadamard product. Given Theorem \ref{th:positive}, $K_{RC}$ is semi-positive definite and can be used to construct GP on manifolds. The proof of Theorem \ref{th:positive} is given in Section A.1 of the Appendix. The Riemannian corrected kernel is also differentiable. This framework is very flexible. The RBF kernel can be replaced by other valid kernels such as the Mat\'ern class. 

\begin{theorem} \label{th:positive}
The Riemannian corrected kernel matrix defined in equation \eqref{eqn:RCkernel} is semi-positive definite. 
\end{theorem}


Under an RC-AGP prior for the unknown regression function as $f \sim GP(0,K_{RC}(.,.) )$, 
\begin{align*}
p( \mathbf{f} | s_1, s_2, ..., s_n) \sim \mathcal{N} (0, \Sigma_{\f \f}),
\end{align*}
where $\mathbf{f}$ is the discretisation of $f$ over the labeled points $s_1,s_2,\cdot\cdot\cdot,s_{n_d}$ so that $f_i = f(s_i)$. It is straightforward to identify which subsets the labeled points belong to and therefore construct the corresponding $n_d \times n_d$ $\tilde{K}^t_h$ as in \eqref{eqn:tilK}. The covariance matrix $\Sigma_{\f \f}$, also of size $n_d \times n_d$,  is computed using the Riemannian corrected kernel from \eqref{eqn:RCkernel}. The $(i,j)$-th entry of $\Sigma_{\f\f}$ is given by $\Sigma_{\f\f _{i,j}} = \sigma_r^2 K_{RC}(s_i,s_j)$. There are three hyper-parameters in this construction -- the diffusion time $t$ of $K_h$, the length-scale $l$ of $K_{rbf}$ and the rescaling parameter $\sigma_r$. Optimisation of the hyper-parameters can be done by maximising the log marginal likelihood in \eqref{eq:loglike}: 
\begin{align}
\label{eq:loglike}
\log p( \bm{y}| s) &= \log \int p( \bm{y}| {\text {\bf f} }) p({\text {\bf f} }|s) d{\text {\bf f} }   \nonumber \\
&= -\frac{1}{2} \bm y^T (\Sigma_{ {\text {\bf ff}} } + \sigma_{noise}^2 I)^{-1} \bm y - \frac{1}{2} \log|\Sigma_{ {\text {\bf ff}} } +\sigma_{noise}^2I | - \frac{n}{2}\log2\pi.
\end{align}

Let $\bf{f}_*$ denote the function values at the unlabelled test points.  $\Sigma_{\f_* \f_*}$ and  $\Sigma_{\f \f_*}$ can also be derived from the Riemannian corrected kernel by checking which subset $x_*$ belongs to. The predictive distribution is
\begin{align}
\label{eq:rcPredict}
p( \mathbf{f}_* | \bm{y}) = \mathcal{N} ( \Sigma_{\f_* \f} ( \Sigma_{\f \f} + \sigma_{noise}^2 I   )^{-1} \bm{y} , \Sigma_{\f_* \f_*} - \Sigma_{\f_* \f}( \Sigma_{\f \f} + \sigma_{noise}^2 I   )^{-1} \Sigma_{\f \f_*} ).
\end{align}

\section{Applications} \label{simdata}

\subsection{Torus}

\begin{figure}[h!]
\includegraphics[width=0.4\textwidth,height=0.3\textwidth]{ 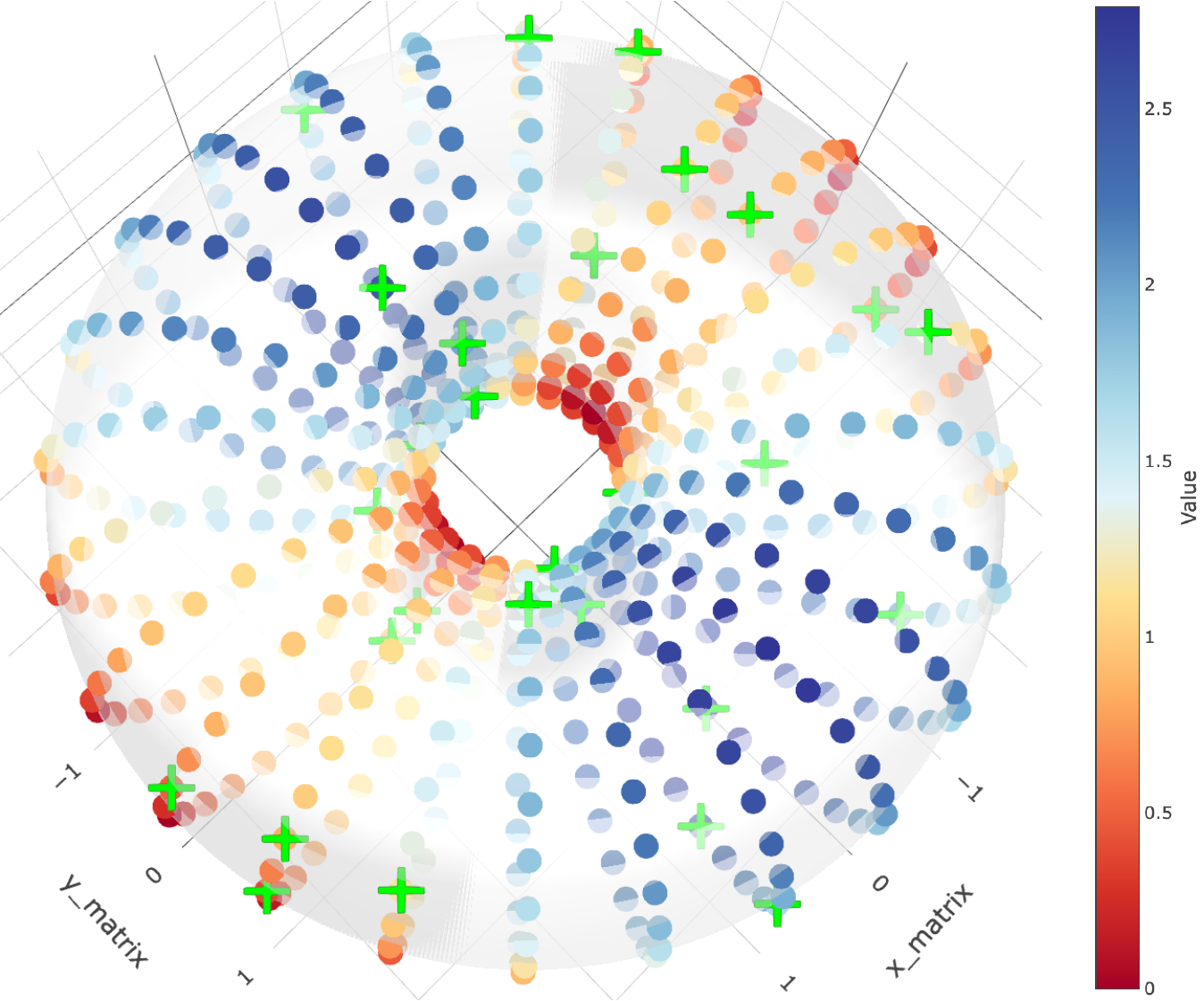 } 
\centering\caption{ \label{fig:Torus3d} The color of the points indicates the value of the true function on the torus point cloud in $\R^3$, while the green crosses represent the training observations.} 
\end{figure}
In this section, we conduct a simulation study for a regression model with synthetic data on a Torus, a two-dimensional manifold depicted by a point cloud in $\R^3$. The point cloud and the true function values, indicated by color codes, are plotted in Fig.~\ref{fig:Torus3d}. The point cloud comprises $n = 625$ points, with $n_d= 30$ labeled points marked as green crosses in Fig.~\ref{fig:Torus3d}. The analytical chart is parameterised by the angles of the outer and inner circles of the torus. The regression function varies according to these angles. However, in the context of the point cloud, these angles are unknown. Only the three-dimensional coordinates of the point cloud and the function values at the labeled points are observed. To learn the atlas, the torus cloud is divided into  overlapping subsets, as shown in Fig.~\ref{fig:Toruslatent}. GPLVM is employed to learn the charts for each subset. The union of these learned charts constitutes the atlas of the Torus point cloud in Fig.~\ref{fig:Toruslatent} C. 

\begin{figure}[h!]
    \centering
    \subfigure[True function on Torus 2D]{ \label{fig:Tpoint}  \includegraphics[width=0.4\textwidth,height=0.3\textwidth]{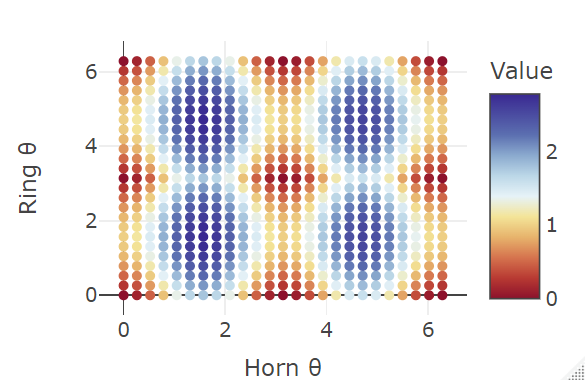}} 
    \subfigure[RC-AGP prediction]{ \label{fig:TrcGP}  \includegraphics[width=0.4\textwidth,height=0.3\textwidth]{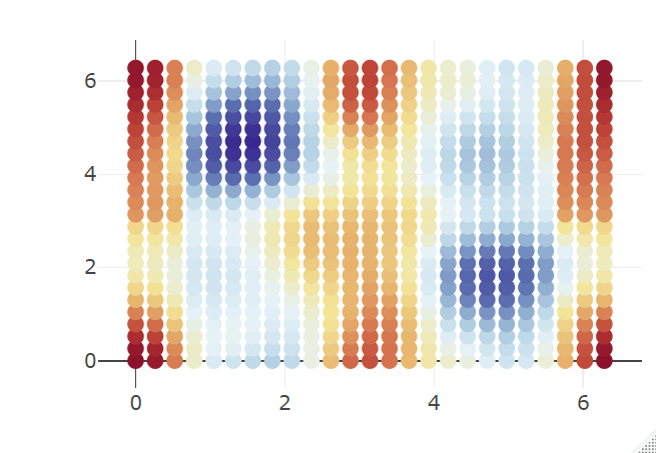}}
 \subfigure[Euclidean GP prediction]{ \label{fig:Trbf}   \includegraphics[width=0.4\textwidth,height=0.3\textwidth]{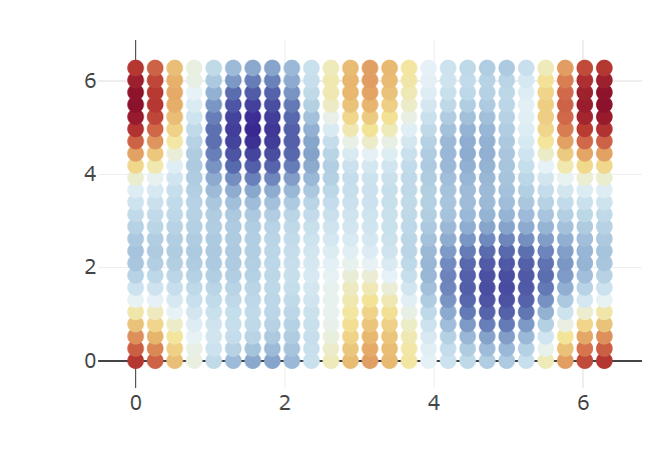}}   
  \subfigure[GL-GP prediction]{ \label{fig:Tglf}   \includegraphics[width=0.4\textwidth,height=0.3\textwidth]{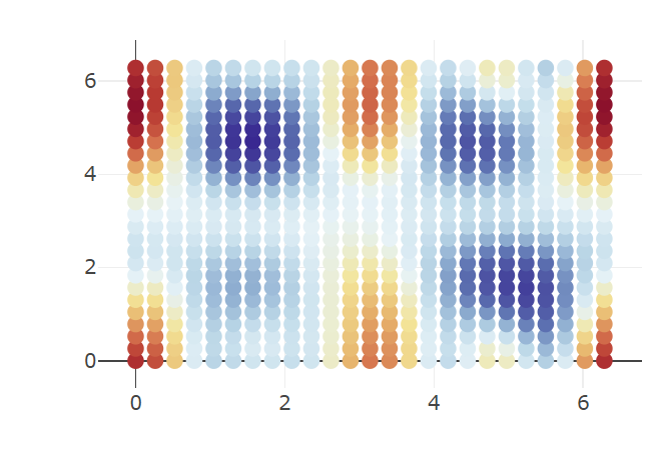}}   
 \caption{\label{fig:Torus}
    \footnotesize
    {\bf Comparison of different methods on Torus: (a) true function in the point cloud; The red to blue color range represents the true function values at the grid points. (b) RC-AGP prediction at the grid points; (c) Euclidean GP prediction using RBF kernel; (d) GL-GP prediction.}  }
\end{figure}

For clearer visualisation, we also plot the true function on the torus in two dimensions, as illustrated in Fig.~\ref{fig:Tpoint}, where the coordinates correspond to the outer and inner circular angles of the Torus. We first applied the standard Euclidean $\R^3$ GP (following \cite{Rasmussen2006}), utilising an RBF kernel based on $\R^3$ Euclidean distance. This method neglects the torus's intrinsic geometry. The predictive means at the unlabeled points are shown in the 2D Torus coordinates in Fig.~\ref{fig:Trbf}. In the center of Fig.~\ref{fig:Trbf}, the prediction appears predominantly blue, which is completely different from the true function, where the centre is dark red. A similar discrepancy is observed in the predictive mean of the GL-GP in Fig.~\ref{fig:Tglf}. Both the GL-GP and Euclidean GP fail to incorporate the manifold structure during regression. Conversely, the RC-AGP predictive mean, depicted in Fig.~\ref{fig:TrcGP}, closely aligns with the true function. The color distribution in \ref{fig:TrcGP} is consistent with that in Fig.~\ref{fig:Tpoint}. The S-AGP results are shown in Section H. of the Appendix. It outperforms those of the GL-GP and Euclidean GP but does not match the accuracy of the RC-AGP. We further evaluated all methods across a range of signal-to-noise ratios(SNRs) by perturbing the true function with Gaussian noise at levels of 10 dB, 20 dB, and 30 dB, performing 10 replicates for each noise level. Each method was subsequently applied to estimate the test function at the grid points for each replicate. The mean and standard deviation of the root mean squared error(RMSE) across the replicates at different noise levels are summarised in Table \ref{tab:TorusMean}. Values in brackets are the standard deviation. The RC-AGP is significantly better than all other methods under all conditions.
\begin{table}
\caption{\label{tab:TorusMean}Comparison of the root mean squared error of
predictive means for various methods on Torus point cloud.}
  \centering
\fbox{%
\begin{tabular}{*{10}{c}}
SNR&\em Euclidean $\R^3$ GP &\em GL-GP&\em RC-AGP&\em  S-AGP\\
\hline
30db&0.42(0.09)& 0.42(0.04)&0.33(0.06)&0.41(0.08)\\
20db&0.43(0.10)& 0.42(0.04)&0.34(0.05)&0.44(0.13)\\
10db&0.48(0.09)& 0.44(0.04)&0.40(0.05)&0.48(0.12)\\
\end{tabular}}
\end{table}

\subsection{U-shape point cloud}
\label{sec:u}

\begin{figure}[h!]
    \centering
    \subfigure[True function in U-shape]{ \label{fig:UtrFun}  \includegraphics[width=0.44\textwidth,height=0.37\textwidth]{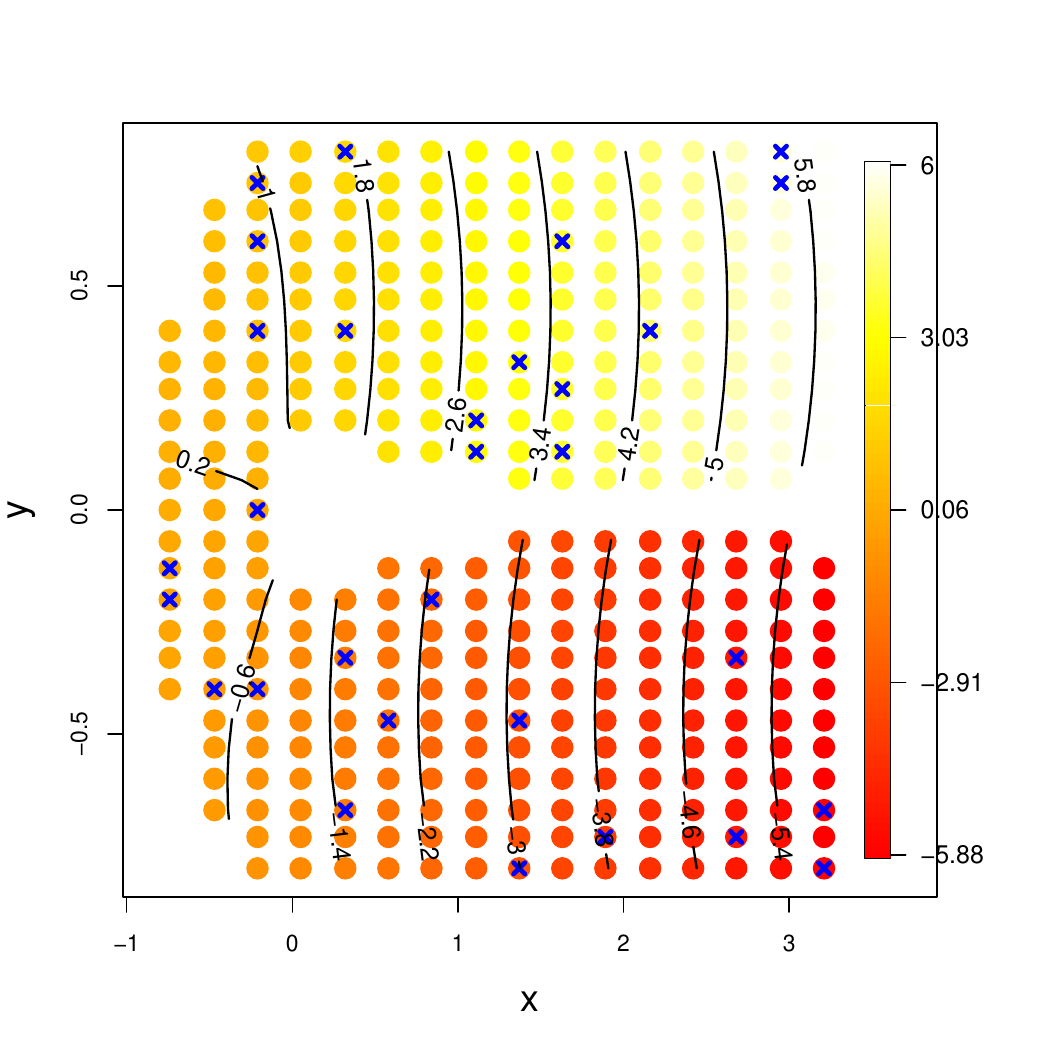}}
    \subfigure[RC-AGP prediction]{ \label{fig:URCGP}  \includegraphics[width=0.44\textwidth,height=0.37\textwidth]{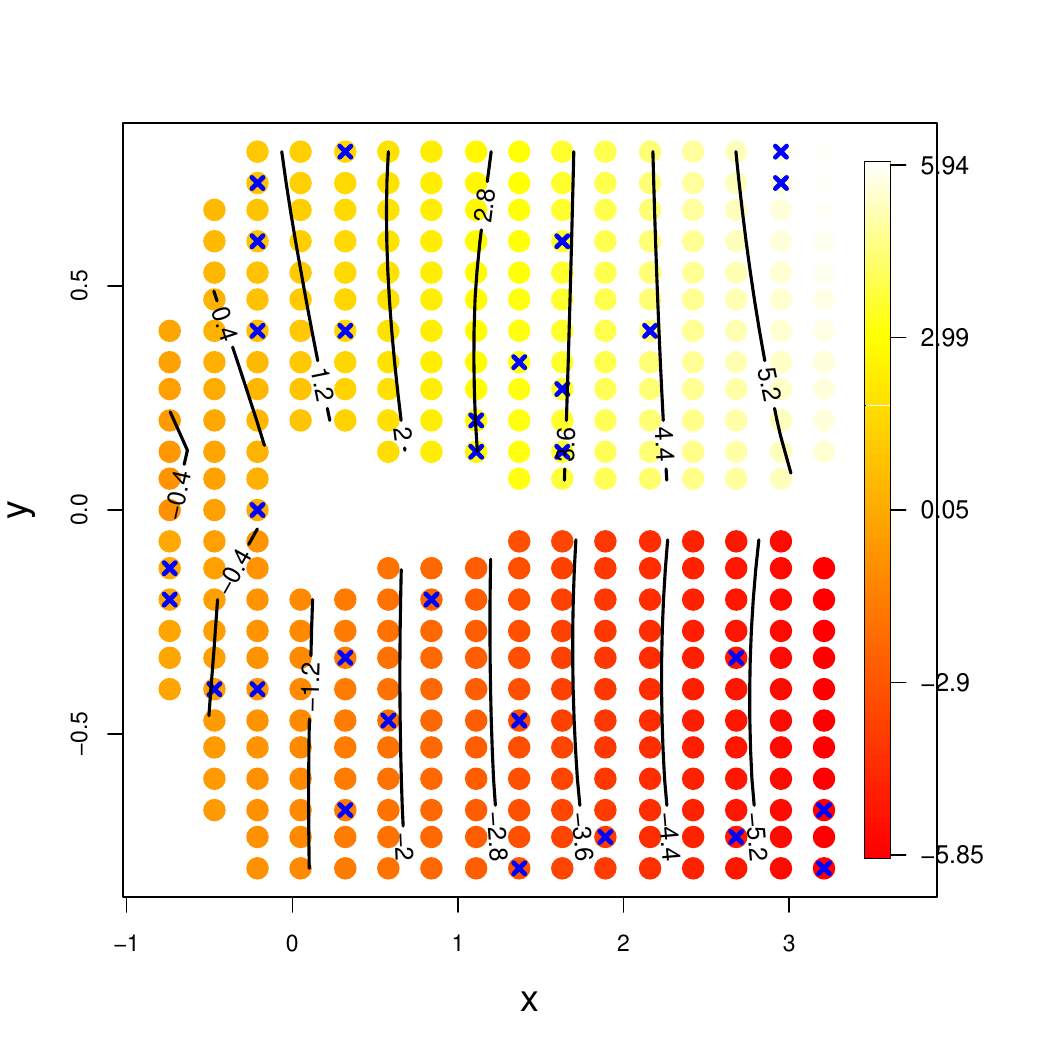}}
 \subfigure[Euclidean GP prediction]{ \label{fig:Urbf}   \includegraphics[width=0.44\textwidth,height=0.37\textwidth]{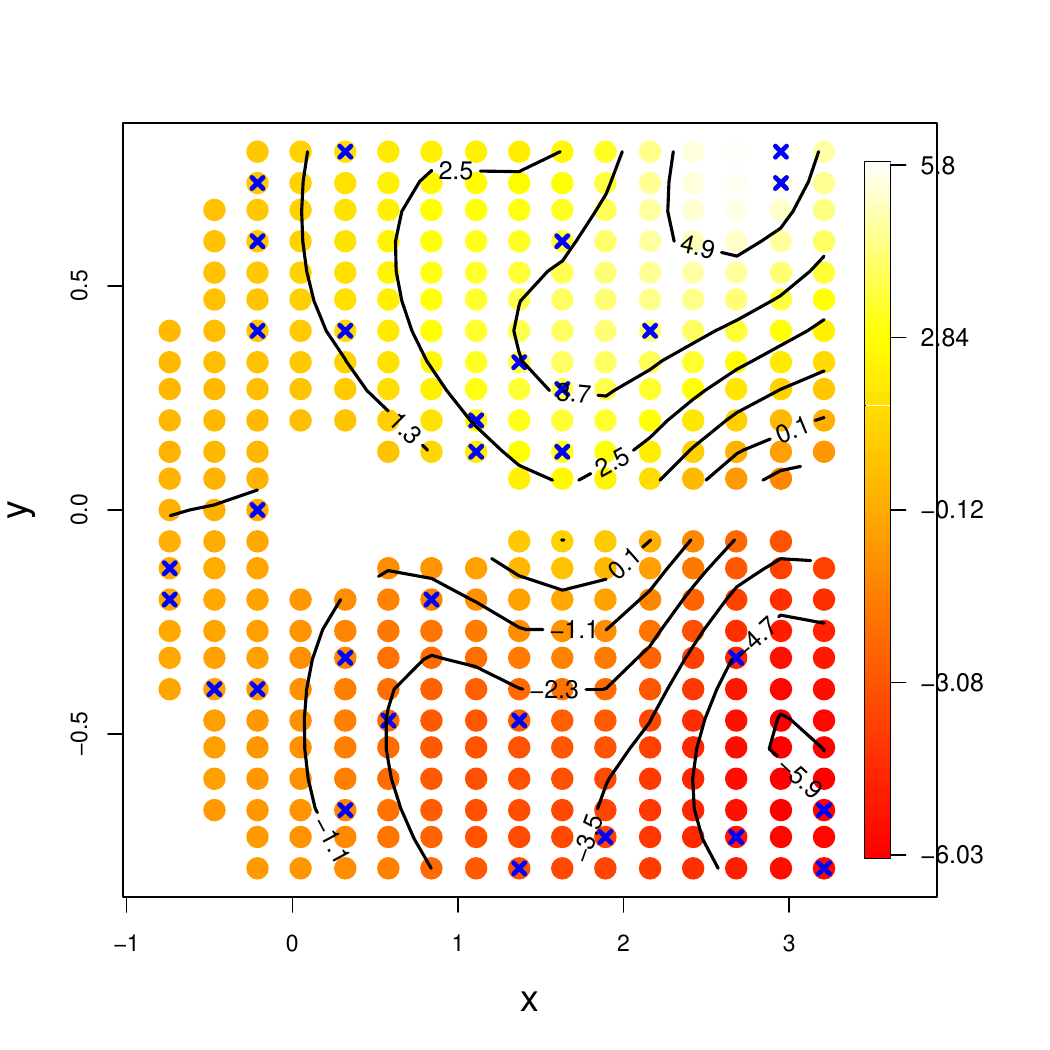}}   
  \subfigure[GL-GP prediction]{ \label{fig:UGLGP}   \includegraphics[width=0.44\textwidth,height=0.37\textwidth]{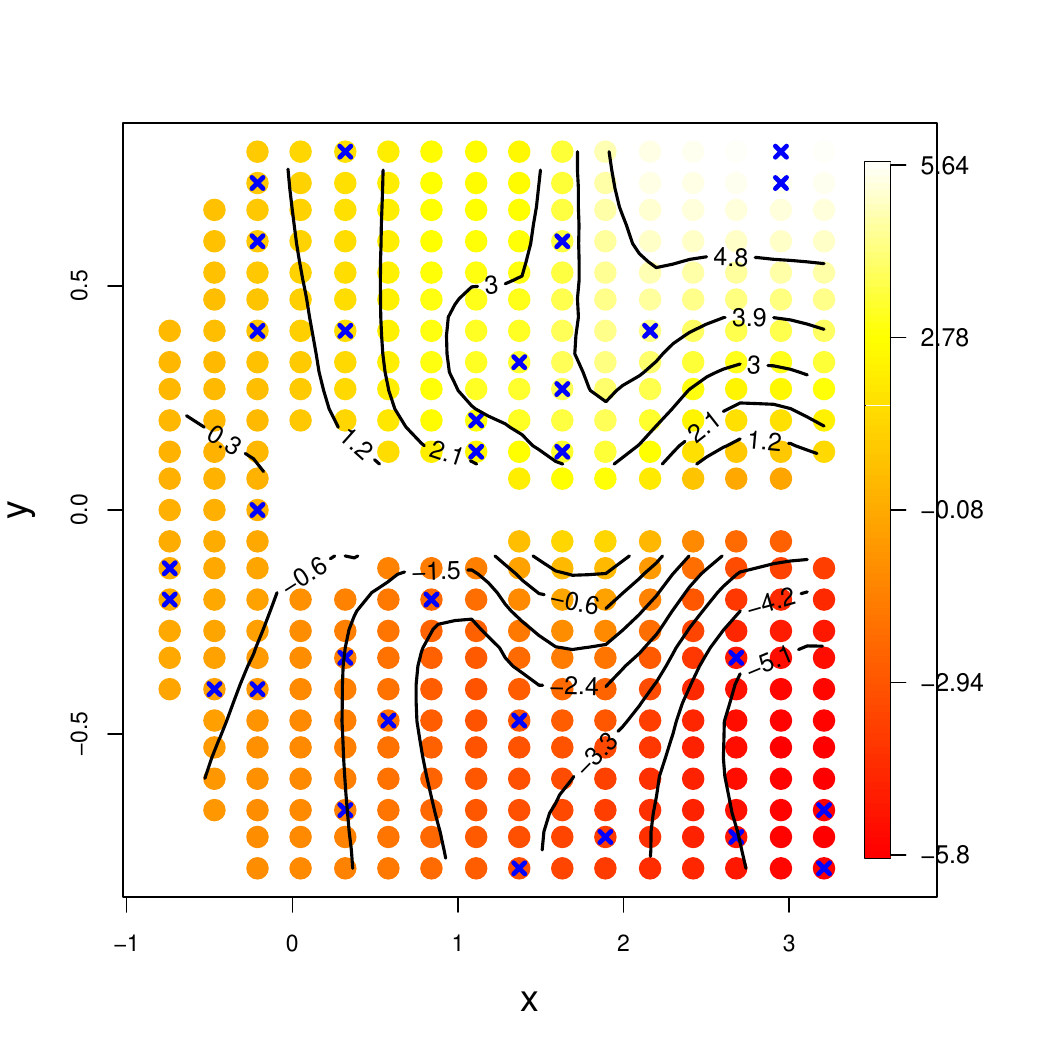}}  
 \caption{\label{fig:udomain}
    \footnotesize
    {\bf Comparison of different methods in U-shape: (a) True function in U-shape(blue crosses represent the training observations); (b) RC-AGP prediction at the grid points; (c) Euclidean GP prediction using RBF kernel; (d) GL-GP prediction. }  }
\end{figure}

We defined a U-shape point cloud as shown in Fig.~\ref{fig:UtrFun}. A smooth regression function is applied to the points \citep{wood}. The values (i.e., the color of the points) vary smoothly from the lower right-hand corner to the upper right-hand corner of the domain, ranging from $-$6 to 6. The blue crosses represent 30 observations randomly located within the domain of interest. The boundaries of the domain are unknown. The goal is to estimate the regression function and make predictions at equally spaced grid points within the domain. There are 355 grid points in the U-shaped domain. All grid points are divided into four subsets, as shown in Fig.~8 in Appendix. Since the U-shaped point cloud is a subset of $\R^2$, each chart is represented as a domain of $\R^2$ that covers the subset. The BM within a chart becomes the standard BM in $\R^2$. The change of coordinates operation is also straightforward. The BM steps can be transferred to a different chart in the overlapping region of subsets. 

In this study, we compared the RC-AGP, Euclidean $\R^2$ GP, GL-GP, and S-AGP at various levels of signal-to-noise ratio. The values of the true function were perturbed by Gaussian noise (with SNRs of 10db, 20db and 30db, respectively), with 10 replicates for each noise level. For each replicate, the different methods were applied to estimate the test function at the grid points. As a demonstration in the high SNR case, the predictive mean of the RC-AGP at the grid points is shown in Fig.~\ref{fig:URCGP}. The colored contours of the prediction closely resemble those of the true function in Fig.~\ref{fig:UtrFun}. In contrast, the contours of the Euclidean $\R^2$GP predictive mean in Fig.~\ref{fig:Urbf} smooth over the gap between the upper and lower arms of the point cloud, as these two arms are close in Euclidean distance. The $\R^2$GP using the RBF kernel assigns strong covariance between these two regions. However, the BM simulation starting from the lower arm needs to traverse different charts to reach the upper arm. This results in lower covariance between these two regions and more accurate predictions by the RC-AGP. This example was also used to evaluate the performance of GL-GP, as shown in Fig.~\ref{fig:UGLGP}. The result of GL-GP is similar to the $\R^2$GP.

The mean and standard deviation of the root mean-squared error for the replicates at different noise levels are reported in Table \ref{tab:U}. The RC-AGP is significantly better than all other methods in almost all cases. When the SNR level is low (10db), the  difference between RC-AGP and S-AGP is not significant. More results of S-AGP is shown in Section H. of Supplementary materiel.


\begin{table}
\caption{\label{tab:U}Comparison of the root-mean-squared error of
predictive means for various methods on U shape point cloud}
  \centering
\fbox{%
\begin{tabular}{*{10}{c}}
\em SNR &\em Euclidean $\R^2$ GP &\em GL-GP&\em RC-AGP&\em S-AGP\\
\hline
30db&1.73(0.25)&1.58(0.20)&0.168(0.05)&0.282(0.02)\\
20db&1.73(0.26)&1.61(0.22)&0.265(0.05)&0.312(0.04)\\
10db&1.83(0.30)&1.78(0.47)&0.682(0.13)&0.630(0.13)\\
\end{tabular}}
\end{table}

\subsection{Shark prey LEGO image cloud}
\begin{figure}[h!]
\includegraphics[width=0.7\textwidth,height=0.28\textwidth]{ 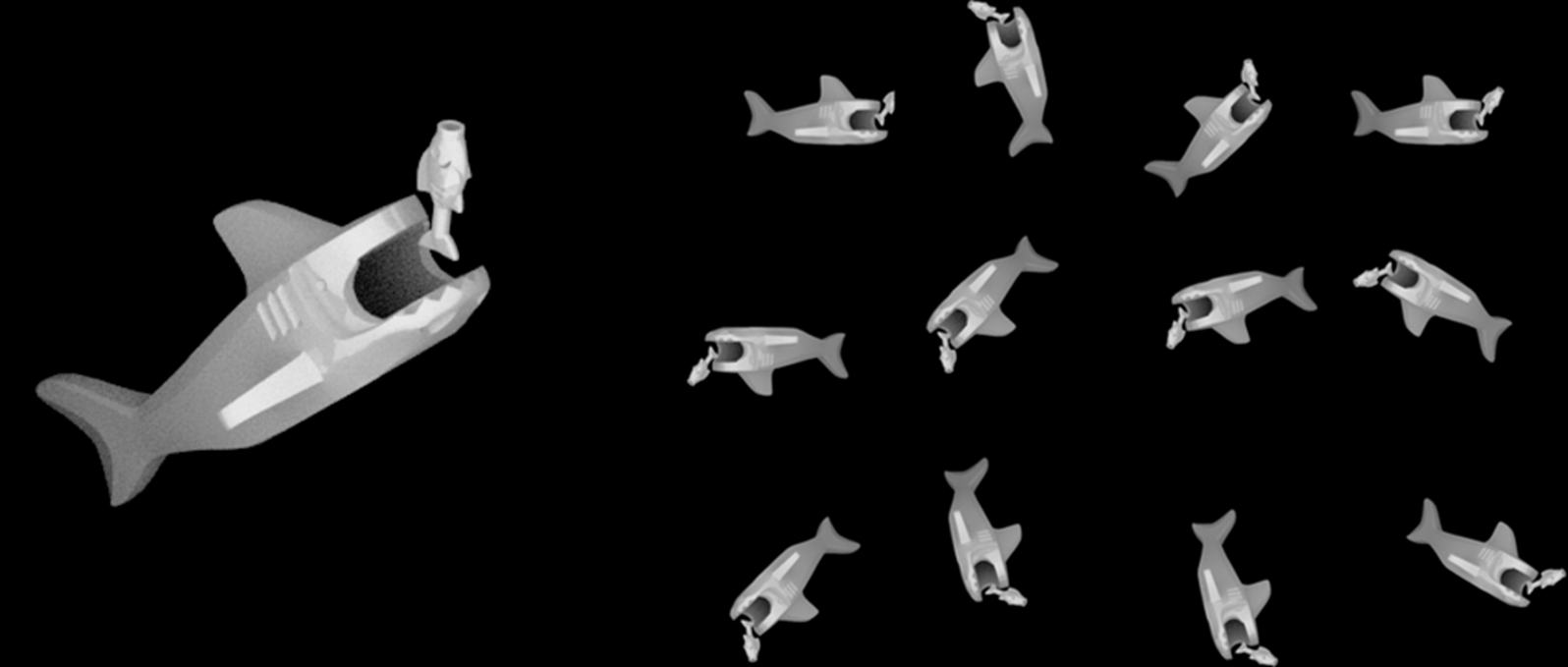} 
\centering\caption{ \label{fig:shark}   { Shark Prey image examples.}  }
\end{figure}

We analyze a dataset comprising 576 images depicting a dynamic scenario involving a LEGO shark and its prey\citep{watkins2012getting}. In this sequence, the small LEGO fish rotates around the shark while both entities simultaneously follow a circular trajectory within the background. The system's movement resembles a double rotation, analogous to the Earth's rotation and revolution. The regression function is defined as a function of the angles of revolution and rotation. The dimension of the original image space is $p = 900$. 12 image examples are given in Fig.~\ref{fig:shark}. Despite the high dimensionality, the movements of the LEGO shark prey occur on low-dimensional manifolds. Eight subsets are generated from the original set of 576 images. For each subset, we estimated a latent space with $q = 2$ dimensions using GPLVM. An example of the resulting latent space is shown in Fig.~9 of the Appendix, where the dark regions indicate areas of high uncertainty. The atlas for the LEGO images is composed of these latent spaces.

\begin{figure}[h!]
    \centering
    \subfigure[True function on image cloud in 3D]{ \label{fig:SP-3d}  \includegraphics[width=0.4\textwidth,height=0.3\textwidth]{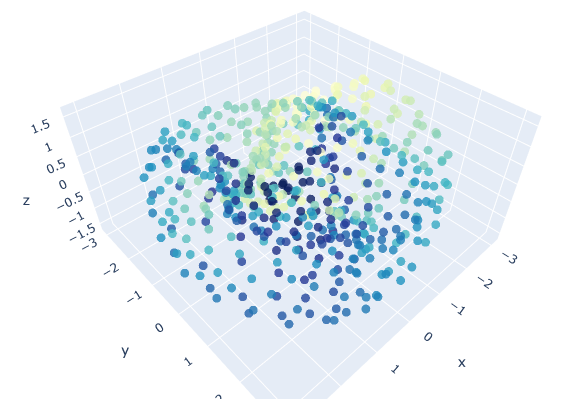}}  
     \subfigure[RC-AGP prediction]{ \label{fig:SP-RCGP}   \includegraphics[width=0.4\textwidth,height=0.3\textwidth]{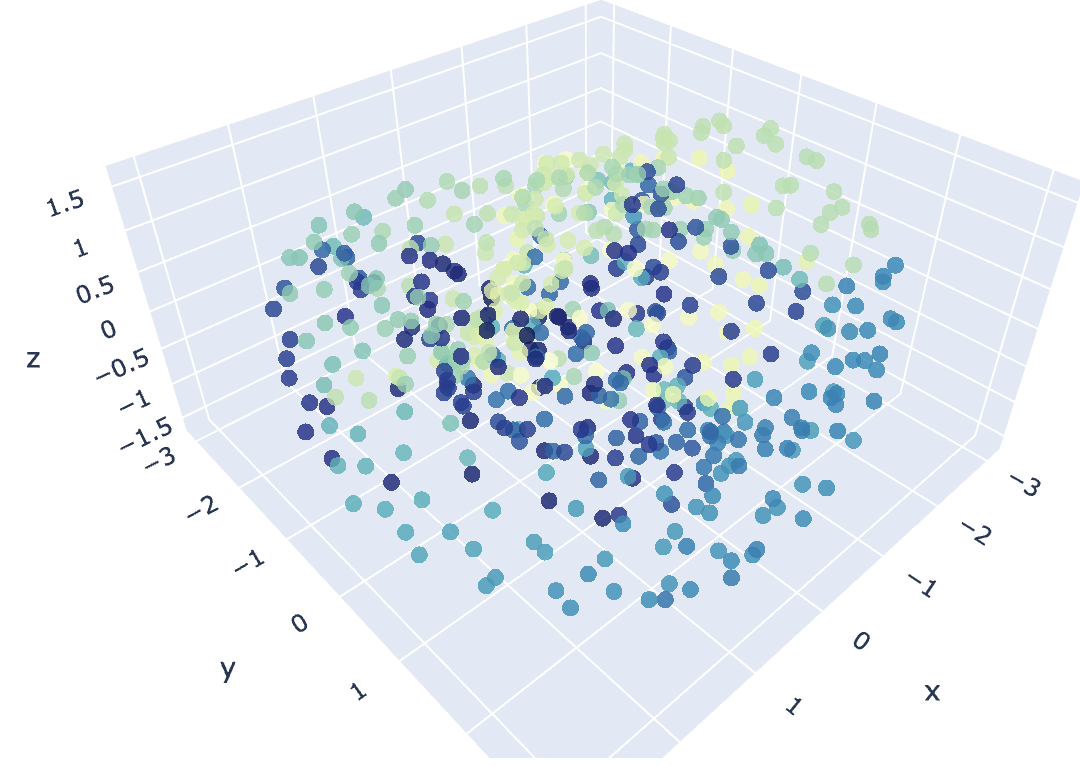}} 
     \subfigure[Euclidean $\R^{900}$ GP prediction]{ \label{fig:SP-Rbf}   \includegraphics[width=0.4\textwidth,height=0.3\textwidth]{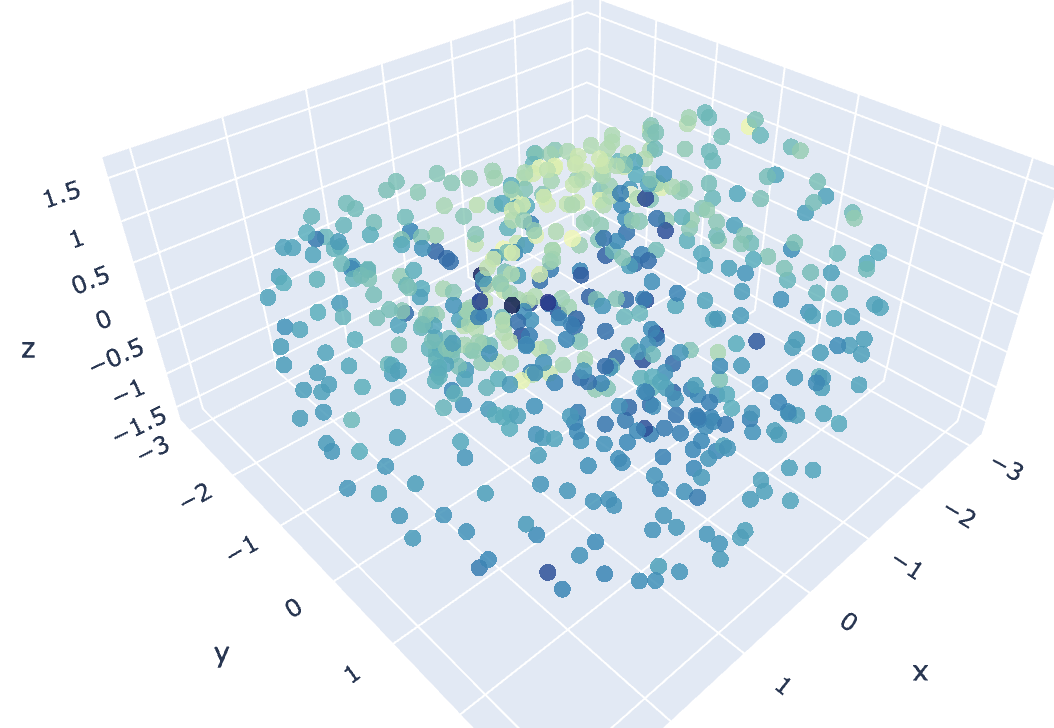}} 
  \subfigure[GL-GP prediction]{ \label{fig:SP-GLGP}   \includegraphics[width=0.4\textwidth,height=0.3\textwidth]{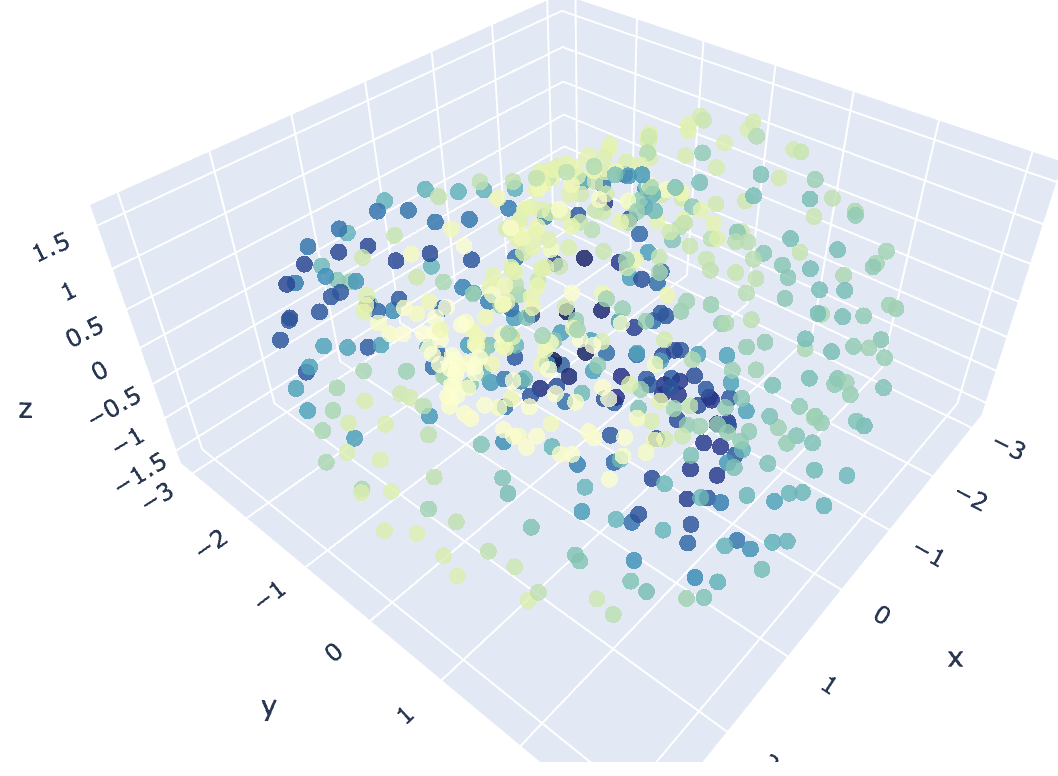}}  
 \caption{\label{fig:SP}
    \footnotesize
    {\bf Comparison of different methods in Shark prey LEGO image cloud: (a) The true function values are depicted on the 3D points, with colors ranging from blue to green to represent varying magnitudes; (b) RC-AGP prediction; (c) Euclidean $\R^{900}$ GP prediction; (d) GL-GP prediction.  }  }
\end{figure}

In this study, a set of $n_d=35$ images are randomly selected and used as the training set (labeled data). The remaining 541 images are used as the test set (unlabeled data). Different methods have been applied to estimate the regression function. Given the high dimensionality of the image space, visualising the 900-dimensional point cloud is impractical. To facilitate visualisation, we  plotted all 576 images in a three-dimensional space using PCA, as shown in Fig.~\ref{fig:SP-3d}. The color density represents the regression function values at each point. The predictive mean of the RC-AGP is illustrated in Fig.~\ref{fig:SP-RCGP}, displaying a pattern closely aligned with the true function depicted in Fig.~\ref{fig:SP-3d}. In contrast, the predictive means of the GL-GP, shown in Fig.~\ref{fig:SP-GLGP}, reveal a notable discrepancy: the central region of the cloud is dominated by bright yellow and green hues, whereas the true function indicates dark blue in the same area. The predictive means of the $\R^{900}$ GP, utilising the RBF kernel with $\R^{900}$ Euclidean distance, are presented in Fig.~\ref{fig:SP-Rbf}. The results display a color distribution that deviates from the true function. We also created ten different training and test sets through random selection. The mean and standard deviation of the root mean square errors calculated over the ten test sets are shown in Table \ref{tab:Shark}. The RC-AGP achieves the smallest mean RMSE. It is significantly better than all other methods (Euclidean GP, GL-GP and S-AGP). The plot of S-AGP results is shown in Section H. of the Appendix.

\begin{table}
\caption{\label{tab:Shark}Comparison of the root-mean-squared error of
predictive means for various methods on Shark LEGO point cloud}
  \centering
\fbox{%
\begin{tabular}{*{10}{c}}
\em Euclidean $\R^{900}$ GP &\em GL-GP&\em RC-AGP&\em S-AGP\\
\hline
1.09(0.06)& 0.98(0.17)& 0.62 (0.04)&0.81(0.07)\\
\end{tabular}}
\end{table}

\subsection{Aral sea point cloud}

\begin{figure}[h!]
    \centering
    \subfigure[Truth in Aral sea]{ \label{fig:Apoint}  \includegraphics[width=0.44\textwidth,height=0.35\textwidth]{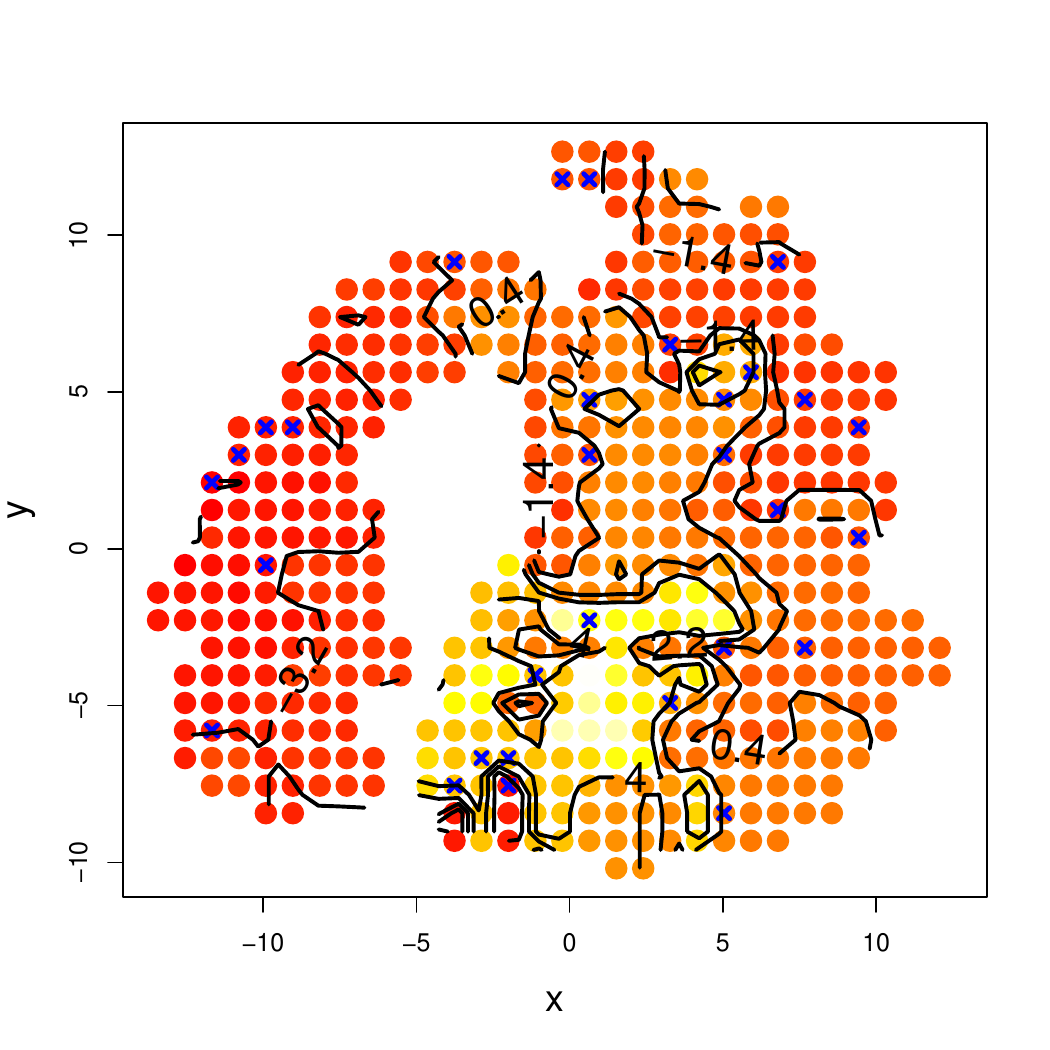}}
    \subfigure[RC-AGP prediction]{ \label{fig:ARCGP}  \includegraphics[width=0.44\textwidth,height=0.35\textwidth]{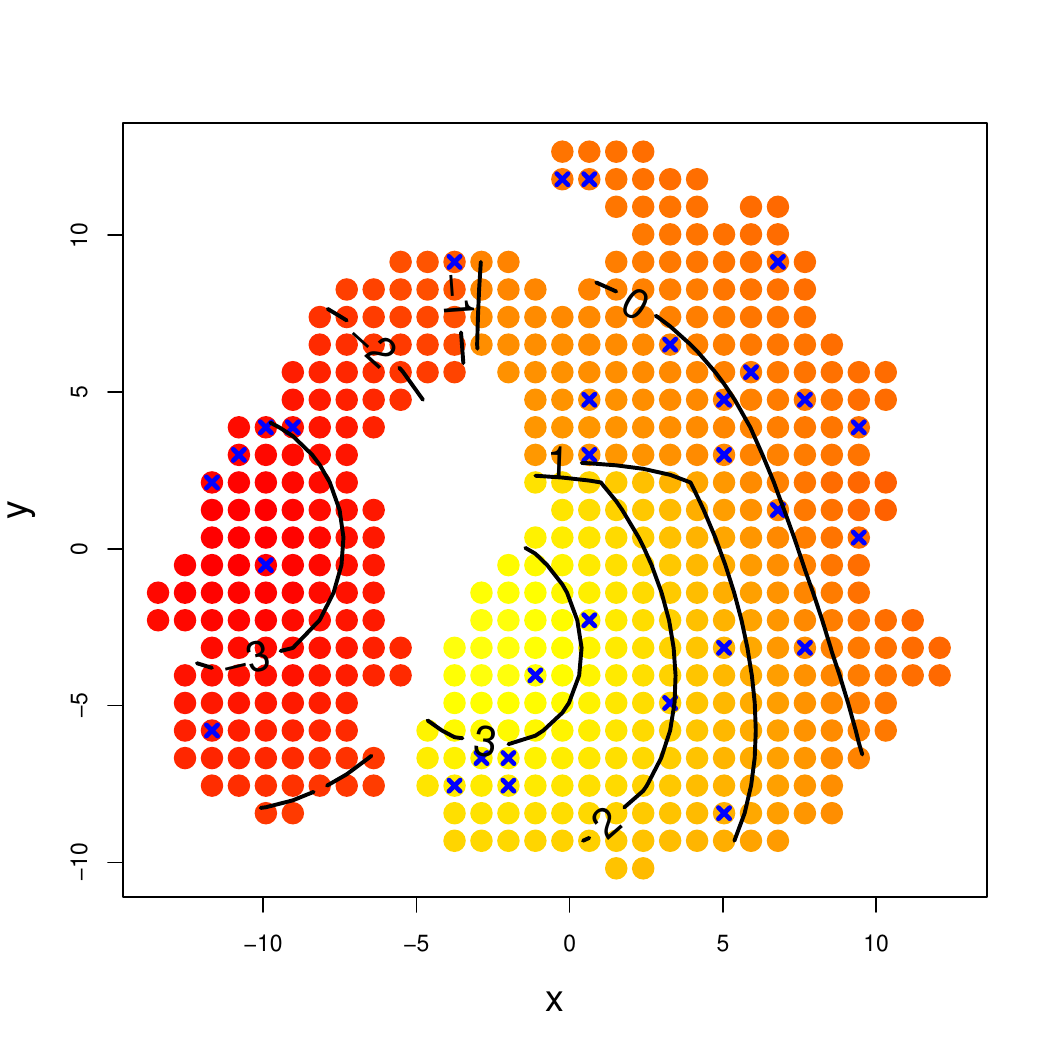}}
 \subfigure[Euclidean GP prediction]{ \label{fig:Arbf}   \includegraphics[width=0.44\textwidth,height=0.35\textwidth]{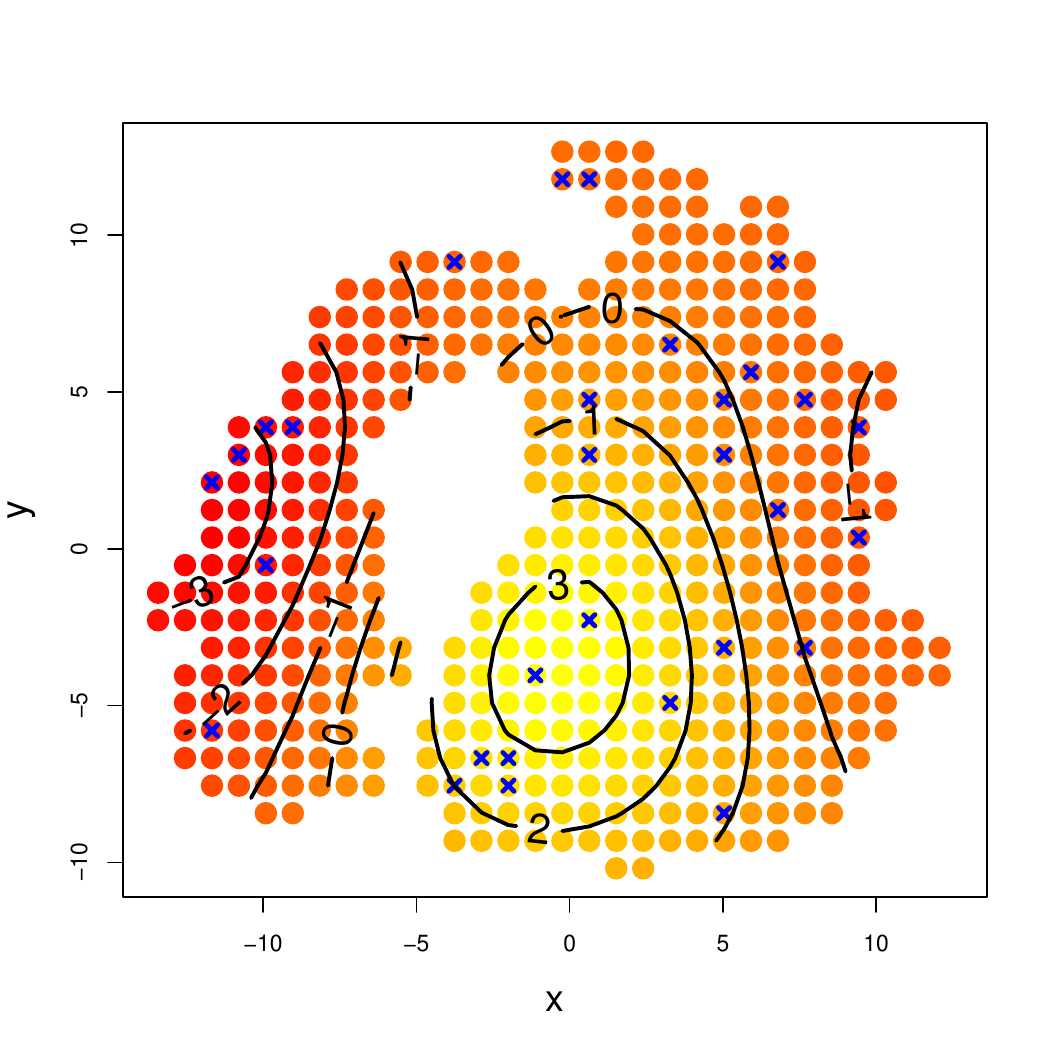}}   
  \subfigure[GL-GP prediction]{ \label{fig:AGLGP}   \includegraphics[width=0.44\textwidth,height=0.35\textwidth]{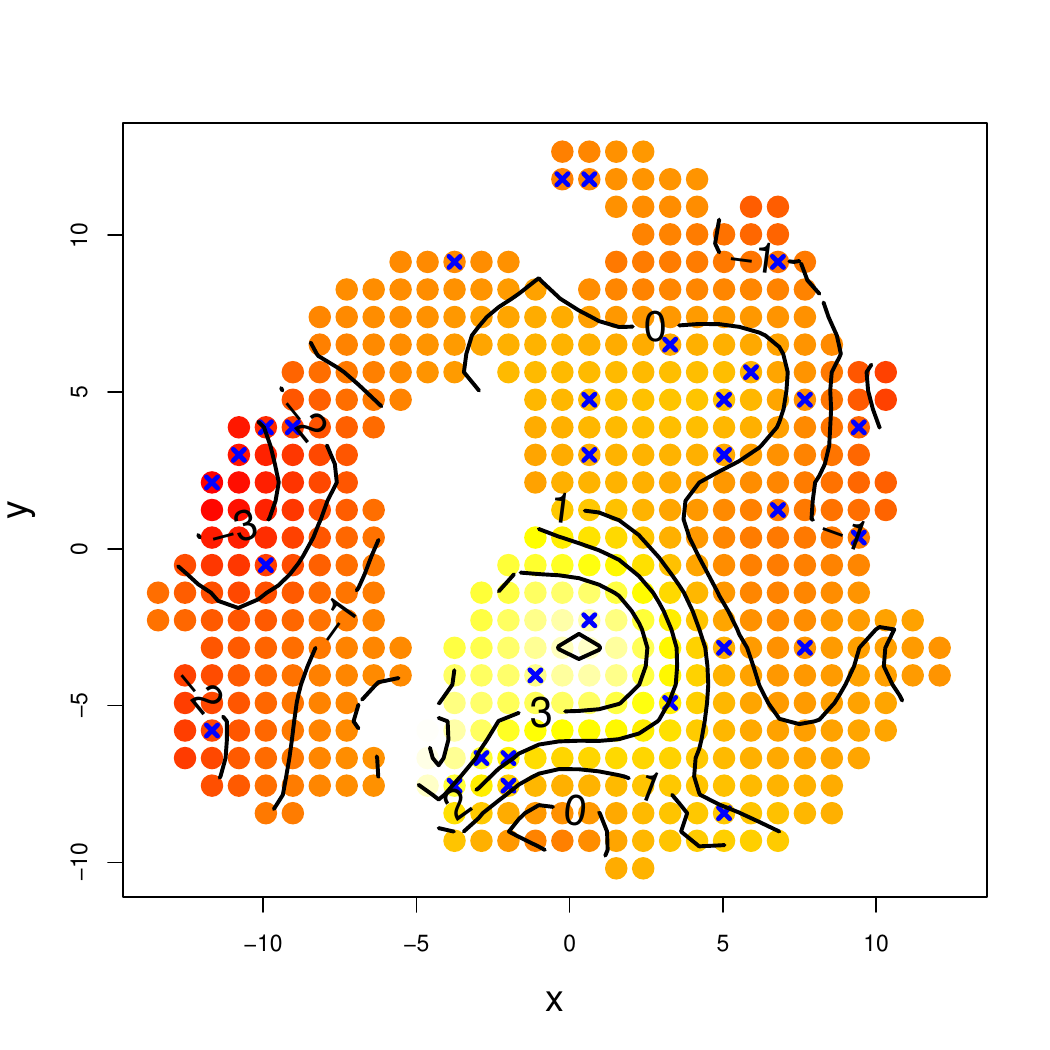}}  
 \caption{\label{fig:Aralsea}
    \footnotesize
    {\bf Comparison of different methods in Aral sea: (a) True function in Aral sea(blue crosses represent the training observations); (b) RC-AGP prediction at the grid points; (c) Euclidean GP prediction using RBF kernel; (d) GL-GP prediction.}  }  
\end{figure}

The chlorophyll concentration at 485 locations in the Aral sea is plotted in Fig.~\ref{fig:Apoint} \citep{wood2006}.  The concentration level is indicated by the intensity of the color. The remote sensing data is noisy and not as smooth as the U-shape example. Although the east and west part of the Aral sea is very close in the Euclidean distance, they are separated by the isthmus of the peninsula. The west part has a lower concentration level compared to the east part.  The blue crosses represent 30 observations randomly located within the Aral sea. We applied different methods to estimate the spatial pattern of the chlorophyll density. The scaled chlorophyll concentration level is modeled as a function of the location, such as the latitude and longitude. 

All grid points are divided into three subsets, as shown in Fig.~12 of the Appendix. The predictive mean of the RC-AGP at the grid points is displayed in Fig.~\ref{fig:ARCGP}. The colored contours indicate that the prediction does not smooth across the isthmus of the peninsula. The prediction in the western part of the Aral Sea is lower and remains unaffected by the observations in the eastern part. The overall pattern closely resembles the data in Fig.~\ref{fig:Apoint}. In contrast, the contours of the Euclidean GP predictive mean in Fig.~\ref{fig:Arbf} smooth over the gap. The Euclidean GP with an RBF kernel assigns a high covariance between these regions. A similar effect can be observed in the GL-GP prediction, as shown in Fig.~\ref{fig:AGLGP}. We randomly create 10 batches of training data. Each batch contains 30 training points. The mean and standard deviation of the RMSE for the 10 replicates are listed in Table \ref{tab:Aral}. The RC-AGP achieves the minimum RMSE. The difference between the Euclidean GP and GL-GP is not significant.

\begin{table}
\caption{\label{tab:Aral}Comparison of the root-mean-squared error of
predictive means for various methods on Aral sea point cloud}
  \centering
\fbox{%
\begin{tabular}{*{10}{c}}
\em Euclidean $\R^2$ GP &\em GL-GP&\em RC-AGP&\em S-AGP\\
\hline
2.65(0.42)& 2.49(0.32)& 2.31 (0.23)&2.33(0.16)\\
\end{tabular}}
\end{table}

\section{Discussion}
\appendix
In this paper, we introduce the Riemannian corrected Atlas Gaussian processes, a novel framework for modeling data on unknown manifolds with nontrivial topologies. Our key contributions are twofold: (1) we developed an efficient method for heat kernel estimation on unknown manifolds by constructing a probabilistic atlas and simulating BM paths on the atlas; (2) we introduced Riemannian-corrected Atlas GPs, which leverage the estimated heat kernel combined with an RBF kernel to enhance computational efficiency and accuracy in manifold-based inference. Our method outperforms single-chart BM and graph Laplacian approximations, requiring fewer data points for accurate kernel estimates and maintaining robustness to variations in point cloud distributions. We also proved the semi-positive definiteness of the Riemannian-corrected kernel. In regression tasks on datasets such as U-Shape, Aral Sea, Torus, and high-dimensional image clouds, RC-AGPs consistently outperformed Euclidean GPs and GL-GPs, demonstrating their effectiveness. Future work could extend this framework to dynamic and time-varying manifolds and explore adaptive atlas construction to handle large-scale data, further broadening the applicability of RC-AGPs.

\appendix
\section{Proof of theorem}
\subsection{Proof of Theorem 5.1 in the main paper}
The Riemannian corrected kernel matrix can be calculated as:
\begin{align} \label{eqn:RCkernel}
K_{RC}= K_{rbf} \odot \tilde{K}_h
\end{align}
where $K_{rbf}$ is a $n \times n$ matrix created from the RBF kernel. $\odot$ stand for the Hadamard product. $\tilde{K}_h$ is the expanded heat kernel matrix.
\begin{theorem*} \label{theorem:positive}
The Riemannian corrected kernel matrix defined in equation \eqref{eqn:RCkernel} is semi-positive definite. 
\end{theorem*}

\begin{proof}
$K_h$ is derived from the heat kernel of $\M$. Therefore $K_h$ is positive semi-definite. All eigenvalues of $K_h$ is bigger than zero. $\tilde{K}_h$ is derived by multiplying matrix of ones for each element of $K_h$. The eigenvalues of $\tilde{K}_h$ is bigger or equal to zero. $\tilde{K}_h$ is positive semi-definite. $K_{rbf}$ is constructed from the RBF kernel. $K_{rbf}$ is positive definite. $K_{RC}$ is constructed as the hadamard product of $K_{rbf}$ and $\tilde{K}_h$. Using the Schur product theorem, we can prove $K_{RC}$ is also positive sem-definite.
\end{proof}

\subsection{Proof of Theorem 4.1 in the main paper}
For each chart in the atlas, the BM can be defined as the SDE in the $k_{th}$ chart as 
\begin{align}
\label{eqn:swBM}
dx_k^i(t) = \frac{1}{2}{G_k}^{-1/2} \sum^{q}_{j=1}\frac{\partial}{\partial x_k^j} \left(  {\G_k}^{ij} {G_k}^{1/2} \right) dt + \left( {\G_k}^{-1/2} dB_k(t)\right)_i
\end{align}
where $x_k^i$ represents the coordinate in the $i_{th}$ dimension of chart $k$ (latent space $k$). $\G_k$ is the metric tensor,
$G_k$ is the determinant of $\G_k$ and $B_k(t)$ represents an independent BM in Euclidean space.

\begin{theorem*}
\label{th:sw}
If the region on $\M$ can be parameterised by multiple charts, the stochastic process defined in \eqref{eqn:swBM} is chart independent. With a given diffusion time, simulations in any choice of charts are equivalent to the same step in $\M$.
\end{theorem*}

Let $\M$ be a manifold of dimension $q$ with a Riemannian metric $g_{\mathbb{M}}$. $\varphi_1: \mathbb{U}_1\rightarrow \mathbb{M}$ and $\varphi_2: \mathbb{U}_2 \rightarrow \mathbb{M}$ are two local coordinate charts of $\M$ where $\mathbb{U}_1$ and $\mathbb{U}_2 $ are open subsets of $\R^q$. 
\begin{itemize}
\item $\psi={\varphi_2}^{-1} \circ \varphi_1: \mathbb{U}_1 \rightarrow {\mathbb{U}}_2$ is the change of coordinate.
$\psi^{-1}=\varphi_1^{-1} \circ {\varphi}_2: {\mathbb{U}_2} \rightarrow \mathbb{U}_1$ is the inverse change of coordinate. (assume $\varphi_1 \left( \mathbb{U}_1 \right) = {\varphi_2}\left( {\mathbb{U}_2} \right)$ without loss of generality.)
\item $\left(x^{1}, \ldots, x^{q}\right)$ denotes the standard coordinates in $\mathbb{U}_1 \in \R^q$.
$\left(\bar{x}^{1}, \ldots, \bar{x}^{q}\right)$ denotes the standard coordinates in ${\mathbb{U}_2} \in \R^q$.
\item $g_1$ denotes matrix representation of $g_{\mathbb{M}}$ in $\mathbb{U}_1$ (via $\varphi_1$) i.e. $g_{1,i j}=g_{\mathbb{M}}\left(\varphi_{1,*} \frac{\partial}{\partial x^{i}}, \varphi_{1,*} \frac{\partial}{\partial x^{j}}\right)$. $g_1^{ij}$ denotes the element of $g_1^{-1}$. $G_1 = det(g_1)$. Similarly for ${g}_2$ , ${g}_{2,ij}$, ${g}_2^{ij}$, ${G}_2$.
\item  $D \psi$  is the matrix derivative of  $\psi = \left(\psi^{1}\left(x^{1}, \ldots x^{q}\right), \ldots, \psi^{q}\left(x^{1}, \ldots, x^{q}\right)\right)$, i.e. $(D \psi)_{i}^{j}=\frac{\partial \psi^{j}}{\partial x^{i}} $. Using the chain rule  $\Rightarrow  D \psi^{-1} \cdot D \psi=I_{q}$. 
\item $\frac{\partial}{\partial x^{i}}=\sum_{j=1}^{q} \frac{\partial \psi^{j}}{\partial x^{i}} \frac{\partial}{\partial \bar{x}^{j}}=\sum_{j=1}^{q}(D \psi)_{i}^{j} \frac{\partial}{\partial \bar{x}^{j}} \quad \Rightarrow \quad g_{1,i j}=(D \psi)_{i}^{k} {g}_{2,k \ell}(D \psi)_{j}^{\ell} $ $ \quad \Rightarrow g_1^{-1}=\left(D \psi^{-1}\right)^{\top}  {g}_2^{-1} D \psi^{-1}$,
$G_1=(\operatorname{det} \mathcal{D} \psi)^{2} {G}_2$.
\end{itemize}

\begin{proof}
Consider a stochastic process in $\mathbb{U} \in \R^q$ defined by:
\begin{align}
d x^{i} &= \frac{1}{2} \frac{1}{\sqrt{ G_1 }} \sum_{j=1}^{q} \frac{\partial}{\partial x^{j}}\left(\sqrt{ G_1 } g_1^{i j}\right) dt + \left(g_1^{-\frac{1}{2}} d B\right)^{i}, \quad i=1, \ldots, q. 
\end{align}
Note that
\begin{align*}
\quad d x^{i} d x^{j}&=\left(g_1^{-\frac{1}{2}} d B\right)^{i}\left(g_1^{-\frac{1}{2}} d B\right)^{j} 
=\left( \sum_{k=1}^{q}\left(g_1^{-\frac{1}{2}}\right)^{i k} d B^{k}\right) \cdot\left(\sum_{l=1}^{q}\left(g_1^{-\frac{1}{2}}\right)^{j l} d B^{l}\right) \\ 
&=g_1^{i j} d t  \ \   \   i,j = 1,\ldots , q.
\end{align*}
$\psi$ maps the above process in $\mathbb{U}_1$ to a process in ${\mathbb{U}_2}$ defined by:
\begin{align}
\label{apxeqn:coord}
d \bar{x}^{i} &=d \psi^{i}\left(x^1, \ldots, x^{q}\right) \nonumber \\
&=\sum_{j=1}^{q} \frac{\partial \psi^{i}}{\partial x^{j}} d x^{j}+\frac{1}{2} \sum_{j=1}^{q} \sum_{k=1}^{q} \frac{\partial^{2} \psi^{i}}{\partial x^{j} \partial x^{k}} d x^{j} d x^{k} \nonumber \\
&= \frac{1}{2} \sum_{j=1}^{q} \sum_{k=1}^{q}\left[(D \psi)_{j}^{i} \cdot \frac{1}{\sqrt{G_1}} \frac{\partial}{\partial x^{k}}\left(\sqrt{G_1} g_1^{jk} \right)+\frac{\partial^{2} \psi^{i}}{\partial x^{j} \partial x^{k}} g_1^{j k}\right] d t +\sum_{j=1}^{q}(D \psi)_{j}^{i}\left(g_1^{-\frac{1}{2}} d B\right)^{j}.
\end{align}

The $dt$ term in \eqref{apxeqn:coord} can be written as follows

\begin{align}
&\frac{1}{2} \sum_{j=1}^{q} \sum_{k=1}^{q} \left[ (D \psi)_{j}^{i} \cdot \frac{1}{det D \psi} \cdot \frac{1}{\sqrt{{G_2}}} \cdot \sum_{\ell=1}^{q}(D \psi)_{k}^{l} \cdot \frac{\partial}{\partial \bar{x}^{\ell}} \left(det D \psi \cdot \sqrt{G_2} \sum_{r=1}^{q} \sum_{s=1}^{q}\left(D \psi^{-1}\right)_{r}^{j} {g_2}^{r s}(D \psi^{-1})_{s}^{k}\right) \right. \nonumber \\
&+\left(\sum_{l=1}^{q}(D \psi)_{j}^{l} \frac{\partial}{\partial \bar{x}^{\ell}}(D \psi)_k\right)  \left.\left(\cdot \sum_{r=1}^{q} \sum_{s=1}^{q}\left(D \psi^{-1}\right)_{r}^{j} {g}_2^{r s}\left(D \psi^{-1}\right)_{s}^{k}\right)\right] d t \nonumber \\
=& \frac{1}{2} \sum_{l=1}^{q} \frac{1}{\sqrt{{G_2}}} \frac{\partial}{\partial \bar{x}^{\ell}}\left(\sqrt{{G_2}} {g}_2^{i l}\right) d t +\frac{1}{2} \sum_{s=1}^{n} \operatorname{Tr}\left[D \psi \cdot \frac{\partial}{\partial \bar{x}^{s}}\left(D \psi^{-1}\right)\right] \cdot {g}_2^{i s} dt \nonumber \\
&+\frac{1}{2} \sum_{j=1}^{q} \sum_{l=1}^{q} \sum_{r=1}^{q} \frac{\partial}{\partial \bar{x}^{\ell}}\left[(D \psi)_{j}^{i}\left(D \psi^{-1}\right)_{r}^j\right] \cdot {g}_2^{rl} dt +\frac{1}{2} \sum_{l=1}^{n} \operatorname{Tr}\left[\frac{\partial}{\partial \bar{x}^{\ell}}(D \psi) \cdot\left(D \psi^{-1}\right)\right] {g}_2^{i \ell} d t  \nonumber \\
=& \frac{1}{2} \sum_{l=1}^{q} \frac{1}{\sqrt{{G_2}}} \frac{\partial}{\partial \bar{x}^{\ell}}\left(\sqrt{{G_2}} {g}_2^{i l}\right) d t 
+\frac{1}{2} \sum_{s=1}^{q} \operatorname{Tr}\left[\frac{\partial}{\partial \bar{x}^{s}}\left(D \psi \cdot D \psi^{-1}\right)\right] \cdot {g}_2^{i s} dt +0 \nonumber \\
=&\frac{1}{2} \sum_{\ell=1}^{q} \frac{1}{\sqrt{{G_2}}} \frac{\partial}{\partial \bar{x}^{\ell}}\left(\sqrt{{G_2}} {g}_2^{i \ell}\right) dt +0.
\end{align}

The $dB$ term in \eqref{apxeqn:coord} can be written in vector form as
\begin{align}
(D \psi)^{\top} \cdot g_1^{-\frac{1}{2}} \cdot d B.
\end{align}
Since
\begin{align*}
(D \psi)^{\top} \cdot g_1^{-\frac{1}{2}} \cdot d B \quad\left((D \psi)^{\top} g_1^{-\frac{1}{2}} d B\right)^{\top}  = {g}_2^{-1} d t,
\end{align*}
we have
\begin{align}
(D \psi)^{\top} \cdot g_1^{-\frac{1}{2}} \cdot d B \sim \mathcal{N}\left(0, {g}_2^{-1} d t\right),
\end{align}
i.e. the same distribution as ${g}_2^{-1/2}dB$.

Conclusion: $\psi$ maps the process in $\mathbb{U}_1 \subset \R^n$ defined by
\begin{align}
d x^{i}=\frac{1}{2} \frac{1}{\sqrt{G_1}} \sum_{j=1}^{q} \frac{\partial}{\partial x^{j}}\left(\sqrt{G_1} g_1^{i j}\right) d t+\left(g_1^{-\frac{1}{2}} d B\right)^{i}, \quad i=1, \ldots, q,
\end{align}
to the process in $\mathbb{{U}}_2 \subset \R^n$ defined by
\begin{align}
d \bar{x}^{i}=\frac{1}{2} \frac{1}{\sqrt{{G}_2}} \sum_{j=1}^{q} \frac{\partial}{\partial \bar{x}^{j}}\left(\sqrt{{G}_2} {g}_2^{i j}\right) d t+\left( {g}_2^{-\frac{1}{2}} d B\right)^{i}, \quad i=1, \ldots, q.
\end{align}
This verifies that the process defined by the above formula is coordinate independent.

As a result, with a given $dt$, simulating one step using any choice of local coordinates as above is equivalent as a step in $\mathbb{M}$.
\end{proof}

\subsection{Proof of Theorem 4.2 in the main paper}
\begin{theorem*} \label{theorem:positive}
The estimator $\hat{h}^t$ in equation (8) of the main paper is asymptotically unbiased and consistent.
\end{theorem*}

\begin{proof}
Let $\M$ be a manifold parameterised by an atlas $\mathcal{A}$ consisting of $n_v$ charts. Simulating Brownain Motion $\{S(t):t>0\}$ on $\M$ is equivalent to simulate a stochastic process governed by the Stochastic Differential Equations(SDEs) on the atlas $\mathcal{A}$. For each chart, the SDEs are derived as follows:
\begin{equation}
dx_k^i(t) = \frac{1}{2}{G_k}^{-1/2} \sum^{q}_{j=1}\frac{\partial}{\partial x_k^j} \left(  {\G_k}^{ij} {G_k}^{1/2} \right) dt + \left( {\G_k}^{-1/2} dB_k(t)\right)_i
\end{equation}
where $\G_k$ is the metric of chart $k$ and $G_k$ is the determinant of $\G_k$. $x_{k}^i$ is the coordinate in the $i_{th}$ dimension of chart $k$, and $B_{k}(t)$ is an independent BM in Euclidean space. 
Transition between different charts in the atlas is handled via the change of coordinates operations.
Consider a Brownian motion $\{S(t):t>0\}$ on $\M$ with $S(0)=s_0$. For any $t>0, s \in \M$, the true BM transition probability evaluated at $S(t)$ reaching a small neighborhood set $\mathbb{A}$ is given by:
\begin{equation}
    p(S(t) \in \mathbb{A} |S(0)=s_0)=\int_{\mathbb{A}} h^t(s_0,s)ds. 
\end{equation}
where $h^t(s_0,s)$ is the heat kernel on $\M$. Given a point $s$ with local coordinates $(r_1,\cdots,r_p)$, a window size $w$ as the measurement of the volume for neighborhood set $\mathbb{A}$, 
and assuming $r_1(s)= \cdots =r_p(s)=0$ for simplifying the calculation, 
the heat kernel can be approximated by:
\begin{equation}
    {h}^{t'}=\frac{1}{Vol(\mathbb{A})} p[|r_i(S(t))|<w \quad \text{for 
 }i=1,\cdots,p]
\end{equation}
where $Vol(\mathbb{A})=\{|r_i|<w, \quad i=1,\cdots,q\}$ is the Riemannian volume of $\mathbb{A}$. By applying the Taylor expansion at point $s$,
the  kernel estimator will be
\begin{equation}
    h^{t'}=h^t+ \mathcal{O}(w^2)
\end{equation}
where the order of the magnitude of $h^{t'}-h^t$ is $\mathcal{O}(w^2)$. Increasing $w$ thus increases the error term.

By counting how many of these paths reach $\mathbb{A}$ at time $t$ the transition probability can be evaluated. In other words, the transition density ${h^t}'$ of BM at $s$ is approximated by $\hat{h}^t$:
\begin{align}
\label{eqn:1stapprox}
\hat{h}^t = \frac{1}{Vol(\mathbb{A})}\cdot\frac{N_{A}}{N},\quad
\t{where } N_{A}\sim\t{Bin}(N,Vol(\mathbb{A} ) {h^t}' ).
\end{align}
$N_{A}$ is the number of sample paths with $\|S(t)-s\|< w$ (the number of paths reach the neighbourhood at time $t$). 
And follows a binomial distribution with $N$ trails (the total number of simulated BM paths) and probability of success of $Vol(\mathbb{A} ) {h^t}'$. The expectation of $\hat{h}^t$ is:
\begin{equation}
    E[\hat{h}^t]=E[\frac{1}{Vol(\mathbb{A})}\cdot\frac{N_{A}}{N}]=\frac{1}{Vol(\mathbb{A})} \cdot \frac{E[N_{A}]}{N}
\end{equation}
Since $N_A$ is binomially distributed, $E(N_A)=N \cdot Vol(\mathbb{A}){h^t}' $, which gives:
\begin{equation}
    E[\hat{h}^t]=\frac{1}{Vol(\mathbb{A})} \cdot \frac{N \cdot Vol(\mathbb{A}){h^t}'}{N} ={h^t}' = h^t +\mathcal{O}(w^2) 
\end{equation}
Besides, the standard error of $\hat{h}^t$ is:
\begin{align}
    sd ( \hat{h}^t )  & = \sqrt{ (\frac{1}{ Vol(\mathbb{A}) N})^2 \cdot Var(N_A)  } \\
     &=  \frac{1}{ Vol(\mathbb{A}) N} \cdot \sqrt{ N \cdot Vol(\mathbb{A}) {h^t}' (1-Vol(\mathbb{A}){h^t}') } \\
     &\leq \sqrt{ \frac{ {h^t}' }{ N \cdot Vol(\mathbb{A})}  } =\mathcal{O}( N^{-1/2}w^{-d/2})
\end{align}

The standard deviation decreases with $w$ and $N$. As $w^2 \rightarrow 0$, we obtain $E(\hat{h}^t(s_0, s)) = h^t(s_0, s)$, and as $\frac{w^{-d}}{N} \rightarrow 0$, we have $\text{Var}(\hat{h}^t(s_0, s)) \rightarrow 0$. Therefore, the estimator $\hat{h}^t(s_0, s)$ is asymptotically unbiased and consistent.
\end{proof}


\section{Autoencoder latent space of Torus.}
The autoencoder (AE, \cite{baldi2012}) can also be used to construct the latent space and learn the mapping function. It has been widely used as a neural network-based dimensionality reduction method. The encoder of the AE projects the original input into a lower-dimensional latent space, and the decoder of the AE maps the points in the latent space back to the original space. In fact, GPLVM can also be thought of as a GP decoder \citep{moreno2022}. Unlike GPLVM, which learn the mapping as a probabilistic function, 
the AE-learned mapping function, which is a neural network, is deterministic. For each subset $\S_i =\{ s_{ij} | j =1, \cdots, ns_i \}$, we use AE to learn the mapping $\varphi_i$ and construct latent space $\X_i$, where $s_{ij}=\varphi_i(x_{ij})$. 

An autoencoder (AE) is a neural network composed of nodes (or artificial neurons) organized in multiple layers to match the input data. 
The aim of Autoencoders is to reconstruct the output as similar as possible to the original input. To achieve this, a reconstruction error (or loss function) is introduced, which measures the discrepancy between the input $S_i$ and its reconstructions (the output) $\hat{S_i}$. It is important to note that separate AEs are modeled for each subset $S_i$. The goal of training an AE is to minimize this reconstruction error. In this context, we use mean squared error (MSE) as the reconstruction error.
\begin{equation}
    L(S_i,\hat{S_i}) = \frac{1}{ns_i} \sum_{j=1}^{ns_i} L(s_{ij}, \hat{s}_{ij})
\end{equation}
where $ns_i$ is the number of points in the $i$th subset $S_i$ and $s_{ij}$ is the $j$th observation of $i$th subset.

\begin{figure}[h!]
    \centering
    \subfigure[A subset of torus]{ \label{fig:torusSub}   \includegraphics[width=0.4\textwidth,height=0.4\textwidth]{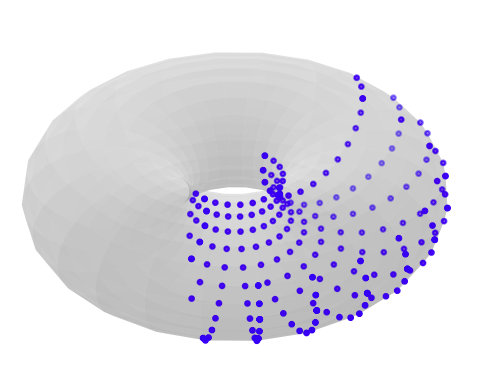}} 
    \subfigure[ AE Latent space]{ \label{fig:GPLVMlatent}  \includegraphics[width=0.4\textwidth,height=0.4\textwidth]{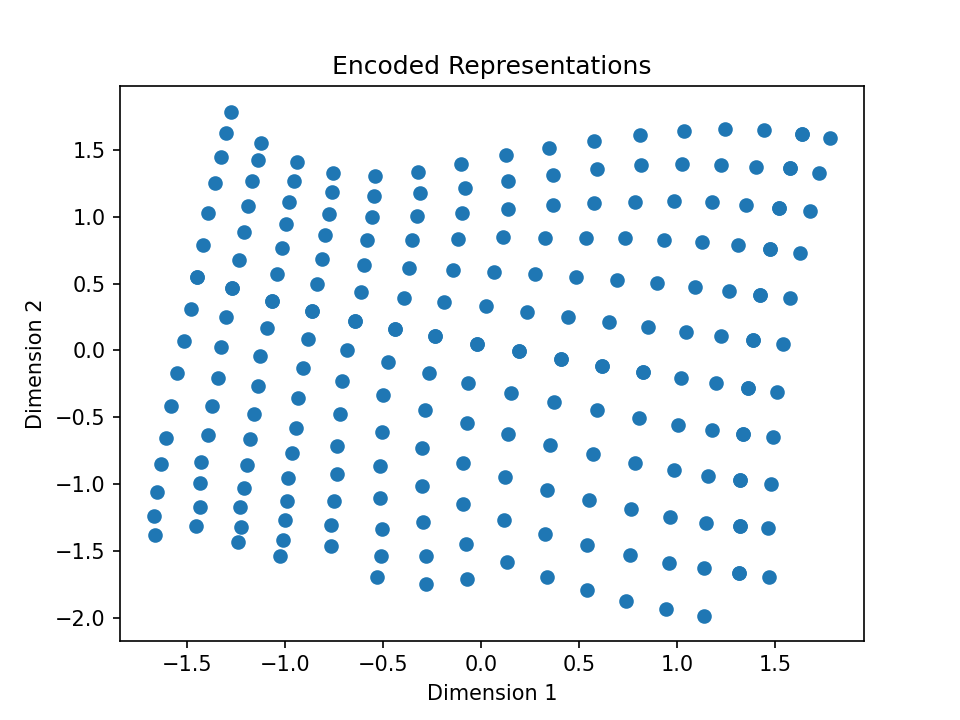}}
\caption{\label{fig:TorusAELatent}
    \footnotesize
    {\bf A subset latent spaces leaned from autoencoder of torus cloud.}  }
\end{figure}

Once an AE model is constructed, it allows for straightforward transformation of points between the latent space and the original space. We denote an encoder as $\varphi_i^{-1}$ that is a feature-extracting function in a pre-determined parametric closed form(e.g., Rectified Linear Activation Function). The latent variable are estimated as $X_i=\varphi_i^{-1}({S_i})$.  In this context, the encoder can also be considered as the backward mapping. The Decoder is the forward mapping represented as $\varphi_i$, which maps from the latent space back to the original input space, generating a reconstruction $\hat{S_i}=\varphi_i({X_i})$. Fig. \ref{fig:torusSub} shows one subset of the Torus point cloud, and Fig. \ref{fig:TorusAELatent}(b) is the corresponding latent space generated by the AE decoder.

As the Riemannian metric defined in the main parper, the metirc tensor $\mathbf{g}_i$ for each subset $S_i$ are
\begin{equation}
    \mathbf{g}_{i}=\mathcal{J}_{i}^T \mathcal{J}_{i}
\end{equation}
where the $\mathcal{J}_{i}$(the Jacobian matrix of subset $S_i$) can be regarded as the  partial derivative $\frac{\partial \varphi_i(x_*)}{\partial x}$.
It can be calculated in AE decoder model by applying the chain rule of the partial derivative of the activate function and the inputs in each layers~\citep{rifai2011contractive}.


\section{Gaussian Process Latent Variable Model}

The Gaussian Process Latent Variable Model (GPLVM, \cite{lawrence2005})is a probabilistic latent variable model. The probabilistic mapping links the latent variable in latent space $x \in \R^q$ and its corresponding observations in the original dimension $s \in \R^p$. For each subset $\mathcal{S}_i$, independent mappings are defined. We have
\begin{equation}
    s^l_{ij}=\varphi^l_i(x_{ij})+\epsilon^l_{ij}
\end{equation}
where $s_{ij}^l$ is $i$th chart, $l$th dimension of $j$th observations and  $\epsilon_{ij}^l$ is a i.i.d noise which is typically taken to be Gaussian $p(\epsilon)=\mathcal{N}(0, \beta^2)$. 
GPLVMs place a prior on the mapping, $\varphi^l_i(x_{ij})\sim \mathcal{N}(0, \mathcal{K}_{xx})$ where $\mathcal{K}$ is a RBF kernel. Let $X_i$ denotes the set of all latent variables $x_{ij}$ correspondent to $\mathcal{S}_i$.
With the assumption that the prior is independent across the dimension in the orginal data space, the marginal likelihood can be written as:
\begin{align}
    p(S_i|X_i)&=\int \prod_{l=1}^p p(S_{i}| \varphi_i^l,\beta) p(\varphi_{i}^l|X_{i}) d \varphi_i \\
    &=\prod_{l=1}^p \mathcal{N}(s_{i:}^l, \mathcal{K}_{XX}+\beta^2I)
\end{align}
where $\mathcal{K}_{XX}$ is the covariance matrix generated by the RBF kernel. $s_{i:}^l$ represents the lth dimension of all points in $\mathcal{S}_i$. The latent variables and the kernel hyper-parameters can be estimated by maximum likelihood estimates.
\begin{equation*}
X_{i}=\mathbf{arg} \max_{X_i} p(S_i|X_i)
\end{equation*}
When simulating BM in latent space, we can observe a new location in each time step as testing data denoted as $x_*$. $\varphi_i(x_*)$ is the projection of $x_*$ in the original high dimensional space.  The joint distribution of $\varphi_i(X_i)$ and $\varphi_i(x_*)$ is:

\begin{equation}
    \begin{bmatrix}
       \varphi_i(X_i)   \\
       \varphi_i(x_*)
    \end{bmatrix} \sim \mathcal{N}( \begin{bmatrix}
       0   \\
      0
    \end{bmatrix} , \begin{bmatrix}
      \mathcal{K}_{XX}  & \mathcal{K}_{Xx^*}   \\
       \mathcal{K}_{x^*X}  & \mathcal{K}_{x^*x^*}
    \end{bmatrix})
\end{equation}
where $\mathcal{K}_{Xx^*}$ is the covariance matrix between given points and new points. The predictive distribution is
 
\begin{align}
    p(\varphi_i(x_*)|S_i) & \sim \mathcal{N}(\mathcal{K}_{x^*X} (\mathcal{K}_{XX}+\beta^2 I_n)^{-1} S_i,\mathcal{K}_{x^*x^*}-\mathcal{K}_{x^*X} (\mathcal{K}_{XX}+\beta^2 I_n)^{-1} \mathcal{K}_{x^*X}^T) 
\end{align}

In addition to GPLVM, a variational method, the Bayesian Gaussian Process Latent Variable Model (Bayesian GPLVM), can also be applied. A key advantage of Bayesian GPLVM is its incorporation of uncertainty in the latent variables, which allows for a distribution over the latent space rather than a fixed point estimate~\citep{titsias2009variational}. This leads to more robust handling of small datasets and improved generalization by accounting for uncertainty in the latent space. However, the increased number of hyperparameters introduces additional complexity to the optimization process, which can lead to convergence difficulties in some cases. In the following section, we will focus on GPLVM.

\subsection{GPLVM mappings}
Since we know how to infer latent variables from the original manifold and generate new points in high dimensions from latent space points, we can construct both forward and backward mappings between the latent space and the original manifold.

Denote $x_*$ as a point in the latent space, by applying the predictive distribution (forward mapping), the corresponding point $s^*$ in the manifold can be generated.

\begin{equation}
    s^* | \mathcal{S}_i \sim \mathcal{N}(\mathcal{K}_{x^*X} (\mathcal{K}_{XX}+\beta^2 I_n)^{-1} S_i ,\mathcal{K}_{x^*x^*}-\mathcal{K}_{x^*X} (\mathcal{K}_{XX}+\beta^2 I_n)^{-1} \mathcal{K}_{x^*X}^T) 
\end{equation}
On the contrary, when a point $s^*$ on the manifold is given in the high dimensional space, we can estimate the latent variables in a chart ($x_*$ in the latent space) by maximising the likelihood(backword mapping).
\begin{equation*}
x_{*}=\mathbf{arg} \max_{x_*} p(s_*,S_i|X_i,x_*)
\end{equation*}

\subsection{ GPLVM metric}
The Jacobian of $i$th chart, $\mathcal{J}_i$ can be computed as the partial derivative $\frac{\partial \varphi_i(x_*)}{\partial x^l} $ with respect to the $l$th dimension for any point $x_*$ in the latent space.
\begin{equation}
    \mathcal{J}_i=\frac{\partial \varphi_i(x_*)}{\partial x} =
    \begin{bmatrix}
        \frac{\partial \varphi_i^1(x_*)}{\partial x^1}&  \cdots  & \frac{\partial \varphi_i^1(x_*)}{\partial x^k} \\
        \vdots & \cdots &  \vdots \\
        \frac{\partial \varphi_i^l(x_*)}{\partial x^1} &  \cdots  & \frac{\partial \varphi_i^l(x_*)}{\partial x^k} 
    \end{bmatrix}
\end{equation}
where the Jacobian is a $q\times p$ matrix. Recall that $p$ represents the dimension of the original space and $q$ denotes the dimension of the latent space. Since the RBF Kernel is differentiable, the derivative of a GP is also a GP. Therefore the distribution of Jacobian for chart $i$ is
\begin{equation} p(\mathcal{J}_i|X_i,S_i)=\prod_{l=1}^p \mathcal{N}(\mu_{\mathcal{J}_i}^l,\Sigma_{\mathcal{J}_i})
\end{equation}
The expression for $\mu_{J_i}^l,\Sigma_{J_i}$(the mean and covariance for the Jacobian distribution) are shown later. The expected metric tensor $\mathcal{G}_i$ for $ith$ chart is defined as:
\begin{align} 
\mathcal{G}_i = \E(\mathcal{J}_i^T) \E(\mathcal{J}_i) + p \Sigma_{\mathcal{J}_i} 
\end{align}
 
The joint distribution of $l$th dimension of the mapping $\varphi_i$ and the corresponding column of the Jacobian are
\begin{equation}
     \mathcal{J}_i^{l,k} = \frac{\partial \varphi_i^l }{ \partial  x^k} ,\ \ \ 
    \begin{bmatrix}
        \varphi_i^l(x) \\
        \frac{\partial \varphi_i^l(x_*)}{\partial x_*} 
    \end{bmatrix} 
    \sim
    \mathcal{N} \begin{pmatrix}
        0 & \begin{bmatrix}
            \mathcal{K}_{xx} & \partial \mathcal{K}_{x*} \\
            \partial \mathcal{K}^T_{x*}  & \partial^2 \mathcal{K}_{**} 
        \end{bmatrix}
    \end{pmatrix}
\end{equation}
where the superscript indicates the $l$th dimension of the $i$th chart and $k$th dimension of the observation space. The covariance matrix is computed as
\begin{equation*}
    \mathcal{K}_{xx_*}= \gamma \mathbf{exp}(-\rho \|x_{ij}-x_*\|^2)
\end{equation*}
where $\gamma $ and $\rho$ regarded as hyperparameters are length scale and variance for the RBF kernel. The derivatives of $\mathcal{K}$ are:
\begin{align*}
    (\partial \mathcal{K}_{x*})^l_j &=\frac{\partial \mathcal{K}_{xx_*}}{\partial x^l_*}=2 \rho(x_{ij}^l-x_*^l)\mathcal{K}_{xx_*} \quad j=1,\cdots, n, \quad l=1,\cdots,q \\
    (\partial^2 \mathcal{K}_{**})^{r,l} &=\frac{\partial^2 \mathcal{K}_{x^{'}_*x_*}}{\partial x^r_* \partial x^l_*} = 
    \left \{ \begin{array}{l}
         -4 \rho^2 ({{x^{'}}_*^r}-x_*^r)({{x^{'}}_*^l}-x_*^l)\mathcal{K}_{x^{'}_*x_*} \quad r \neq l\\
        2\rho(1-2 \rho ({{x^{'}}_*^r}-x_*^r)^2)\mathcal{K}_{x^{'}_*x_*} \quad r = l
    \end{array}\right.
\end{align*}
When $x_*=x_*'$
\begin{equation*}
       (\partial^2 \mathcal{K}_{**})^{r,l} =\frac{\partial^2 \mathcal{K}_{x_*x_*}}{\partial x^r_* \partial x^l_*} = \left \{ \begin{array}{l}
         0 \quad r \neq l \\
        2\rho \mathcal{K}_{x_*x_*}=2\rho \gamma \quad r = l
    \end{array}\right.
\end{equation*}
The conditional probability over the Jacobian also follows a Gaussian distribution. For any point $x_*$ in the latent space, the distribution of the Jacobian can be expressed as
\begin{equation}
    p(\mathcal{J}_i |S_i,\varphi_i)=\prod^p_{l=1} \mathcal{N}( \partial \mathcal{K}^T_{x,*}\mathcal{K}^{-1}_{x,x}s_{i:}^l, \partial^2 \mathcal{K}_{*,*}-\partial \mathcal{K}^T_{x,*}\mathcal{K}^{-1}_{x,x} \partial \mathcal{K}_{x,*})
\end{equation}


\subsection{ Magnification factor}
The magnification factor(MF) refers to the square root of the determinant of the Riemannian metric~\citep{bishop1997magnification}.
\begin{equation}
    \mathcal{MF}=\sqrt{det(\mathcal{G})}
\end{equation}
The magnification factor can express how much a part of latent space in $\R^q$ is stretched or bend when mapped to $\M$. In other words, it quantifies how distances in the latent space $X_i$ are transformed when mapped to the high-dimensional observed space $S_i$. A higher magnification factor indicates more stretching in the latent space.

\begin{figure}[h!]
\centering
\subfigure[\footnotesize MF with true mapping]{
\begin{minipage}[t]
{0.32\linewidth}\label{Fig:true-MF}
\centering
\includegraphics[width=2. in, height= 2  in]{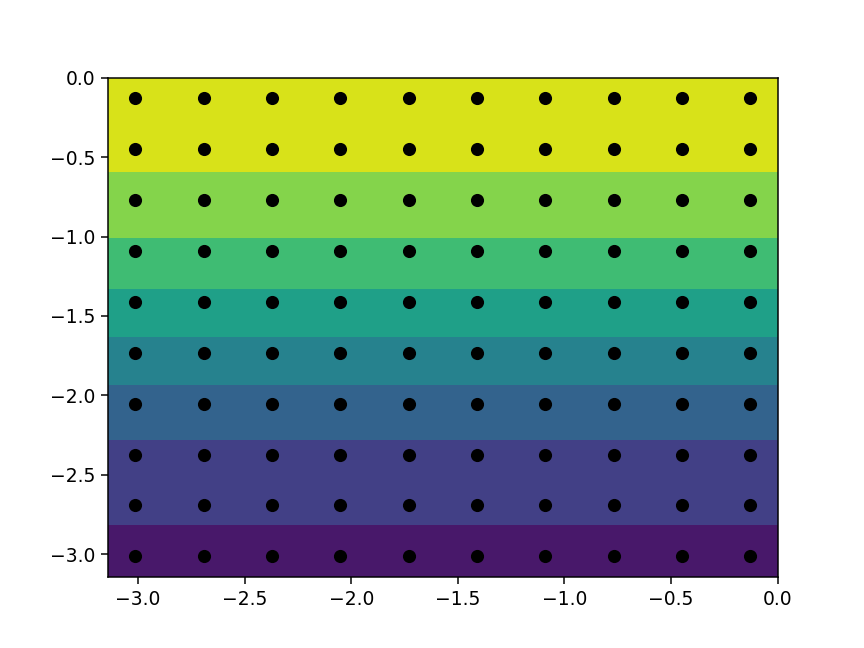}
\end{minipage}%
}
\centering
\subfigure[\footnotesize GPLVM MF ]{
\begin{minipage}[t]{0.32\linewidth}\label{Fig:BGP-MF}
\centering
\includegraphics[width=2. in, height= 2  in]{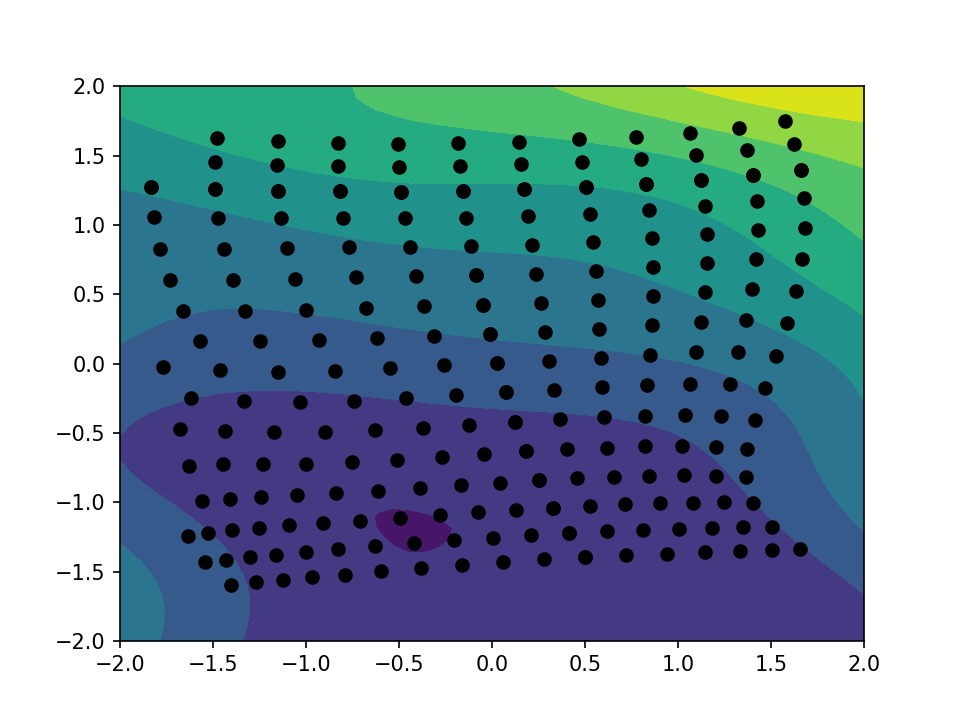}
\end{minipage}%
}%
\centering
\subfigure[\footnotesize AE MF]{
\begin{minipage}[t]{0.32\linewidth}\label{Fig:NN-MF}
\centering
\includegraphics[width=2. in, height= 2  in]{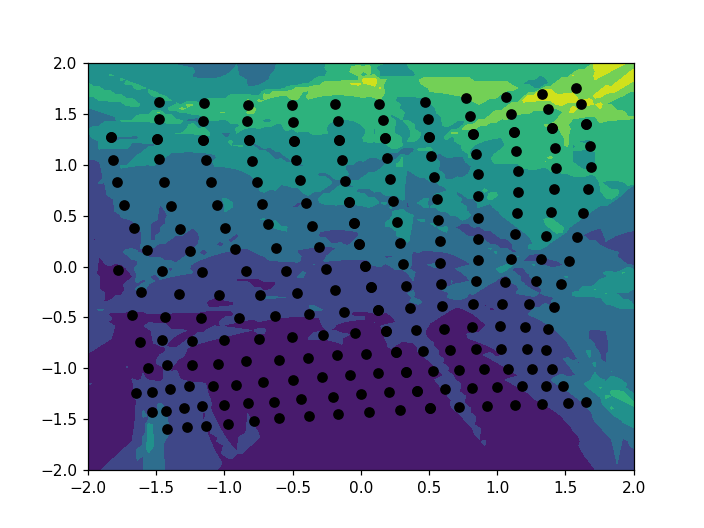}
\end{minipage}%
}%
\centering
\caption{\small Plots of the magnification factor in one of the latent spaces of the torus, constructed using different methods: (a) MF with true  mapping; The background color corresponds to the scale of MF. (b)MF with GPLVM mapping. (c) MF with Autoencoder mapping.  }\label{Fig:MFplot}
\end{figure}

For example, Fig.~\ref{Fig:MFplot} illustrates a latent space generated from $S_i$ using different methods. The background color of Fig.~\ref{Fig:MFplot} represents the magnification factor, where darker color indicate larger magnification. In Fig.~\ref{Fig:true-MF}, the vertical axis corresponds to the angle of the inner circle, while the horizontal axis represents the angle of the outer circle. As the inner circle angle approaches $- \pi$, corresponding to the inner hole of the torus, the latent space becomes increasingly stretched when mapped back to the original manifold. The magnification factor computed from the analytical mapping function is plotted in Fig.~\ref{Fig:true-MF}. In Fig.~\ref{Fig:BGP-MF}, where the mapping function is generated by GPLVM, the magnification factor appears in a similar pattern, but the contours are curved. The the magnification factor generated from AE mapping is plotted in Fig.~\ref{Fig:NN-MF}. The density of the magnification factor varies less smoothly, and the overall pattern deviates notably from the analytical result.

\section{ Kernel estimates on scatter Torus.}

\begin{figure}[h!]
\centering
\subfigure[\footnotesize Non-uniform and irregular distributed Torus point cloud]{
\begin{minipage}[t]
{0.32\linewidth}\label{Fig:scatter3d}
\centering
\includegraphics[width=1.8in, height=1.5in]{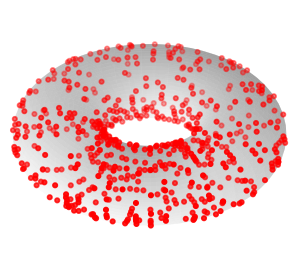}
\end{minipage}%
}
\centering
\subfigure[\footnotesize  Comparision kernel density while varying $\theta$ ]{
\begin{minipage}[t]{0.32\linewidth}\label{}
\centering
\includegraphics[width=1.8in, height=1.5in]{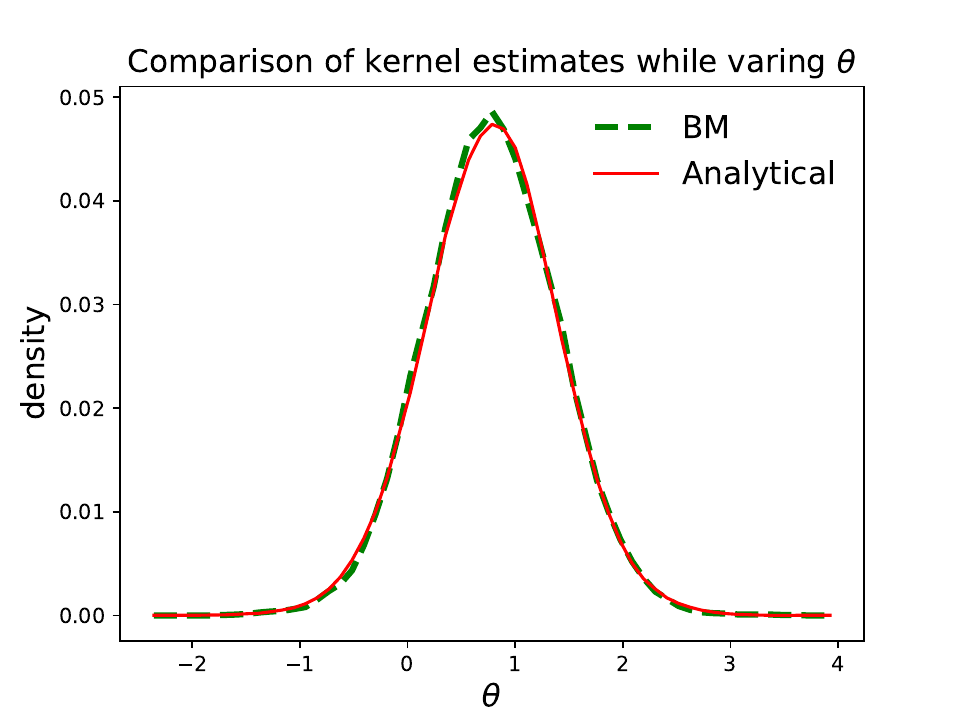}\label{Fig:scatterthata}
\end{minipage}%
}%
\centering
\subfigure[\footnotesize Comparision kernel density while varying $\phi$ ]{
\begin{minipage}[t]{0.32\linewidth}\label{}
\centering
\includegraphics[width=1.8in, height=1.5in]{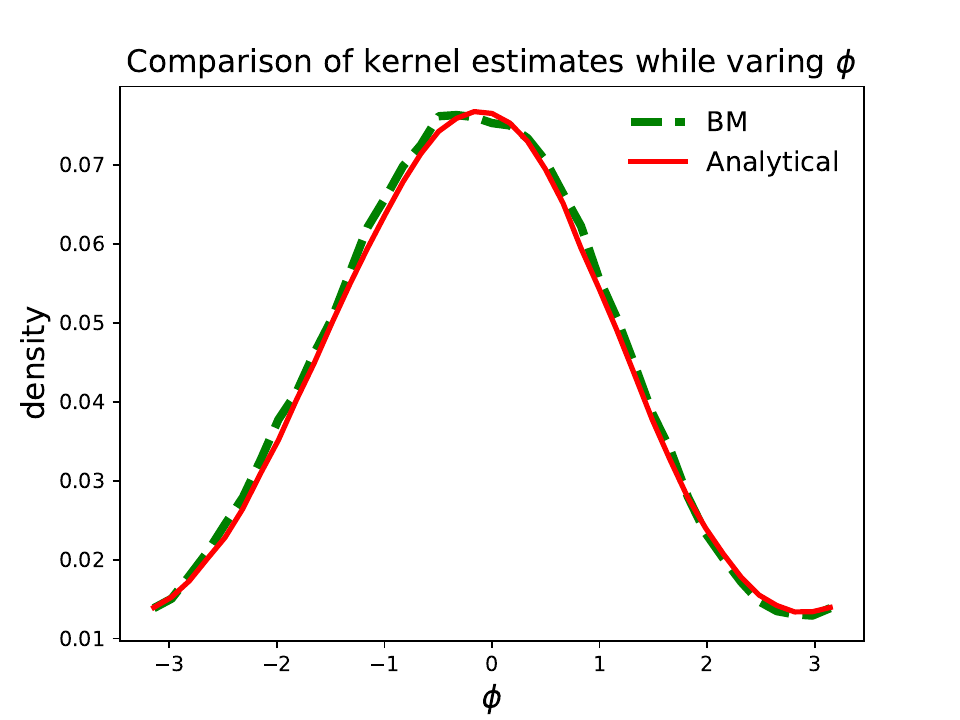}\label{Fig:scatterphi}
\end{minipage}%
}%
\centering
\caption{\small  Kernel estimates of scatter torus. The horizontal axes represent the outer circle angle $\theta$ and inner circle angle $\phi$, while the
vertical axis shows the transition density. The truth is presented as the red line and the green dashed line represents the kernel estimates generated by atlas BM approach. }\label{Fig:Scatter}
\end{figure}

We tested the robustness of our Atlas Brownian Motion (BM) framework by varying the configuration of the torus, specifically the radius of the inner circle $\phi$ and the outer circle $\theta$, and the distribution of points on the torus. A total of 625 points were sampled from a torus with an outer circle radius of $R=2$ and an inner circle radius of $r=1$, forming a point cloud with a non-uniform and irregular distribution, as illustrated in Fig. \ref{Fig:Scatter}(a).

In this experiment, we implemented the Atlas Brownian Motion (BM) framework outlined in the main paper. The probabilistic atlas was constructed from the non-uniform torus point cloud using latent variable models. Kernel density estimation was then performed based on BM simulations across the atlas.

Fig.~\ref{Fig:scatterthata}~and~\ref{Fig:scatterphi} show the inner and outer circle angles of the scatter torus on the horizontal axes, while the vertical axis represents the transition density. The kernel estimates derived from the Atlas BM approach (the green dashed line) are closely aligned with the true values (the red solid line). This demonstrates the robustness and accuracy of the Atlas BM framework in estimating kernel densities, even with complex, irregular point clouds.

\section{ Problematic Single chart for Torus point cloud.}
\begin{figure}[htbp]
    \centering
    \subfigure[Single chart latent space]{ \label{Fig:Onechart} \includegraphics[width=0.45\linewidth,height=0.35\textwidth]{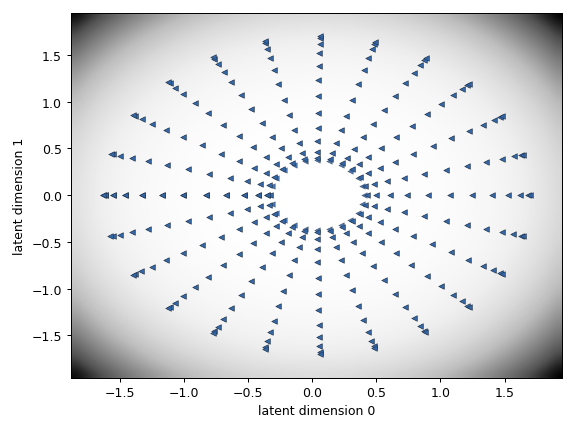}} 
    \subfigure[ One sample path on $\M$ generated by the single chart ]{ \label{Fig:onechart-sde}  \includegraphics[width=0.45\linewidth,height=0.4\textwidth]{onechart-3d.png}}
\caption{\label{fig:}
    \footnotesize
    {\bf (a)The latent space of the torus cloud generated by the single-chart approach; (b) The top view of the sample path generated by the single-chart approach.  }  }
\end{figure}

A manifold cannot be covered by a single chart, unless it is homeomorphic to an open subset of an Euclidean space \citep[Section~1]{Lee13}. In this section we will show the problem of using a single chart to parameterise a torus point cloud which have non-trivial topology. As shown in Fig.~\ref{Fig:Onechart}, we created a latent space for the torus point cloud using GPLVM.
The distribution of latent variables exhibits a pattern resembling a sunflower. 
This suggests that the latent space is generated by compressing the torus, leading to points stacking on top of each other.

\begin{figure}[h!]
    \centering    \includegraphics[width=0.55\textwidth,height=0.4\textwidth]{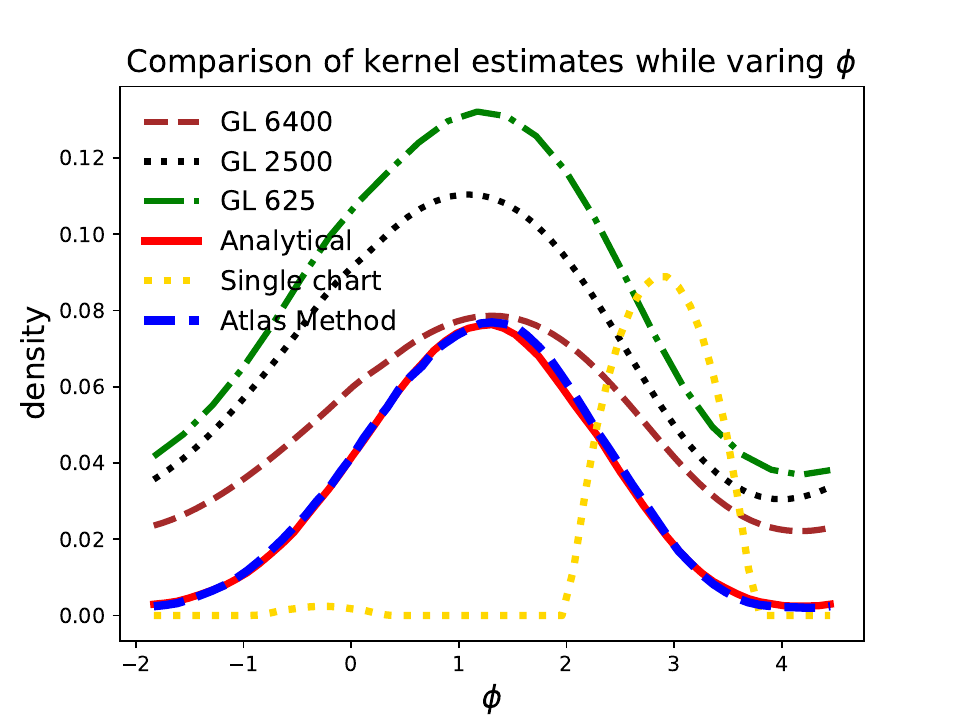}
 \caption{\label{fig:compKphi}
    \footnotesize
    {\bf  Kernel estimates comparison for different methods. The analytical kernel density is represented by the solid red line, while the kernel density generated from a single chart is depicted by the yellow dashed line. The kernel estimates obtained using the BM on Atlas approach are illustrated as a dashed blue line, with the atlas constructed from 625 points on the torus. The GL kernel estimates are plotted as a green dash-dotted line when the number of points is 625. Additionally, the GL estimates are represented as a black dotted line and a brown dashed line for increased point counts of 2,500 and 6,400, respectively.} } 
\end{figure}

A stochastic process has been simulated in the latent space and projected back to the observational space, illustrated in Fig.~\ref{Fig:onechart-sde}. It is clear that the simulated path travels inside the Torus, treating the manifold as a disk. It indicates the stochastic process is not a valid BM on $\M$. Fig.~\ref{fig:compKphi} presents the kernel density estimates from different methods. The horizontal axis represents the inner circle angle $\phi$ of the torus, while the vertical axis represents the transition density. The yellow dashed line represents the kernel estimates from the single-chart BM approach, which deviate notably from the analytical kernel, with a skewed peak compared to other estimates. The kernel density estimates using the Atlas BM method introduced in the main paper, is depicted as the blue dashed line in Fig. \ref{fig:compKphi}. It closely aligns with the analytical kernel(the red solid line). Additionally, the kernel can also be estimated using the Graph Laplacian (GL) method. By adjusting the number of grid points, the GL kernel estimates shift from the green dash-dotted line to the black dotted and brown dashed lines, respectively. These results highlight the limitations of the single-chart approach in capturing the manifold's geometry, leading to unreliable kernel estimates.


\section{Define the overlapping region in the latent space of Torus subsets}
Given the torus subsets generated from the Section 3 in main paper, we can identify overlapping points among different subsets. For example, points that belong to both subsets \( {S}_1 \) and \( {S}_2 \) will be classified as Category 12. In contrast, points exclusive to a single chart(either \( {S}_1 \) or \( {S}_2 \)) will be labeled as either Category 1 or 2, depending on the respective chart. This labeling process can be extended to assign labels to points across all charts within the atlas. Once the shared points between different charts are identified, classifiers such as decision trees \citep{hastie2009elements}, can be used to define overlapping regions in the latent spaces, facilitating the switching of SDEs between different charts. 
\begin{figure}[h!]
\centering
\subfigure[\footnotesize One of the latent space of Torus]{
\begin{minipage}[t]
{0.32\linewidth}\label{points}
\centering
\includegraphics[width=1.8in, height=1.5in]{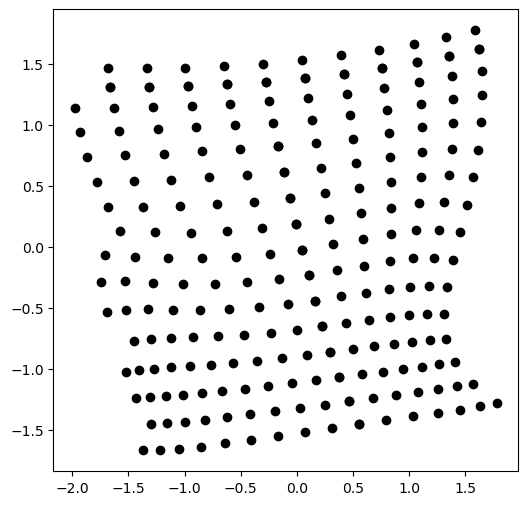}
\end{minipage}%
}
\centering
\subfigure[\footnotesize Points labeled by categories]{
\begin{minipage}[t]{0.32\linewidth}\label{classifier}
\centering
\includegraphics[width=1.8in, height=1.5in]{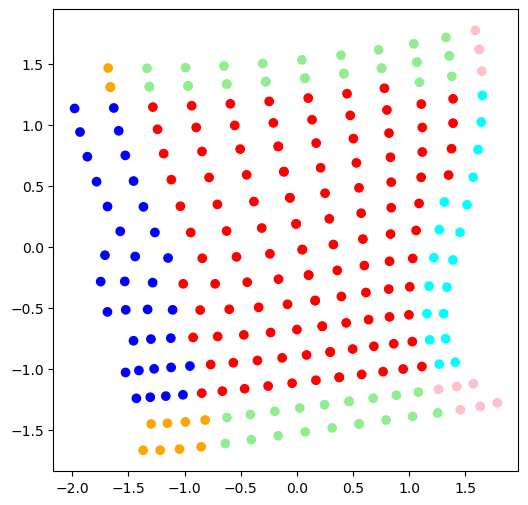}
\end{minipage}%
}%
\centering
\subfigure[\footnotesize Points with labels and overlapping regions ]{
\begin{minipage}[t]{0.32\linewidth}\label{boundaries}
\centering
\includegraphics[width=1.8in, height=1.5in]{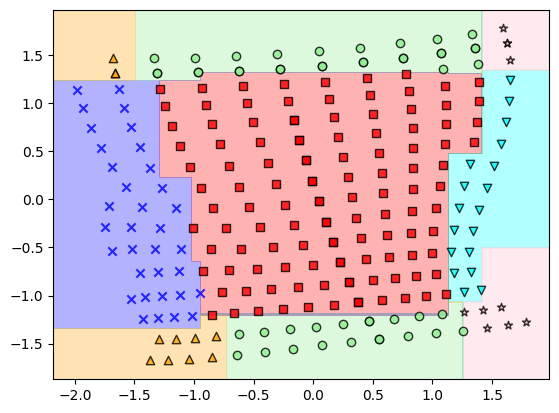}
\end{minipage}%
}%
\centering
\caption{\small   
Identify overlapping regions in the latent space: (a) Visualization of a latent space of Torus. (b) Data points labeled by colors, where each color denotes a category. (c) Overlapping regions in the latent space: The regions highlighted in yellow, pink, blue, green, and cyan represent areas that overlap with other subsets in the latent space. The red region does not overlap with any other subsets.}\label{Fig:decisiontree}
\end{figure}

As shown in Fig.~\ref{Fig:decisiontree}, the decision tree method is applied to identify the overlapping regions in the latent space. Fig.~\ref{points} illustrates one of the latent spaces of Torus, while Fig.~\ref{classifier} presents the points labelled with six different colors, each representing a different category. The red points correspond to points belonging exclusively to a single chart, whereas the remaining colors (green, yellow, cyan, blue, and pink) indicate points shared across at least two charts. 
The decision tree method is applied and identifies the overlapping region in the latent space in Fig.~\ref{boundaries}. The background colors represent the regions associated with each category. 
When simulating SDE steps in the latent space, if the simulated point enters an overlapping region, such as the left middle area (blue area), the point can be transferred to another latent space via the corresponding mapping function.


\section{Sparse Atlas GP}
As a comparison to the RC-AGP, the Sparse Atlas GP (S-AGP) are developed by combining the Atlas BM framework with sparse GP inference. In this approach, we first introduce a set of inducing points on $\M$, $\bm{s_z}=[s_{z1},...,s_{zm} ]$, $s_{zi}\in \M$. $zm$ is the number of inducing points. The realisation of the regression function on these inducing points can be represented as $zm$ random variables $\bm{u}=[f(s_{z1}),...,f(s_{zm})]$. The distribution of $\bm{u}$ is a multivariate Gaussian, $\bm{u} \sim \mathcal{N}( 0 , \Sigma_{\bf uu} )$, with mean zero and covariance matrix $\Sigma_{\bf uu}$. Each element of  $\Sigma_{\bf uu}$ can be computed as the heat kernel estimates in Section 4 of the main paper.  To facilitate efficient inference, we employ the Subset of Regressors (SoR) approximation \citep{smola2000,quin07}. Let $\bf{f}$ represent the vector of function values at the labeled points, and $\bf{f}_*$ denote the function values at the unlabeled test points. The prior for the Sparse Atlas GP (S-AGP) can then be expressed as the following

\begin{align*}
  q({\bf f,f_*})= \mathbf{N} \left( 0, \left[\begin{array}{cc}  
Q_{\bf ff} & Q_{\bf ff_*} \\[0.3em]
Q_{\bf f_*f} & Q_{\bf f_*f_*} \\[0.3em]
           \end{array} \right] \right) 
 \end{align*}
\noindent where $Q_{\bf a,b} = \Sigma_{\bf a, u}\Sigma_{\bf u,u}^{-1}\Sigma_{\bf u,b}$. By applying Algorithm in Section 4, the matrices $\Sigma_{ {\bf uu}}$, $\Sigma_{{\bf uf}}$, and $\Sigma_{{\bf uf_*}}$ are derived by estimating the transition densities of the BM simulation paths, using the inducing points as the initial points. With this approximation, the marginal distribution of the S-AGP can be expressed as:
\begin{equation}
p(\bm{y}|{\bf f}) \approx q(\bm{y}|\bm{u}) = \prod_{i=1}^{n} \mathbf{N} \left( y_i| \Sigma_{f_i \bf u} \Sigma_{\bf uu}^{-1} {\bf u} , \sigma_{noise}^2 {\bf I} \right). 
\end{equation}
The inducing points in the marginal likelihood can be integrated out by incorporating their prior distribution:
\begin{equation} \label{sparseLik}
p(\bm{y}) = \int q(\bm{y}|\bm{u}) p(\bm{u})d\bm{u} = \mathbf{N} \left( 0, \Sigma_{\bf fu} \Sigma_{\bf uu}^{-1} \Sigma_{\bf uf}+\sigma_{noise}^2 {\bf I} \right).
\end{equation}

The hyper-parameters such the diffusion time and rescaling parameter can be optimised by maximising the marginal likelihood. Using the proposed model, the predictive distribution can be derived as follows:
\begin{align}
q( {\bf f}_*| \bf{y} ) &= \mathbf{N} \left( Q_{\bf f_*f}  \left( Q_{\bf ff}+ \sigma^2 \bf{I} \right)^{-1} {\bf y} , Q_{\bf f_*f_*}-Q_{\bf f_*f}(Q_{\bf ff}+\sigma^2 {\bf I})^{-1} Q_{\bf ff_*} \right).
\end{align}
By employing the SoR approximation, the number of required BM sample path simulations is significantly reduced from $n_d \times N_{bm}$ to $m \times N_{bm}$, where $m$ denotes the number of inducing points, $n_d$ represents the number of data points, and $N_{bm}$ is the number of BM paths per starting point. Additionally, the computational complexity of inverting the covariance matrix is reduced from $O(n_d^3)$ to $O(n_d \times m^2)$. 

\begin{table}
\label{tab:induce}
\caption{Root-Mean-Squared Error of predictive means for S-AGP on Torus point cloud with different numbers of inducing points}
\centering
\begin{minipage}[t]{\linewidth}
\centering
    \begin{tabular}{|c|c|}
        \hline
         \em The Number of inducing points &\em Sparse Atlas GP\\
        \hline
         8 & 0.48(0.05) \\
        10 & 0.39(0.03) \\
        16 & 0.38(0.02) \\
        \hline
    \end{tabular}
\end{minipage}
\end{table}

The number of inducing points in sparse GP can affect the regression performance~\citep{galy2021adaptive,moss2023inducing}. Sparse GPs approximate the posterior using a smaller number of inducing points. However, if the number of inducing points is too small, the model may fail to capture the underlying trends in the dataset, leading to reduced predictive performance. To illustrate the impact of inducing point quantity on model performance, we apply S-AGP on the torus point cloud example and vary the number of inducing points. The mean RMSE (calculated over 10 replicates) presented in Table \ref{tab:induce}. In each replicate, there are 30 labeled points and 625 grid points. The observed labeled points are considered noise free. From Table \ref{tab:induce}, we can see that when the number of inducing points increases from 8 to 16, the Mean RMSE of S-AGP reduces from 0.48 to 0.38. Additionally, the standard deviation decreases from 0.05 to 0.02, indicating improved stability and accuracy of the model with more inducing points. Although the inducing points are randomly selected, we ensure that each chart contains at least one inducing point to maintain coverage across the dataset.

\section{Regression results of Sparse Atlas GP}
In the main paper, we have already discussed results generated by different methods across various applications. In this section, we will focus on the results produced by the S-AGP method for these applications.

\begin{figure}[h!]
    \centering
    \subfigure[True function on Torus in 2D]{ \label{fig:Tpoint}  \includegraphics[width=0.44\textwidth,height=0.4\textwidth]{Fig/TorusFun2d.png}} 
   \subfigure[S-AGP prediction]{ \label{fig:saGP}   \includegraphics[width=0.44\textwidth,height=0.4\textwidth]{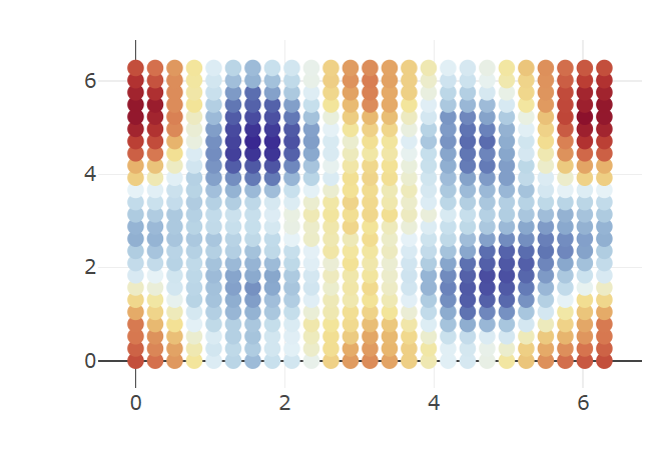}}   
 \caption{\label{fig:Torus}
    \footnotesize
    {\bf True function and prediction on Torus: (a) true function in the point cloud; The red to blue color indicate the true function values on the grid points (b) S-AGP prediction at the
 grid points  }  }\label{fig:torus2d}
\end{figure}

Fig.~\ref{fig:torus2d} presents the true function and S-AGP predictions on the torus point cloud. 
For better visualization, the true function values are displayed in two-dimensional plot(Fig.~\ref{fig:Tpoint}), where the horizontal axis corresponds to the horn angle $\theta$ and the vertical axis to the ring angle $\theta$. Fig.~\ref{fig:saGP} display the S-AGP predictions. The blue regions in Fig.~\ref{fig:saGP} appear more bracket-like rather than four circles in the Fig.~\ref{fig:Tpoint}. Despite this discrepancy, S-AGP outperforms GL-GP and Euclidean $\R^3$GP, as the central region in the S-AGP prediction appears orange, rather than light blue from other methods.


\begin{figure}[h!]
    \centering
    \includegraphics[width=0.5\linewidth]{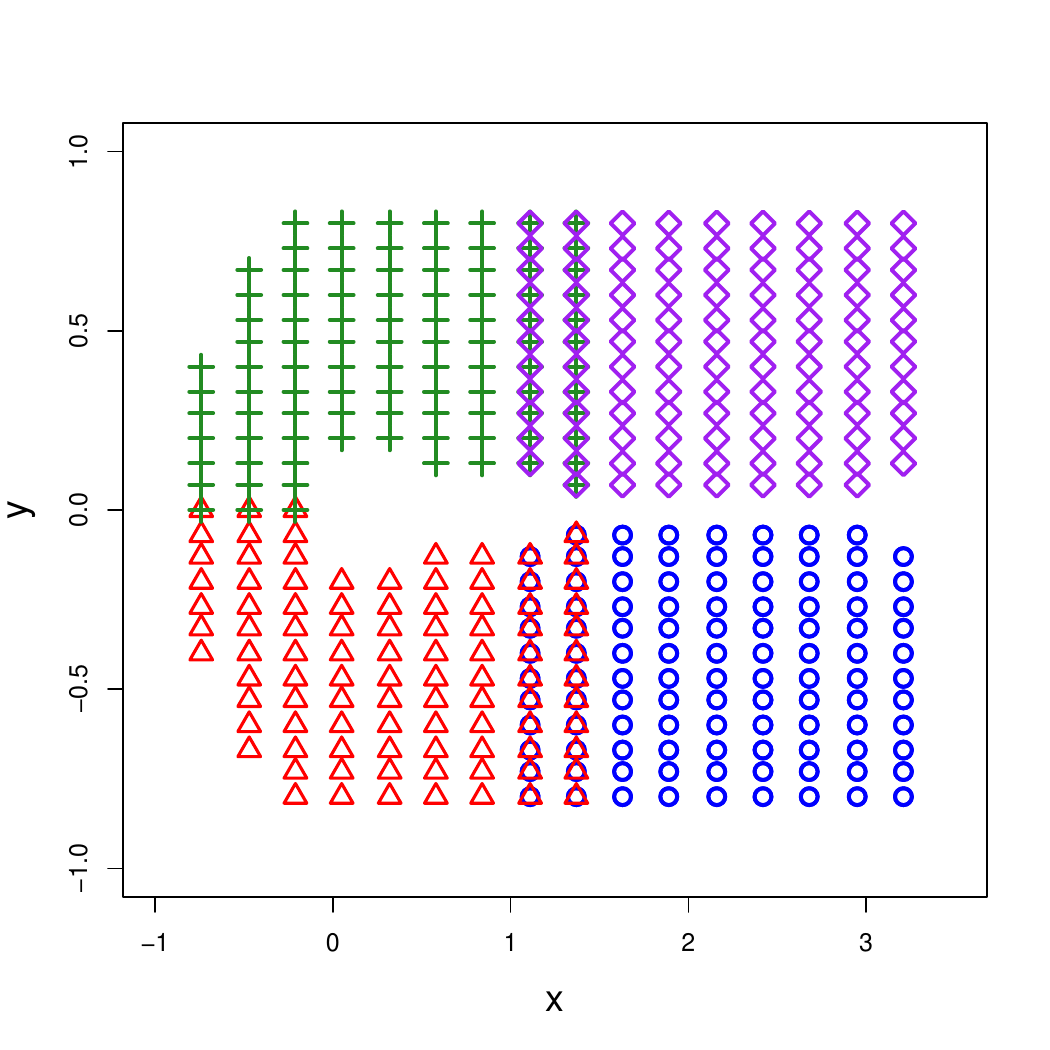}
    \caption{The U-shape subsets}
    \label{fig:ushape-sub}
\end{figure}

\begin{figure}[h!]
    \centering
    \subfigure[True function in U-shape]{ \label{fig:UtrFun}  \includegraphics[width=0.44\textwidth,height=0.4\textwidth]{Fig/u_true.pdf}}
 \subfigure[S-AGP prediction]{ \label{fig:URC2GP}   \includegraphics[width=0.44\textwidth,height=0.4\textwidth]{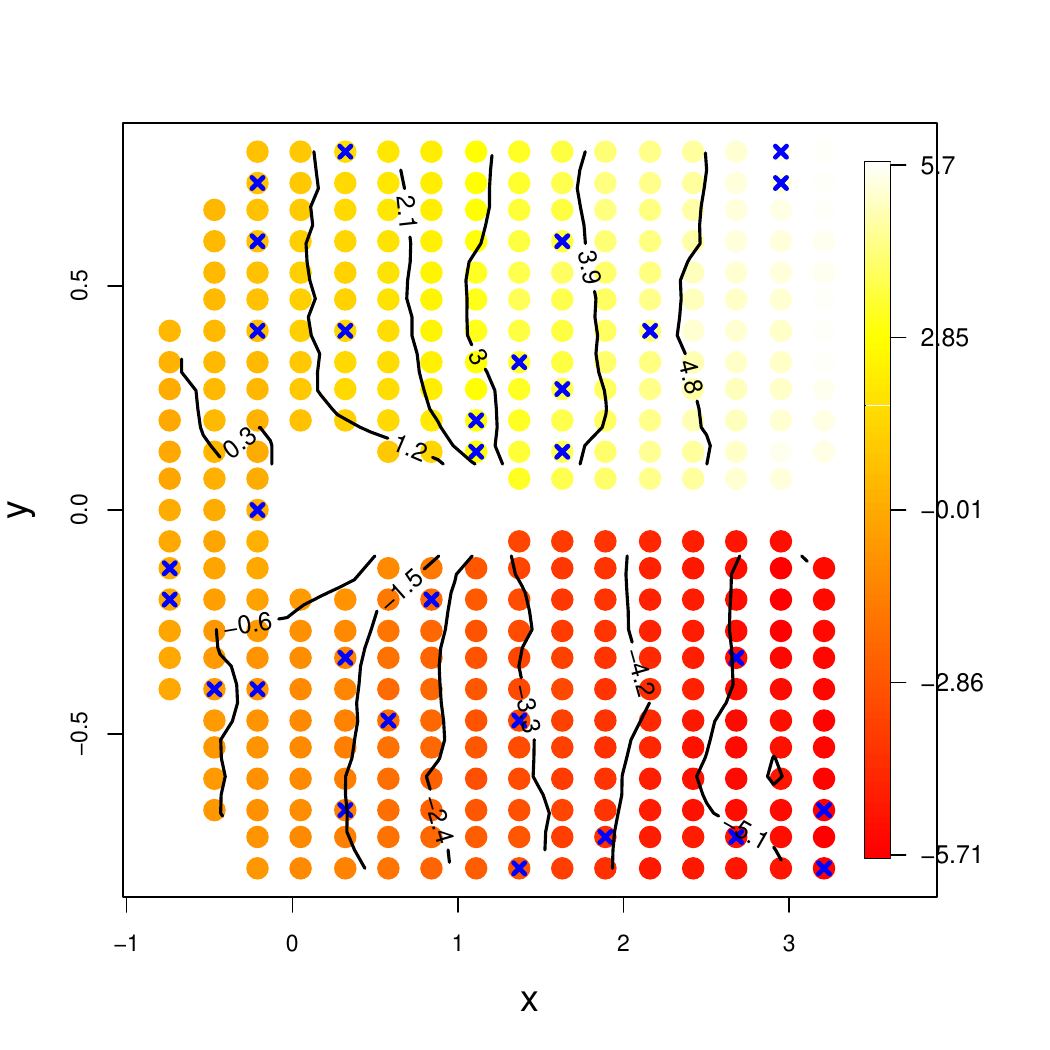}} 
 \caption{\label{fig:udomain}
    \footnotesize
    {\bf True function and prediction in U-shape: (a) true function in the point cloud(blue crosses represent the training observations); (b) S-AGP prediction.   }  }\label{fig:Ushape} 
\end{figure}

Fig.~\ref{fig:Ushape} compares the true function values and the S-AGP prediction for the U-shape point cloud. Four subsets are defined in Fig.~\ref{fig:ushape-sub}. The blue crosses indicate the 30 randomly selected observations used for the analysis. The true function values range from approximately -6 at the lower-right part to 6 at the upper-right part in Fig.~\ref{fig:UtrFun}. The contours of the S-AGP predictive mean in Fig.~\ref{fig:URC2GP} closely match the true function, although the contour line is slightly irregular compared to the true function. The values in the lower-right and upper-right parts of the plot show no interaction, suggesting that the performance of S-AGP is better than that of both GL-GP and $\R^2$GP.  


\begin{figure}[h!]
    \centering
    \subfigure[\footnotesize Latent Space of subset $S_1$]{ \label{fig:fishl1}  \includegraphics[width=0.3\textwidth,height=0.3\textwidth]{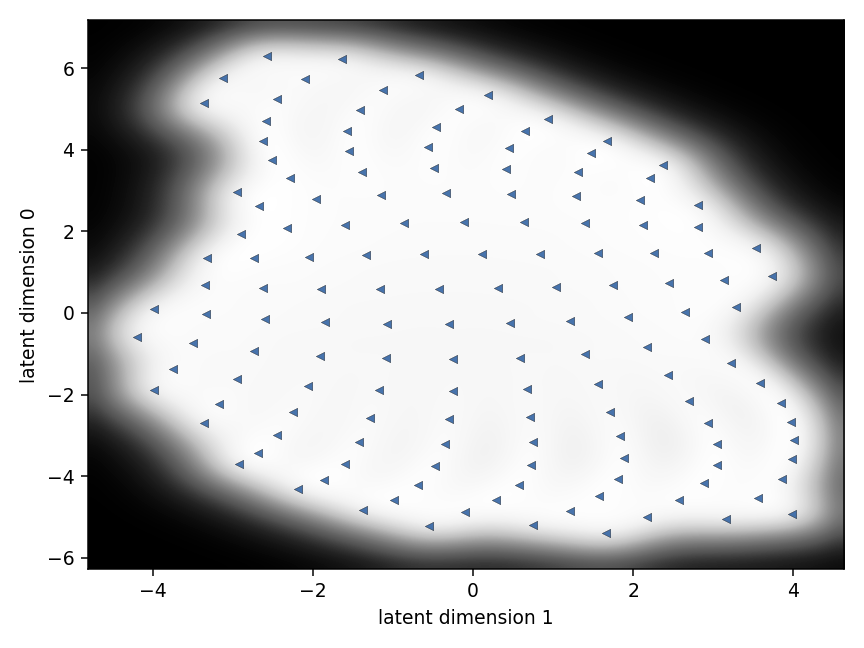}}
    \subfigure[\footnotesize Latent Space of subset $S_2$]{ \label{fig:fishl2}  \includegraphics[width=0.3\textwidth,height=0.3\textwidth]{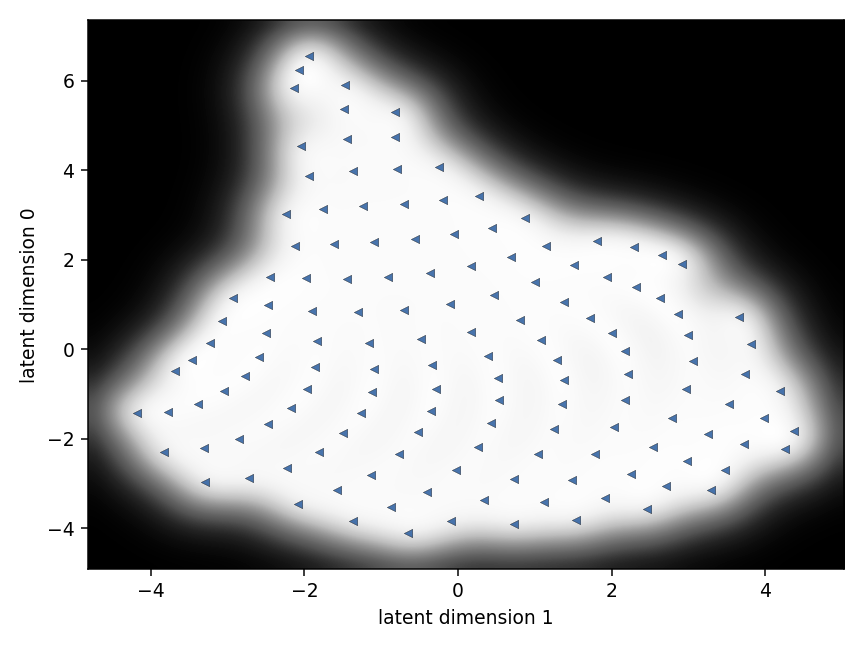}}  
 \subfigure[\footnotesize Latent Space of subset $S_3$]{ \label{fig:fishl3}   \includegraphics[width=0.3\textwidth,height=0.3\textwidth]{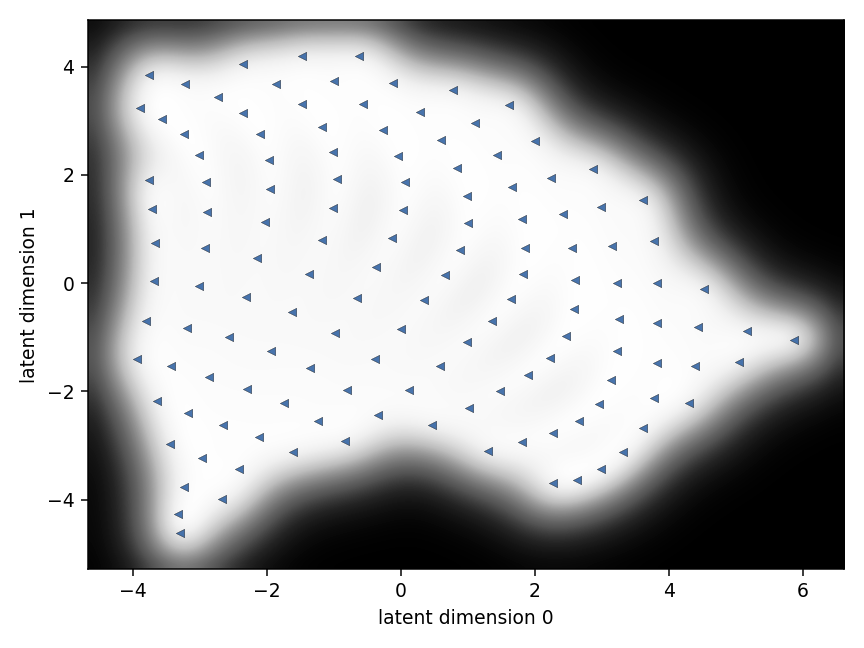}}
 \caption{ \bf Visualization of three latent spaces among eight for shark prey dataset; The background color indicates the variance of the mapping, with darker areas representing higher uncertainty in the latent representation. }\label{fig:fish_l} 
\end{figure}

The shark-prey LEGO image dataset is divided into eight subsets, with each subset used to construct a latent space through GPLVM.
Fig.~\ref{fig:fish_l} visualizes three of these latent spaces. The background color reflects the variance of the mapping, with darker areas indicating higher uncertainty. The shapes of the bright regions differ across the three latent spaces: the points in Fig.~\ref{fig:fishl1} and Fig.~\ref{fig:fishl3} form quadrilateral patterns, while those in Fig.~\ref{fig:fishl2} have a triangular shape.

\begin{figure}[h!]
    \centering
    \subfigure[Image cloud in 3D]{ \label{fig:SP-3d}  \includegraphics[width=0.44\textwidth,height=0.4\textwidth]{Fig/FishTrue.png}}  
 \subfigure[S-AGP prediction]{ \label{fig:SP-}   \includegraphics[width=0.44\textwidth,height=0.4\textwidth]{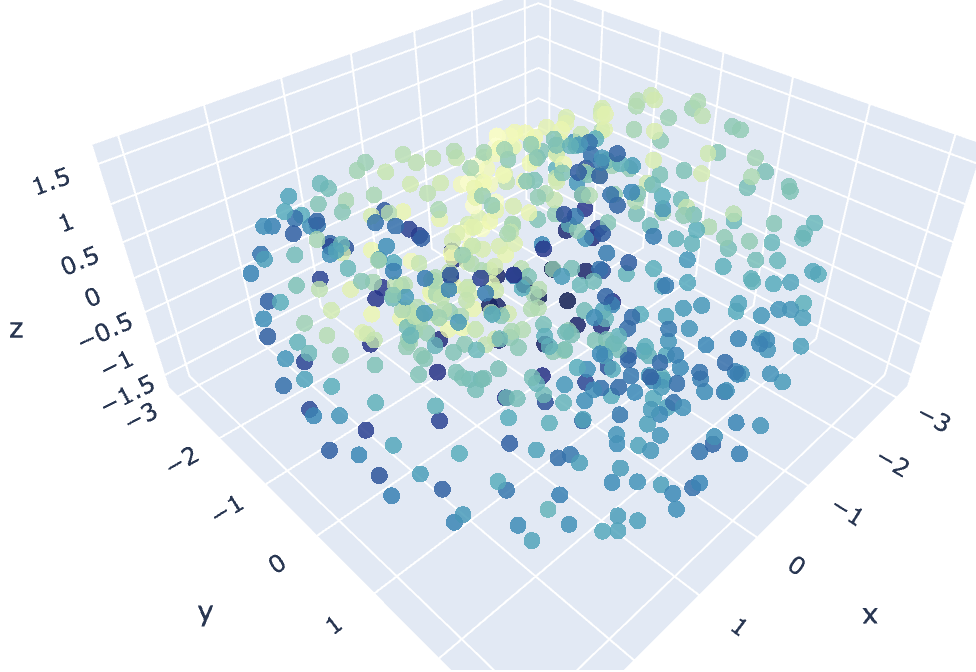}}
 \caption{
    \footnotesize
    {\bf Comparison of different methods in Shark prey LEGO image cloud:(a) true function in the point cloud in 3D;  (b) S-AGP prediction.   }  }\label{fig:SP}
\end{figure}

For better visualization of the regression function on shark prey images, each point in Fig.~\ref{fig:SP}, projected in $\R^3$ using PCA, represents an image with the color scale corresponding to the function values. Although the S-AGP predictions shown in Fig.~\ref{fig:SP-} generally maintain the overall pattern of the true function displayed in Fig.~\ref{fig:SP-3d}, there is a noticeable discrepancy where the lower central points are shown in green rather than dark blue, differing from that of the true function.

\begin{figure}[h!]
    \centering
    \includegraphics[width=0.5\linewidth]{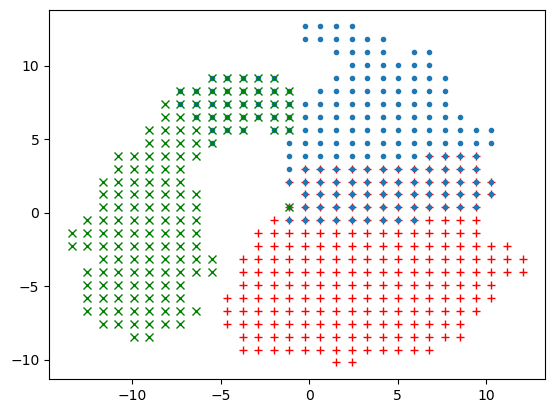}
    \caption{The Aral sea subsets}
    \label{fig:aral-sub}
\end{figure}
\begin{figure}[h!]
    \centering
    \subfigure[Truth in Aral sea]{ \label{fig:Apoint}  \includegraphics[width=0.44\textwidth,height=0.4\textwidth]{Fig/Aral_true.pdf}}
    \subfigure[S-AGP prediction]{ \label{fig:ARC2GP}   \includegraphics[width=0.44\textwidth,height=0.4\textwidth]{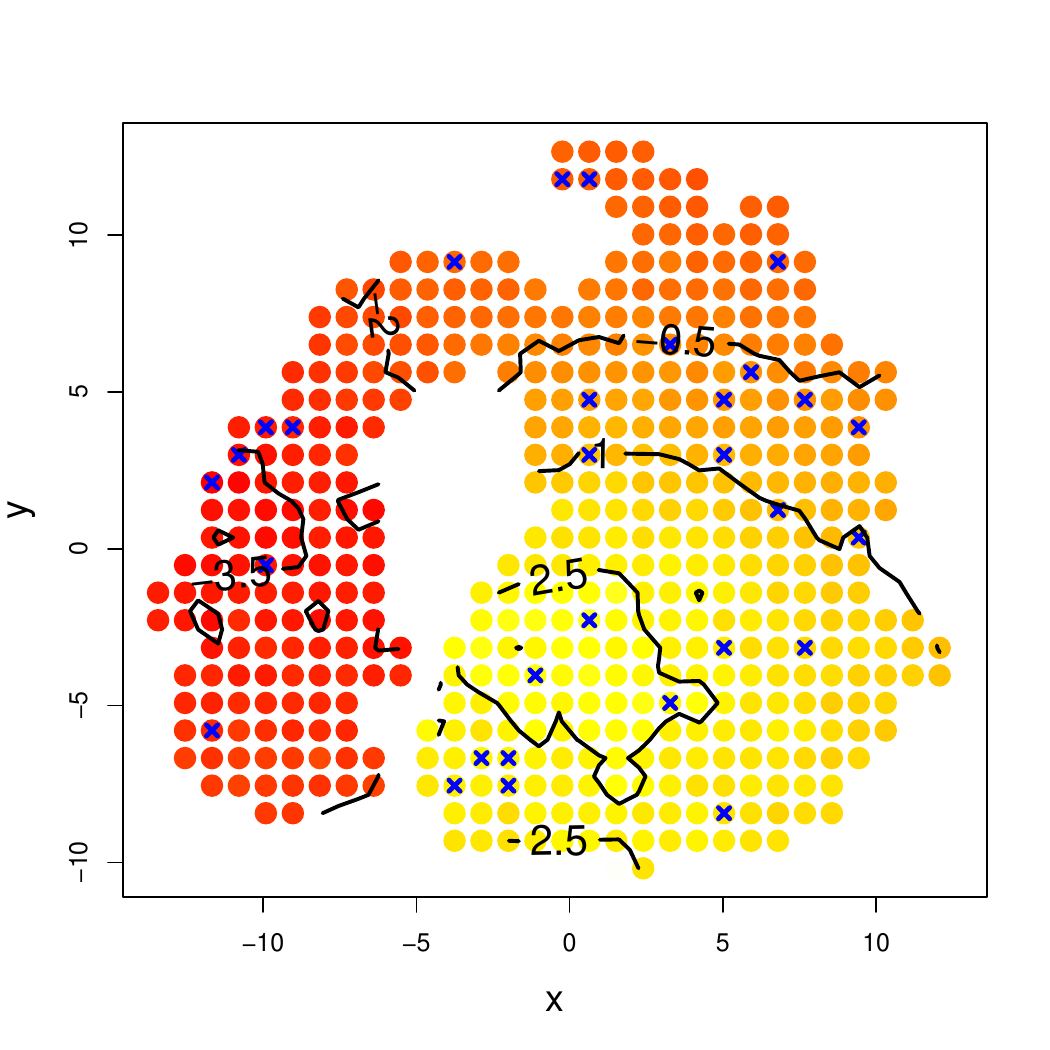}} 
 \caption{\label{fig:Aralsea}
    \footnotesize
    {\bf Comparison of different methods in U-shape: (a) true function in the point cloud; (b) S-AGP prediction at the grid points (blue crosses represent the training observations).  }  }\label{fig:aral}
\end{figure}

The subsets of the Aral Sea point cloud are depicted in Fig.\ref{fig:aral-sub}. A comparison between the S-AGP predictions and the true chlorophyll concentration levels in the Aral Sea is provided in Fig.\ref{fig:aral}. In Fig.\ref{fig:Apoint}, the color intensity represents the true chlorophyll concentration across 485 distinct locations, with blue crosses marking 30 randomly selected observations. The S-AGP prediction shown in Fig.\ref{fig:ARC2GP} broadly reflects a similar pattern to the actual chlorophyll concentration, with the exception of the rightmost region, where the predictions exhibit over-smoothing.


\section{Brownian motion on Torus.}

The three-dimensional coordinates of a Torus $\in \R^3$ can be parametrised by four variables: $r$ radius of tube, $R$ distance from centre of the tube to the centre of the torus, $\theta$ and $\phi$ are angles to make full circles while $\theta$ for angle of torus and $\phi$ for angle of tube. In our case, we fix $R=2$ and $r=1.3$ and let $\theta \in [0,2\pi]$ and $\phi \in [0,2\pi]$. Consider the Torus parametrised by
\begin{align*}
\mathbf{X}(\theta,\phi) &= \left( \left(R + r \cos \theta \right) \cos \phi, \left(R+r\cos \theta\right) \sin\phi, r\sin \theta \right)
\end{align*}

To find its metric tensor, we first compute the partial derivatives
\begin{align*}
\mathbf{X}_{\phi} &= \left( \left(R+r\cos \theta \right) \left(-\sin \phi \right), \left( R + r\cos \theta \right) \cos\phi, 0 \right) \\
\mathbf{X}_{\theta} &= \left( r\cos\phi  \left( - \sin \theta \right),r\sin \phi  \left( -\sin \theta \right),r\cos \theta \right)
\end{align*}

The metric tenor is given by
\begin{align*}
&\quad(\mathbf{X}_{\theta}\cdot\mathbf{X}_{\theta})d\theta^2
+2(\mathbf{X}_{\theta}\cdot\mathbf{X}_{\phi})d\theta\,d\phi
+(\mathbf{X}_{\phi}\cdot\mathbf{X}_{\phi})d\phi^2 \\
&= r^2 d\theta^2 + (R+r\cos\theta)^2d\phi^2
\end{align*}

or in matrix form
\begin{align*}
g= \left[\begin{array}{cc}  
r^2 & 0 \\[0.3em]
0 & (R+r\cos\theta)^2 \\[0.3em]
           \end{array} \right]
\end{align*}

\begin{align*}
 g^{-1}= \left[\begin{array}{cc}  
\frac{1}{r^2} & 0 \\[0.3em]
0 & \frac{1}{(R+r\cos\theta )^2} \\[0.3em]
           \end{array} \right], \qquad
\frac{\partial g}{\partial \theta}= \left[\begin{array}{cc}  
0 &0 \\[0.3em]
 0 & -2(R+r\cos\theta)r\sin\theta \\[0.3em]
                \end{array} \right]            
\end{align*}

By substituting each term into the general equation of BM on manifold, the BM on Torus can be written as
\begin{align}
d\theta(t) &= \frac{1}{2}(-g^{-1} \frac{\partial g}{\partial \theta}g^{-1})_{11} dt + \frac{1}{4} (g^{-1})_{11} tr( g^{-1}\frac{\partial g}{\partial \theta} )dt + (g^{-1/2})_{11} dB_1(t) \\
d\theta(t) &=  -\frac{1}{2} r^{-1}\sin\theta (R+r\cos\theta)^{-1} dt + r^{-1}dB_1(t) \nonumber \\
d\phi(t) &= (g^{-1/2})_{22} dB_2(t) \\
d\phi(t) &=|(R+r\cos\theta)^{-1}| dB_2(t) \nonumber
\end{align}



\begin{figure}
    \centering
    \includegraphics[width=0.7\textwidth,height=0.45\textwidth]{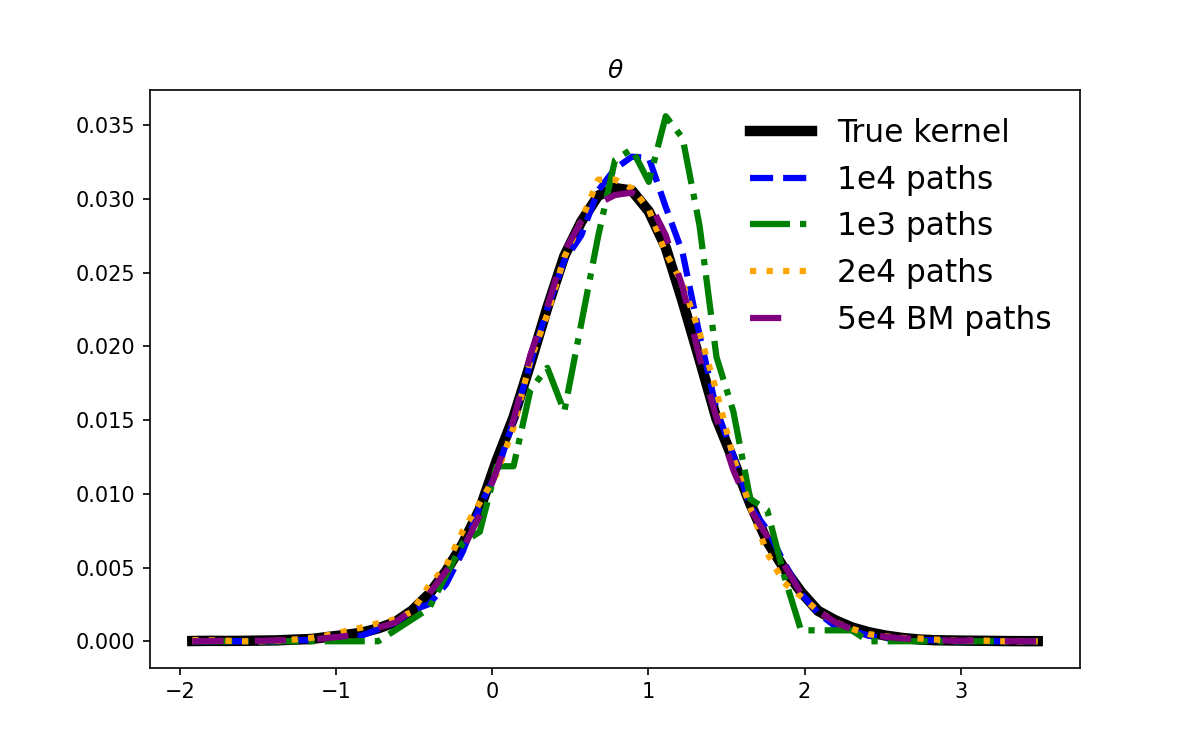}
    \caption{ \label{fig:toruskernel}Comparison of estimates of BM transition density and the analytical kernel on Torus. The black solid line represents the analytical kernel. Estimates obtained from 1,000 simulated paths are shown as the green dash-dotted line, from $10^4$ paths as the blue dashed line, from $2\times10^4$ paths as the yellow dotted line, and from $5\times10^4$ paths as the purple dashed line.
 }
\end{figure}

With the aid of the analytical parameterization of the Torus introduced in this section, we plot the analytical heat kernel as the solid black line in Fig.
\ref{fig:toruskernel}, the ground-truth heat kernel is evaluated at 50 uniformly spaced points $\theta \in (-2,3.5)$ with a diffusion time $t=2$. When using 1,000 simulated paths, the estimated density exhibits noticeable fluctuations. Increasing the number of paths to 10,000 substantially improves the smoothness of the estimates and brings them closer to the analytical kernel, although minor discrepancies persist, particularly near the density peak around $\theta = 0.75$. When the number of simulated paths increases to 20,000 or more, the estimated kernel closely aligns with the ground truth across the entire domain.

\section{Decompose the point cloud into subsets}


Decomposing a point cloud into subsets with simple topological structures has been explored through various approaches \citep{singh2007topological,schonsheck2019chart}. In this work, we follow the framework proposed by \cite{singh2007topological} and employ Mapper as the primary tool for decomposition.  Mapper is a powerful tool that combines dimensionality reduction and clustering to create informative combinatorial representations of high-dimensional data. The process begins by generating a low-dimensional projection using methods such as ISOMAP~\citep{tenenbaum2000global}. Clustering is then applied in the low dimensional space using algorithms such as k-means~\citep{macqueen1967some}. To ensure connectivity and overlap between clusters, Mapper constructs a graph where each data point is linked to its K nearest neighbours, allowing clusters to expand along these connections. This overlap enables certain data points to belong to multiple clusters, facilitating the construction of overlapping coordinate domains.
By clustering these subsets in the original data space, we generate an informative decomposition of the point cloud.  The topological features of each subset can further be analyzed using persistence diagrams \citep{carlsson2009,edelsbrunner2022}.

A natural question is how many charts are required to sufficiently cover the manifold? Theoretically, the minimum number of charts corresponds to the Lusternik-Schnirelmann category of the manifold $\M$ \citep{cornea2003lusternik}. However, it is generally infeasible to compute. An alternative, practical approach suggested by \citet{floryan2022} involves iteratively sampling combinations of chart numbers and dimensions to identify an optimal configuration. A two-step strategy can further simplify this task: first, determine the manifold's intrinsic dimensionality by analyzing small, local data patches, and then tune the number of charts required for full coverage. Manifold dimensionality, being a global property, can be reliably estimated using local methods such as those proposed by \citet{camastra2016intrinsic} and \citet{zwiessele2017}.

In this work, we leverage the Automatic Relevance Determination (ARD) mechanism within the GPLVM framework~\citep{zwiessele2017} to estimate the manifold's dimensionality. ARD assigns weights to latent dimensions, allowing irrelevant dimensions to be effectively pruned. Once the dimensionality is determined, the number and arrangement of charts are adjusted empirically to achieve comprehensive coverage. Graph-theoretical tools and techniques from topological data analysis provide additional insights into the optimal chart configuration. Developing systematic methods for determining the optimal number of charts remains an intriguing direction for future research.

\vskip 0.2in
\bibliography{heatkern}

@article{anderson1946,
  title={The non-central Wishart distribution and certain problems of multivariate statistics},
  author={Anderson, Theodore W},
  journal={The Annals of Mathematical Statistics},
  pages={409--431},
  year={1946},
  publisher={JSTOR}
}

@article{bolin2022,
  title={Gaussian {W}hittle-{M}at{\'e}rn fields on metric graphs},
  author={Bolin, David and Simas, Alexandre B and Wallin, Jonas},
  journal={arXiv preprint arXiv:2205.06163},
  year={2022}
}

@inproceedings{baldi2012,
  title={Autoencoders, unsupervised learning, and deep architectures},
  author={Baldi, Pierre},
  booktitle={Proceedings of ICML workshop on unsupervised and transfer learning},
  pages={37--49},
  year={2012},
  organization={JMLR Workshop and Conference Proceedings}
}

@book{chavel1984,
  title={Eigenvalues in Riemannian geometry},
  author={Chavel, Isaac},
  year={1984},
  publisher={Academic press}
}

@article{carlsson2009,
  title={Topology and data},
  author={Carlsson, Gunnar},
  journal={Bulletin of the American Mathematical Society},
  volume={46},
  number={2},
  pages={255--308},
  year={2009}
}

@article{camastra2016intrinsic,
  title={Intrinsic dimension estimation: Advances and open problems},
  author={Camastra, Francesco and Staiano, Antonino},
  journal={Information Sciences},
  volume={328},
  pages={26--41},
  year={2016},
  publisher={Elsevier}
}

@book{cornea2003lusternik,
  title={Lusternik-Schnirelmann category},
  author={Cornea, Octavian},
  number={103},
  year={2003},
  publisher={American Mathematical Soc.}
}

@article{dunson2020diffusion,
  title={Diffusion Based {G}aussian processes on Restricted Domains},
  author={Dunson, David and Wu, Hau-Tieng and Wu, Nan},
  journal={arXiv preprint arXiv:2010.07242},
  year={2020}
}

@book{edelsbrunner2022,
  title={Computational topology: an introduction},
  author={Edelsbrunner, Herbert and Harer, John L},
  year={2022},
  publisher={American Mathematical Society}
}

@article{floryan2022,
  title={Data-driven discovery of intrinsic dynamics},
  author={Floryan, Daniel and Graham, Michael D},
  journal={Nature Machine Intelligence},
  volume={4},
  number={12},
  pages={1113--1120},
  year={2022},
  publisher={Nature Publishing Group UK London}
}

@article{garcia2020,
  title={Error estimates for spectral convergence of the graph Laplacian on random geometric graphs toward the Laplace--Beltrami operator},
  author={Garc{\'\i}a Trillos, Nicol{\'a}s and Gerlach, Moritz and Hein, Matthias and Slep{\v{c}}ev, Dejan},
  journal={Foundations of Computational Mathematics},
  volume={20},
  number={4},
  pages={827--887},
  year={2020},
  publisher={Springer}
}

@article{he2024scalable,
  title={Scalable Bayesian inference for heat kernel Gaussian processes on manifolds},
  author={He, Junhui and Ma, Guoxuan and Kang, Jian and Yang, Ying},
  journal={arXiv preprint arXiv:2405.13342},
  year={2024}
}

@article{hsu1988,
  title={Brownian motion and {R}iemannian geometry},
  author={Hsu, Pei},
  journal={Contemporary Mathematics},
  volume={73},
  pages={95--104},
  year={1988}
}

@article{hsu2008,
  title={A brief introduction to {B}rownian motion on a {R}iemannian manifold},
  author={Hsu, Elton P},
  journal={Lecture Notes},
  year={2008}
}

@article{KeplerMapper2019,
    doi           = {10.21105/joss.01315},
    url           = {https://doi.org/10.21105/joss.01315},
    year          = {2019},
    publisher     = {The Open Journal},
    volume        = {4},
    number        = {42},
    pages         = {1315},
    author        = {Hendrik Jacob van Veen and Nathaniel Saul and David Eargle and Sam W. Mangham},
    title         = {Kepler Mapper: A flexible Python implementation of the Mapper algorithm.},
    journal       = {Journal of Open Source Software}
    }

@article{lawrence2005,
  title={Probabilistic non-linear principal component analysis with {G}aussian process latent variable models},
  author={Lawrence, Neil},
  journal={Journal of machine learning research},
  volume={6},
  number={Nov},
  pages={1783--1816},
  year={2005}
}

@book {Lee13,
    AUTHOR = {Lee, John M.},
     TITLE = {Introduction to smooth manifolds},
    SERIES = {Graduate Texts in Mathematics},
    VOLUME = {218},
   EDITION = {Second},
 PUBLISHER = {Springer, New York},
      YEAR = {2013},
     PAGES = {xvi+708},
      ISBN = {978-1-4419-9981-8},
   MRCLASS = {58-01 (53-01 57-01)},
  MRNUMBER = {2954043},
}

@book {Lee18,
    AUTHOR = {Lee, John M.},
     TITLE = {Introduction to {R}iemannian manifolds},
    SERIES = {Graduate Texts in Mathematics},
    VOLUME = {176},
   EDITION = {Second},
 PUBLISHER = {Springer, Cham},
      YEAR = {2018},
     PAGES = {xiii+437},
      ISBN = {978-3-319-91754-2; 978-3-319-91755-9},
   MRCLASS = {53-01 (53B20 53B30 53C20 53C21)},
  MRNUMBER = {3887684},
MRREVIEWER = {Robert\ J.\ Low},
}

@article{moreno2022,
  title={Revisiting active sets for gaussian process decoders},
  author={Moreno-Mu{\~n}oz, Pablo and Feldager, Cilie and Hauberg, S{\o}ren},
  journal={Advances in Neural Information Processing Systems},
  volume={35},
  pages={6603--6614},
  year={2022}
}

@article{niu2019,
  title={Intrinsic {G}aussian processes on complex constrained domains},
  author={Niu, Mu and Cheung, Pokman and Lin, Lizhen and Dai, Zhenwen and Lawrence, Neil and Dunson, David},
  journal={Journal of the Royal Statistical Society: Series B (Statistical Methodology)},
  volume={81},
  number={3},
  pages={603--627},
  year={2019},
  publisher={Wiley Online Library}
}

@article{niu2023,
  title={Intrinsic gaussian process on unknown manifolds with probabilistic metrics},
  author={Niu, Mu and Dai, Zhenwen and Cheung, Pokman and Wang, Yizhu},
  journal={Journal of Machine Learning Research},
  volume={24},
  number={104},
  pages={1--42},
  year={2023}
}

@article{paik2022atlas,
  title={Atlas flow: compatible local structures on the manifold},
  author={Paik, Taejin and Park, Jaemin and Park, Jung Ho},
  journal={arXiv preprint arXiv:2210.14149},
  year={2022}
}

@inbook{quin07,
  title = {Approximation Methods for Gaussian Process Regression},
  author = {Quiñonero-Candela, J. and Rasmussen, CE. and Williams, CKI.},
  booktitle = {Large-Scale Kernel Machines},
  pages = {203-223},
  series = {Neural Information Processing},
  editors = {Bottou, L. , O. Chapelle, D. DeCoste, J. Weston},
  publisher = {MIT Press},
  month = sep,
  year = {2007},
  month_numeric = {9}
}

@book{Rasmussen2006,
  title={Gaussian processes for machine learning},
  author={Rasmussen, Carl Edward and Williams, Christopher KI },
  volume={2},
  issue={3},
  year={2006},
  publisher={MIT press Cambridge, MA}
}

@article{smola2000,
  title={Sparse greedy Gaussian process regression},
  author={Smola, Alex and Bartlett, Peter},
  journal={Advances in neural information processing systems},
  volume={13},
  year={2000}
}

@article{stolberg2022,
  title={Atlas generative models and geodesic interpolation},
  author={Stolberg-Larsen, Jakob and Sommer, Stefan},
  journal={Image and Vision Computing},
  volume={122},
  pages={104433},
  year={2022},
  publisher={Elsevier}
}

@article {tramsay,
author = {Ramsay, Tim},
title = {Spline smoothing over difficult regions},
journal = {J. Royal Stat. Soc. Series B},
volume = {64},
number = {2},
publisher = {Blackwell Science, Ltd},
issn = {1467-9868},
doi = {10.1111/1467-9868.00339},
pages = {307--319},
year = {2002},
}

@article{wood,
 author = {Simon N. Wood and Mark V. Bravington and Sharon L. Hedley},
 journal = {J. Royal Stat. Soc. Series B },
 number = {5},
 pages = {931-955},
 title = {Soap Film Smoothing},
 volume = {70},
 year = {2008}
}

@book{wood2006,
 title={Generalized Additive Models: an Introduction with R},
 author={Wood, Simon},
 edition = {1st},
 address = {Boca Raton, Florida},
 year = {2006},
 publisher = {Chapman and Hall/CRC}
}

@phdthesis{zwiessele2017,
  title={Bringing models to the domain: Deploying gaussian processes in the biological sciences},
  author={Zwiessele, Max},
  year={2017},
  school={University of Sheffield}
}

@article{singh2007topological,
  title={Topological methods for the analysis of high dimensional data sets and 3d object recognition.},
  author={Singh, Gurjeet and M{\'e}moli, Facundo and Carlsson, Gunnar E and others},
  journal={PBG@ Eurographics},
  volume={2},
  pages={091--100},
  year={2007}
}

@inproceedings{titsias2009variational,
  title={Variational learning of inducing variables in sparse Gaussian processes},
  author={Titsias, Michalis},
  booktitle={Artificial intelligence and statistics},
  pages={567--574},
  year={2009},
  organization={PMLR}
}

@article{tenenbaum2000global,
  title={A global geometric framework for nonlinear dimensionality reduction},
  author={Tenenbaum, Joshua B and Silva, Vin de and Langford, John C},
  journal={science},
  volume={290},
  number={5500},
  pages={2319--2323},
  year={2000},
  publisher={American Association for the Advancement of Science}
}

@inproceedings{macqueen1967some,
  title={Some methods for classification and analysis of multivariate observations},
  author={Macqueen, J},
  booktitle={Proceedings of 5-th Berkeley Symposium on Mathematical Statistics and Probability/University of California Press},
  year={1967}
}

@book{hastie2009elements,
  title={The elements of statistical learning: data mining, inference, and prediction},
  author={Hastie, Trevor and Tibshirani, Robert and Friedman, Jerome H and Friedman, Jerome H},
  volume={2},
  year={2009},
  publisher={Springer}
}

@inproceedings{rifai2011contractive,
  title={Contractive auto-encoders: Explicit invariance during feature extraction},
  author={Rifai, Salah and Vincent, Pascal and Muller, Xavier and Glorot, Xavier and Bengio, Yoshua},
  booktitle={Proceedings of the 28th international conference on international conference on machine learning},
  pages={833--840},
  year={2011}
}

@article{galy2021adaptive,
  title={Adaptive inducing points selection for gaussian processes},
  author={Galy-Fajou, Th{\'e}o and Opper, Manfred},
  journal={arXiv preprint arXiv:2107.10066},
  year={2021}
}

@inproceedings{moss2023inducing,
  title={Inducing point allocation for sparse Gaussian processes in high-throughput Bayesian optimisation},
  author={Moss, Henry B and Ober, Sebastian W and Picheny, Victor},
  booktitle={International Conference on Artificial Intelligence and Statistics},
  pages={5213--5230},
  year={2023},
  organization={PMLR}
}

@inproceedings{bishop1997magnification,
  title={Magnification factors for the GTM algorithm},
  author={Bishop, Christopher M and Svens{\'e}n, Markus and Williams, Christopher KI},
  booktitle={Fifth International Conference on Artificial Neural Networks (Conf. Publ. No. 440)},
  pages={64--69},
  year={1997},
  organization={IET}
}

@article{li2023inference,
  title={Inference for gaussian processes with mat{\'e}rn covariogram on compact riemannian manifolds},
  author={Li, Didong and Tang, Wenpin and Banerjee, Sudipto},
  journal={Journal of Machine Learning Research},
  volume={24},
  number={101},
  pages={1--26},
  year={2023}
}

@article{lin2019extrinsic,
  title={Extrinsic Gaussian processes for regression and classification on manifolds},
  author={Lin, Lizhen and Mu, Niu and Cheung, Pokman and Dunson, David},
  year={2019}
}

@article{schonsheck2022semi,
  title={Semi-supervised manifold learning with complexity decoupled chart autoencoders},
  author={Schonsheck, Stefan C and Mahan, Scott and Klock, Timo and Cloninger, Alexander and Lai, Rongjie},
  journal={arXiv preprint arXiv:2208.10570},
  year={2022}
}

@article{schonsheck2019chart,
  title={Chart auto-encoders for manifold structured data},
  author={Schonsheck, Stefan and Chen, Jie and Lai, Rongjie},
  journal={arXiv preprint arXiv:1912.10094},
  year={2019}
}

@book{watkins2012getting,
  title={Getting Started in 3D with Maya: Create a Project from Start to Finish—Model, Texture, Rig, Animate, and Render in Maya},
  author={Watkins, Adam},
  year={2012},
  publisher={Routledge}
}

\end{document}